\documentclass[final, onefignum,onetabnum]{siamonline190516}

\usepackage{ulem}

\usepackage{epstopdf}

\ifpdf
  \DeclareGraphicsExtensions{.eps,.pdf,.png,.jpg}
\else
  \DeclareGraphicsExtensions{.eps}
\fi

\usepackage{bm}
\usepackage{bbm}
\usepackage{cleveref}
\usepackage[inline]{enumitem}
\usepackage{amsopn}
\usepackage{amsfonts}
\usepackage{amssymb}
\usepackage{graphicx}
\usepackage{stmaryrd}
\usepackage{xr-hyper}
\usepackage{subcaption}
\usepackage{algpseudocode}
\usepackage{eqparbox}
\usepackage{mathtools}

\Crefname{ALC@unique}{Line}{Lines}

\usepackage{mymacros}

\graphicspath{ {./figures} }

\usepackage{enumitem}
\setlist[enumerate]{leftmargin=.5in}
\setlist[itemize]{leftmargin=.5in}

\newsiamremark{remark}{Remark}
\newsiamremark{example}{Example}
\crefname{example}{Example}{Example}
\newsiamthm{informaltheorem}{Informal Theorem}
\newsiamthm{fact}{Fact}
\headers{Efficient Identification of Butterfly Sparse Matrix Factorizations}{L. Zheng, E. Riccietti, and R. Gribonval}

\title{Efficient Identification of Butterfly Sparse Matrix Factorizations\thanks{Submitted to the editors on April 1st, 2022, revision sent on July 27th, 2022. This work is an extension of \cite{le:hal-03438881}.\funding{This project was supported in part by the AllegroAssai ANR project ANR-19-CHIA-0009 and the CIFRE fellowship N°2020/1643.}}}

\author{L\'eon Zheng\footnotemark[3] \thanks{Univ. de Lyon, ENS de Lyon, UCBL, CNRS, Inria, LIP, F-69342, LYON Cedex 07, France.
(\email{leon.zheng@ens-lyon.fr}, \email{elisa.riccietti@ens-lyon.fr}, \email{remi.gribonval@inria.fr}). $^\ddag$valeo.ai, Paris, France.} 
\and Elisa Riccietti\footnotemark[2]
\and R\'emi Gribonval\footnotemark[2]}

\ifpdf
\hypersetup{
	pdftitle={Efficient Identification of Butterfly Sparse Matrix Factorizations},
	pdfauthor={L. Zheng, E. Riccietti, and R. Gribonval}
}
\fi

\begin{document}
	
	\maketitle
	
	\normalem 

	\begin{abstract}
		Fast transforms correspond to factorizations of the form $\mat{Z} = \matseq{\mat{X}}{1} \ldots \matseq{\mat{X}}{J}$, where each factor $\matseq{\mat{X}}{\ell}$ is sparse and possibly structured. 
		This paper investigates \emph{essential uniqueness} of such factorizations, i.e., uniqueness up to unavoidable scaling ambiguities.
		Our main contribution is to prove that any $N \times N$ matrix having the so-called \emph{butterfly} structure admits an essentially unique factorization into $J$ butterfly factors (where $N = 2^{J}$), and that the factors can be recovered by a hierarchical factorization method,  which consists in recursively factorizing the considered matrix into two factors. This hierarchical identifiability property relies on a simple identifiability condition in the two-layer and fixed-support setting. 
		This approach contrasts with existing ones that fit the product of butterfly factors to a given matrix via gradient descent. The proposed method can be applied in particular to retrieve the factorization of the Hadamard or the discrete Fourier transform matrices of size $N=2^J$.
		Computing such factorizations costs $\mathcal{O}(N^{2})$, which is of the order of dense matrix-vector multiplication, while the obtained factorizations enable fast $\mathcal{O}(N \log N)$ matrix-vector multiplications and have  the potential to be applied to compress deep neural networks.
	\end{abstract}
	
	

	\section{Introduction}
	Sparse matrix factorization with $J \geq 2$ factors is the problem of approximating a given matrix $\mat{Z}$ by a product of $J$ sparse factors $\matseq{\mat{X}}{1} \matseq{\mat{X}}{2} \ldots \matseq{\mat{X}}{J}$. 
	Such a factorization is desired to reduce time and memory complexity for numerical methods involving the linear operator associated to $\mat{Z}$, e.g., in large-scale linear inverse problems \cite{elad2010sparse, 5325694, 7178579, le2016flexible}. 
	It can also be applied for compressing deep neural networks \cite{dao2019learning,dao2020kaleidoscope, lin2021deformable,dao2022monarch, chen2022pixelated}: indeed, dense weight matrices of very large size in modern neural network architectures, like in transformers \cite{devlin2019bert,dosovitskiy2021an} or MLP-mixers \cite{tolstikhin2021mlp}, can be replaced by products of sparse and possibly structured matrices in order to reduce the number of parameters of the model and the number of FLOPs.
	
	The generic sparse matrix factorization problem is formulated as follows:
	\begin{equation}
		\label{eq:smf-general}
		\min_{\matseq{\mat{X}}{1}, \ldots, \matseq{\mat{X}}{J}} \| \mat{Z} - \matseq{\mat{X}}{1} \matseq{\mat{X}}{2} \ldots \matseq{\mat{X}}{J} \|_F, \; \text{ such that } \matseq{\mat{X}}{\ell} \text{ is sparse for all } \ell  = 1, \ldots, J,
	\end{equation}
	where $\| \cdot \|_F$ denotes the Frobenius norm. 
	There are typically two kinds of sparsity constraints:
	\begin{enumerate}
		\item \emph{Classical sparsity constraints}: they are encoded by a family of \emph{allowed} supports that force the factors to have some prescribed sparsity patterns, like $k$-sparsity by row and/or column (i.e., each factor has at most $k$ nonzero entries per row and/or column).
		\item \emph{Fixed-support constraints}: the supports (i.e., the set of indices corresponding to nonzero entries in a matrix) are known a priori and each factor $\matseq{\mat{X}}{\ell}$ is constrained to have a support included in a given prescribed support $\matseq{\mat{S}}{\ell}$, $\ell = 1, \ldots, J$.
	\end{enumerate}
	
	Problem \eqref{eq:smf-general} is known to be difficult in general, but even specific instances have been shown to be NP hard. On the one hand, problem \eqref{eq:smf-general} with classical sparsity constraints and $J=2$ factors is the sparse coding problem \cite{foucart_mathematical_2013, elad2010sparse}, when the dictionary $\matseq{\mat{X}}{1}$ is known and the factor $\matseq{\mat{X}}{2}$ is constrained to be $k$-sparse by column. This specific problem is shown to be NP-hard in \cite[Theorem 2.17]{foucart_mathematical_2013}. On the other hand, even in the case with $J=2$ factors and  fixed-support constraints, problem \eqref{eq:smf-general} has been recently shown to be NP-hard, without further assumptions on the prescribed fixed supports \cite{papierTung}.
	
	In this paper, we identify a particular  choice of fixed-support constraint that makes problem \eqref{eq:smf-general} both well-posed and tractable, in the sense that problem \eqref{eq:smf-general} in the noiseless setting  admits a unique\footnote{up to unavoidable scaling ambiguities.} solution that can be computed with an algorithm of polynomial complexity. 
	This fixed-support constraint is the so-called \emph{butterfly} structure (see \Cref{def:butterfly-structure} below).
	As illustrated in \Cref{fig:butterfly-support} for the case with $J=4$ factors, this fixed-support constraint enforces the factors $\matseq{\mat{X}}{\ell} \in \mathbb{C}^{N \times N}$ ($N = 2^J$, $\ell = 1, \ldots, J$) to have at most two nonzero entries per row and per column, with a block diagonal structure for the factors $\matseq{\mat{X}}{2}, \ldots, \matseq{\mat{X}}{J}$. 
	
	\begin{figure}[!h]
		\centering
		\begin{subfigure}[t]{0.23\textwidth}
			\centering
			\includegraphics[width=0.9\textwidth]{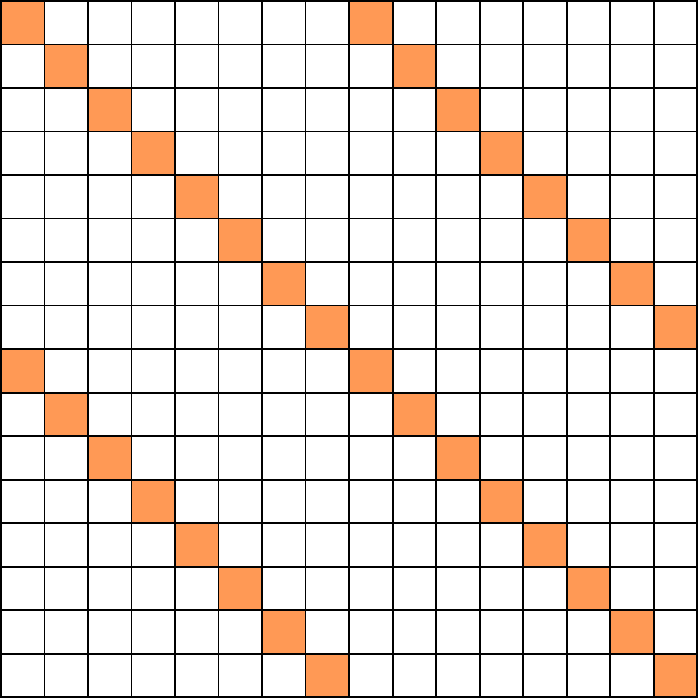}
			\caption{$\bflyindex{1}$}
		\end{subfigure}%
		~ 
		\begin{subfigure}[t]{0.23\textwidth}
			\centering
			\includegraphics[width=0.9\textwidth]{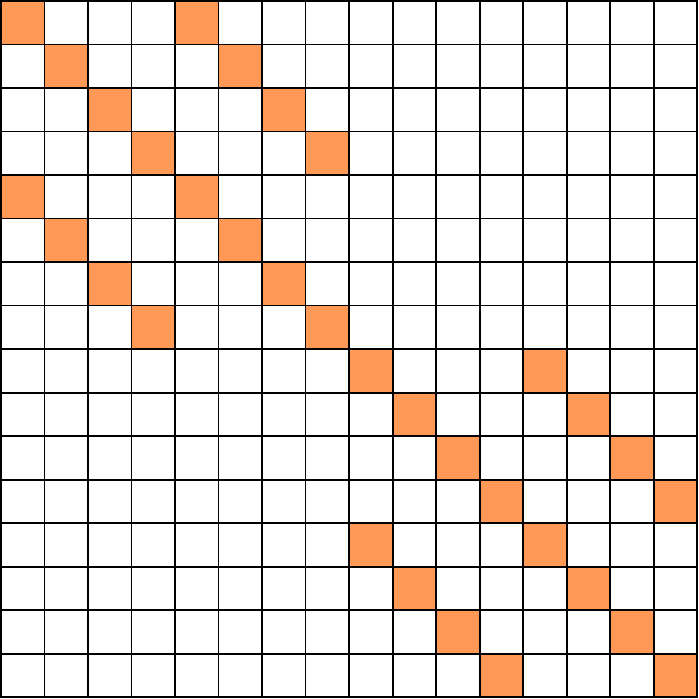}
			\caption{$\bflyindex{2}$}
		\end{subfigure}%
		~ 
		\begin{subfigure}[t]{0.23\textwidth}
			\centering
			\includegraphics[width=0.9\textwidth]{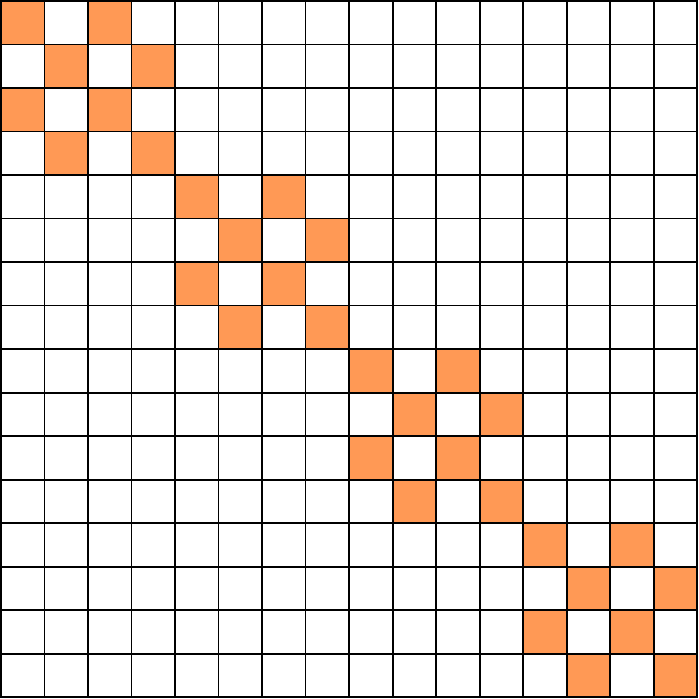}
			\caption{$\bflyindex{3}$}
		\end{subfigure}%
		~ 
		\begin{subfigure}[t]{0.23\textwidth}
			\centering
			\includegraphics[width=0.9\textwidth]{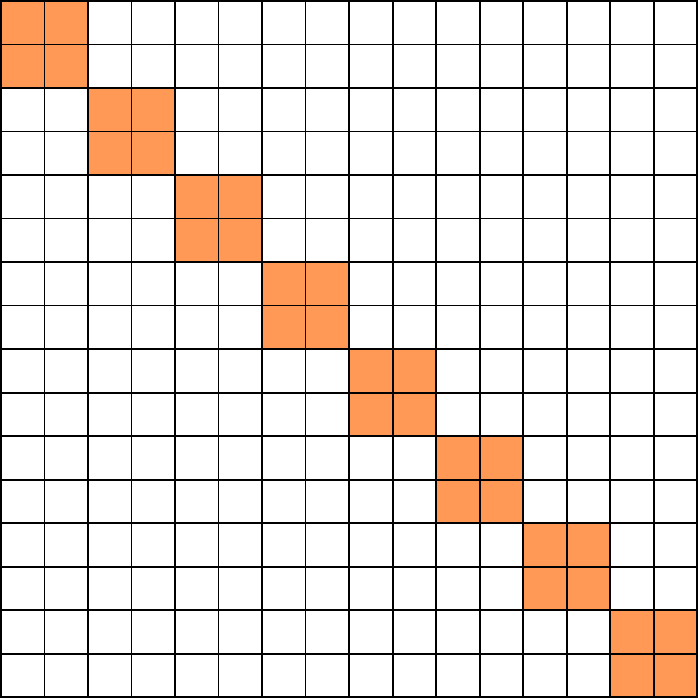}
			\caption{$\bflyindex{4}$}
		\end{subfigure}%
		\caption{Butterfly supports (see \Cref{def:butterfly-suppport} below) of size $N = 16$. Zero entries (white) \emph{vs} entries that are allowed to be nonzero (colored).} 
		\label{fig:butterfly-support}
	\end{figure}

	The butterfly structure is interesting for machine learning applications mainly for two reasons. Firstly, finding the butterfly factors $\matseq{\mat{X}}{\ell}$ ($\ell=1, \ldots, J$) from the product $\mat{Z} := \matseq{\mat{X}}{1} \ldots \matseq{\mat{X}}{J}$ enables \emph{fast} $\mathcal{O}(N\log N)$ matrix-vector multiplication by $\mat{Z}$. Indeed, there are $\mathcal{O}(\log N)$ factors in the butterfly factorization of a matrix of size $N \times N$, and each factor has $\mathcal{O}(N)$ nonzero entries \cite{dao2019learning}. 
	Secondly, the butterfly structure is \emph{expressive}: the composition of matrices with a butterfly structure can accurately approximate any given matrix \cite{dao2019learning, dao2020kaleidoscope}. 
	In practice, the potential of the butterfly structure in terms of expressiveness has been demonstrated in recent empirical works \cite{dao2019learning,lin2021deformable,dao2020kaleidoscope,chen2022pixelated,dao2022monarch}, where it is shown that replacing dense weight matrices in deep neural networks by butterfly-based structured matrices
	can reduce the model's complexity for faster training and inference time \emph{without} harming its performance.	
	The expressiveness property can be clearly illustrated by using the so-called kaleidoscope matrix representation \cite{dao2020kaleidoscope}. Define $\mathcal{B}$ as the class of matrices that admit an exact butterfly factorization; $\mathcal{B} \mathcal{B}^*$ as the class of matrices of the form $\matseq{\mat{M}}{1} {\matseq{\mat{M}}{2}}^*$, with $\matseq{\mat{M}}{1}, \matseq{\mat{M}}{2} \in \mathcal{B}$, and where $*$ denotes the conjugate transpose; $(\mathcal{B} \mathcal{B}^*)^W$ as the class of matrices of the form $\matseq{\mat{M}}{1} \ldots \matseq{\mat{M}}{W}$ with $\matseq{\mat{M}}{w} \in \mathcal{B} \mathcal{B}^*$ for $w = 1, \ldots, W$. 
	Many structured linear maps common in machine learning are shown to be \emph{tightly} captured by this so-called kaleidoscope hierarchy, in the sense that the matrices associated to these linear transforms are in  $(\mathcal{B} \mathcal{B}^*)^W$ 
	with \emph{small} width $W$, resulting in a representation of the linear transform with nearly-optimal memory and time complexity \cite{dao2020kaleidoscope}. For instance, the kaleidoscope hierarchy can express:
	\begin{itemize}
		\item \textbf{Classical fast linear transforms}: this includes the discrete Fourier transform (DFT), the discrete cosine transform, the discrete sine transform and the Hadamard transform. For instance, the Hadamard matrix $\mat{H}$ of dimension a power of two is in $\mathcal{B}$, and the DFT matrix $\mat{F}$ is in $(\mathcal{B} \mathcal{B}^*)^2$, because it can be written as $\mat{F} = \mat{B} \mat{P}$ where $\mat{B} \in \mathcal{B} \subset \mathcal{B}\mathcal{B}^{*}$ and $\mat{P}$ is the bit-reversal matrix, which is shown to be also in $\mathcal{B} \mathcal{B}^*$.
		
		\item \textbf{Circulant matrices}, which are associated to convolutions: any circulant matrix $\mat{C}$ can be expressed as $\mat{C} = \mat{F}^{-1} \mat{D} \mat{F}$ where $\mat{D}$ is a diagonal matrix by \cite[Theorem 2.6.4]{pan2001structured}, and $\mat{C} \in (\mathcal{B} \mathcal{B}^*)^4$, since the DFT matrix $\mat{F}$ as well as its inverse $\mat{F}^{-1} = \mat{F}^{*}$, and their scaled versions $\mat{DF}$, $\mat{F}^{-1}\mat{D}$ are all in $(\mathcal{B} \mathcal{B}^*)^2$. In fact a tighter analysis can show that any circulant matrix $\mat{C}$ is in $\mathcal{B} \mathcal{B}^*$ \cite[Lemma J.5]{dao2020kaleidoscope}.
		
		\item \textbf{Toeplitz matrices}: for any Toeplitz matrix $\mat{T}$ of size $N \times N$, there exists a circulant matrix of size $2N \times 2N$ such that $\mat{T} = \mat{R} \mat{C} \mat{R}^*$
		with $\mat{R} := \left[ \begin{smallmatrix} \identity{N} & \matsize{0}{N} \end{smallmatrix} \right] \in \mathbb{R}^{N \times 2N}$ a reduced identity matrix ($\identity{N}$ denotes the identity matrix of size $N$).
		Moreover, any matrix $\mat{M}$ of size $N \times N$ can be expressed as a product $\mat{M} := \matseq{\mat{T}}{1} \matseq{\mat{T}}{2} \ldots \matseq{\mat{T}}{2N+5}$ of (at most) $2N + 5$ Toeplitz matrices 
		 \cite{ye2016every}, hence
		can be written as $\mat{R}\mat{N}\mat{R}^{*}$ with $\mat{N} \in (\mathcal{B} \mathcal{B}^*)^{2N+5}$.
		Indeed, writing $\matseq{\mat{T}}{i} = \mat{R} \matseq{\mat{C}}{i} \mat{R}^*$ with $\matseq{\mat{C}}{i} \in \mathcal{B} \mathcal{B}^*$ a circulant matrix of size $2N \times 2N$, we have $\mat{M} := \mat{R}\mat{N}\mat{R}^{*}$ with $\mat{N} := \matseq{\mat{C}}{1} (\mat{R}^* \mat{R}) \matseq{\mat{C}}{2} \ldots (\mat{R}^* \mat{R}) \matseq{\mat{C}}{2N+5} \in (\mathcal{B} \mathcal{B}^*)^{2N+5}$, as $(\mat{R}^* \mat{R}) \matseq{\mat{C}}{i} \in  \mathcal{B} \mathcal{B}^*$ for each $i$ due to the fact that $(\mat{R}^* \mat{R})$ is simply a diagonal matrix.

		\item \textbf{Fastfood transform} \cite{le2013fastfood}: the matrix $\mat{V} = \frac{1}{\sigma \sqrt{N}} \mat{S} \mat{H} \mat{G} \mat{P} \mat{H} \mat{B}$, with  $\mat{S}, \mat{G}, \mat{B}$ some diagonal matrices, $\mat{P}$ a permutation matrix and $\mat{H}$ the Hadamard matrix, is used for fast approximation of the Gaussian kernel of scale $\sigma$. Since $\mat{H} \in \mathcal{B} \subseteq \mathcal{B} \mathcal{B}^*$ and $\mat{P} \in \mathcal{B} \mathcal{B}^*$, we get $\mat{V} \in (\mathcal{B} \mathcal{B}^*)^3$, but it is also possible to show that  $\mat{V} \in (\mathcal{B} \mathcal{B}^*)^2$ \cite[Lemma J.7]{dao2020kaleidoscope}. 
	\end{itemize}

	The above properties of the butterfly structure motivate the interest in studying problem \eqref{eq:smf-general} with butterfly constraints, that is, the problem of finding an accurate approximation of any given matrix by a product of butterfly factors. Ideally, one would like to solve this problem efficiently, in order to design a proximal operator \cite{parikh2014proximal} that projects any matrix into the class of butterfly matrices $\mathcal{B}$, and use it to promote sparsity in various learning algorithms, like, e.g., on the learned dictionary matrix in dictionary learning \cite{DictionaryLearning}, or on the learned weight matrices in neural network training.
	However, since problem \eqref{eq:smf-general} with the butterfly constraint is nonconvex, methods based on first-order optimization \cite{dao2019learning} lack theoretical guarantees for finding the optimal solution,
	and their performance is heavily dependent on initialization and hyperparameter tuning. 

	In constrast, this paper shows that every tuple of factors $(\matseq{\mat{X}}{1}, \ldots, \matseq{\mat{X}}{J})$ satisfying the butterfly constraint can be reconstructed \emph{with guarantee} by a \emph{polynomial} algorithm from $\mat{Z} := \matseq{\mat{X}}{1} \ldots \matseq{\mat{X}}{J}$ (see \Cref{algo:butterfly-fact}).
	This algorithm is associated with our \emph{identifiability} results (see \Cref{thm:identifiability-multilayer-butterfly}), 
	which claims that the factorization $\mat{Z} := \matseq{\mat{X}}{1} \ldots \matseq{\mat{X}}{J}$ for any $\matseq{\mat{X}}{1}, \ldots, \matseq{\mat{X}}{J}$ satisfying the butterfly constraint is \emph{essentially unique} up to natural scaling ambiguities. 
	The algorithm is based on a hierarchical approach \cite{le2016flexible} in which the target matrix is iteratively factorized into two factors, 
	until the desired number of sparse factors $J$ is obtained. These successive two-layer factorizations rely on a non-trivial application of the singular value decomposition (SVD) to compute best rank-one approximations of specific submatrices, instead of iterative gradient descent steps \cite{papierTung}. The total complexity of the hierarchical algorithm is only $\mathcal{O}(N^2)$ where $N$ is the size of $\mat{Z}$. This is remarkably of the same order of magnitude as the complexity of matrix-vector multiplication, and needs to be performed only once to enable $\mathcal{O}(N \log N)$ matrix-vector multiplications with the resulting factored representation of $\mat{Z}$.
	In other words, our work shows that problem \eqref{eq:smf-general} with fixed butterfly constraints on the factors can be solved efficiently by our algorithm in the noiseless setting, i.e., under the assumption that the target matrix $\mat{Z}$ admits an exact butterfly factorization. 
	This naturally opens perspective for studying approximation guarantees in the noisy setting and in the quest for a proximal operator associated to the butterfly structure.

	\subsection{Related work}

	\paragraph{Butterfly structure} 
	The butterfly structure as defined in \Cref{def:butterfly-structure} below and illustrated in \Cref{fig:butterfly-support} is related to the divide-and-conquer structure of the fast Fourier transform. 
	It was used in earlier works to design fast randomization methods for preconditioning in computational linear algebra \cite{Parker95randombutterfly}. 
	It also offers an efficient parametrization of orthogonal matrices, which allows for fast approximation of Hessian matrices \cite{mathieu2014fast} and efficient trainable unitary matrices in recurrent neural networks 
	\cite{jing2017tunable}. In the context of kernel approximation via random features, the butterfly structure is used to generate random orthogonal matrices \cite{genz2000methods} in order to design quadrature rules for approximating the spherical part of the integral form of a kernel function \cite{munkhoeva2018quadrature, choromanski2019unifying}. 
	Recently, it has been shown that many common discrete operators associated to fast transforms can be approximated via gradient-descent methods by a data-sparse representation based on the butterfly structure \cite{dao2019learning, dao2020kaleidoscope}. 
	
	\paragraph{Data-sparse representation of structured matrices}
	Our work proposes an efficient hierarchical algorithm to recover the butterfly factors from their product, by computing successive SVDs on specific submatrices. This approach is similar to existing algorithms for approximating other types of structured matrices by a data-sparse representation that allows for fast matrix-vector multiplication. These structured matrices typically include: low-rank matrices \cite{woolfe2008fast, halko2011finding}, $\mathcal{H}$-matrices \cite{wolfgang1999sparse}, $\mathcal{H}^2$-matrices \cite{hackbusch2002data}, hierarchically semi-separable matrices \cite{martinsson2011fast}, and complementary low-rank matrices \cite{michielssen1996multilevel,o2010algorithm,candes2009fast,li2015butterfly}. 
	The  complementary low-rank structure, which is satisfied in many applications involving differential equations \cite{candes2009fast}, is the closest to the butterfly structure considered in this paper.
	Similarly to what we propose it is associated to decomposition algorithms, called {\em butterfly algorithms} \cite{michielssen1996multilevel,o2010algorithm,candes2009fast,li2015butterfly}, involving a recursive use of SVDs and certain types of binary trees.
	On the one hand, in our hierarchical factorization approach, the decomposition algorithm  (\Cref{algo:butterfly-fact}) is parameterized by a single but flexible ``factor-bracketing'' 
	binary tree somehow describing the bracketing of factors associated to their product. 
	On the other hand, butterfly algorithms \cite{li2015butterfly} are applicable to factorize matrices satisfying the complementary low-rank model, which is parameterized by two ``index-partitioning'' binary trees. These two trees are associated to certain hierarchical partitioning of row (resp.~column) indices.
	As further discussed in \cref{subsec:complementary-low-rank}, the butterfly algorithm \cite{li2015butterfly} (using a standard SVD rather than a randomized one) can be interpreted as a specific instance of our hierarchical factorization approach with a ``symmetric'' factor-bracketing tree, while our approach allows a much wider range of factor-bracketing trees. Vice-versa, we show that any matrix admitting a butterfly factorization in the sense of \Cref{def:butterfly-structure} satisfies the complementary low-rank model with a particular set of index-partitioning binary trees, so none of the approaches is more general than the other. Making the best of both is an interesting direction for future work with a potential to further accelerate our approach.
	
	\paragraph{Identifiability in multilinear inverse problems}
	Our paper also provides an analysis of identifiability, i.e., essential uniqueness, in butterfly sparse matrix factorization.
	Many identifiability results in multilinear inverse problems are derived from the \emph{lifting} procedure \cite{choudhary2014identifiability, choudhary2014sparse, malgouyres2016identifiability, li2017identifiability, malgouyres2019multilinear}. This procedure was originally used in the PhaseLift method \cite{li2013sparse, candes2013phaselift, candes2015phase} to address the phase retrieval problem.
	Other bilinear inverse problems, like blind-deconvolution \cite{ahmed2013blind, choudhary2014identifiability, bahmani2015lifting, li2016identifiability, li2016identifiability_bis, kech2017optimal, li2017identifiability} or self-calibration \cite{ling2015self}, have also been addressed using the lifting procedure. 
	A general framework to analyze identifiability for any bilinear inverse problem has been given in \cite{choudhary2014identifiability}. 
	Following the work from \cite[Chapter 7]{le2016matrices}, our analysis of identifiability in the case with $J=2$ factors is a specialization of this general lifting procedure to the matrix factorization problem \emph{with support constraints}. 
	Identifiability results for $J \geq 3$ are more challenging as the usual lifting procedure from bilinear inverse problems
	cannot be directly leveraged. This multi-layer setting requires extending this procedure using a so-called tensorial lifting \cite{malgouyres2016identifiability, malgouyres2019multilinear, malgouyres2020stable}. However, due to this multi-layer structure, conditions obtained via tensorial lifting might be difficult to verify in practice, although stable recovery conditions for convolutional linear networks have been derived \cite{malgouyres2019multilinear, malgouyres2020stable}. In contrast, our work proves identifiability results in matrix factorization with $J \geq 3$ factors \emph{without using tensorial lifting}, using a hierarchical approach that reduces the analysis with multiple factors to the case with only two factors.
	
	\paragraph{Empirical performance of our hierarchical algorithm}
	The hierarchical factorization algorithm (\Cref{algo:butterfly-fact}) has been studied empirically\footnote{These two contributions have been produced in parallel, and the conference publication \cite{le:hal-03438881} refers to the current submission/preprint to support its theoretical claims.} in 
	\cite{le:hal-03438881} where it was shown
	that the butterfly factors of the Hadamard and the DFT matrix can be recovered using \Cref{algo:butterfly-fact}, with reduced computational time and increased accuracy compared to gradient-based methods like \cite{dao2019learning}. As a contribution, our work on identifiability gives theoretical foundations for \Cref{algo:butterfly-fact}, since \Cref{thm:identifiability-multilayer-butterfly} shows exact recovery guarantees of the algorithm. We also justify the observed speed up of the factorization algorithm compared to gradient-based optimization methods, by showing that \Cref{algo:butterfly-fact} has a controlled time complexity of $\mathcal{O}(N^2)$.

	\subsection{Summary} The main contributions of this paper are the following ones:
	\begin{enumerate}
		\item We show in \Cref{thm:identifiability-multilayer-butterfly} that enforcing the butterfly structure on the $J$ sparse factors is sufficient to ensure a hierarchical identifiability property, meaning that we can recover (up to natural scaling ambiguities) sparse factors $(\matseq{\mat{X}}{\ell})_{\ell=1}^J$ from $\mat{Z} := \matseq{\mat{X}}{1} \ldots \matseq{\mat{X}}{J}$, using the hierarchical method described in \Cref{algo:butterfly-fact}.
		\item We show that this algorithm has a time complexity of only $\mathcal{O}(N^2)$ where $N$ is the size of $\mat{Z}$, which is of the order of a few dense matrix-vector multiplications, while the obtained factorizations enable fast $\mathcal{O}(N \log N)$ matrix-vector multiplications.
		\item We illustrate this complexity by implementing\footnote{Implementation available in the FAµST 3.25 toolbox (\url{https://faust.inria.fr/}).} \Cref{algo:butterfly-fact} using  \emph{truncated} SVDs. 
	\end{enumerate}

The paper is organized as follows: \cref{section:two-layer} analyzes identifiability in the two-layer setting with fixed-support constraint;
\cref{section:multilayer} describes how the butterfly structure can ensure the hierarchical identifiability of the sparse factors from their product;  \cref{section:experiments} presents some numerical experiments about the recovery of the sparse butterfly factors from their product; \cref{section:conclusion} discusses perspectives of this work. An annex gathers technical proofs.

\paragraph{Notations}
\label{section:notations}
The set of integers $\{1, \ldots, n\}$ is denoted $\integerSet{n}$.
The \emph{support} of a matrix $\mat{M} \in \mathbb{C}^{m \times n}$ of size $m \times n$ is the set of indices $\supp(\mat{M}) \subseteq \integerSet{m} \times \integerSet{n}$ of its nonzero entries. Depending on the context, such a matrix support can be seen either as a set of indices, or as a binary matrix (belonging to $\mathbb{B}^{m \times n} := \{ 0, 1\}^{m \times n}$) 	with only nonzero entries for indices in this set. 
The \emph{column support}, denoted $\colsupp({\mat{M}})$, is the subset of indices $i \in \integerSet{n}$ such that the $i$-th column of $\mat{M}$, denoted $\matcol{\mat{M}}{i}$, is nonzero. 
The entry of $\mat{M}$ indexed by $(k, l)$ is $\matindex{\mat{M}}{k}{l}$. 
The notation $\matseq{\mat{M}}{i}$ denotes the $i$-th matrix in a collection.
Given a support $\mat{S} \in \mathbb{B}^{m \times n}$, the restriction of $\mat{M}$ to $\mat{S}$ is the matrix $\mat{M} \odot \mat{S} \in \mathbb{C}^{m \times n}$ defined by the entries: $\matindex{(\mat{M} \odot \mat{S})}{k}{l} = \matindex{\mat{M}}{k}{l}$ if $(k, l) \in \mat{S}$, and zero otherwise.
The identity matrix of size $N$ is denoted $\identity{N}$\footnote{It should be clear in the text when the subscript $N$ indicates the size $N$ of the matrix and not its $N$-th column.}. 
The Kronecker product \cite{Kronecker} between $\mat{A}$ and $\mat{B}$ is written $\mat{A} \otimes \mat{B}$.
We recall:
\begin{equation}
	\label{eq:kronecker}
	(\mat{A} \otimes \mat{C})(\mat{B} \otimes \mat{D}) = \mat{A} \mat{B} \otimes \mat{C} \mat{D}.
\end{equation}

\section{Identifiability in two-layer sparse matrix factorization}
	\label{section:two-layer}	
	In order to show hierarchical identifiability results, we first analyze identifiability in two-layer sparse matrix factorization. 	Given a matrix $\mat{Z} \in \mathbb{C}^{m \times n}$, and a subset of pairs of factors $\Sigma \subseteq \mathbb{C}^{m \times r} \times \mathbb{C}^{n \times r}$, the so-called \emph{exact matrix factorization} problem with two factors of $\mat{Z}$ in $\Sigma$ is:
	\begin{equation}
		\label{eq:EMF}
		\text{find if possible } (\mat{X}, \mat{Y}) \in \Sigma \text{ such that } \mat{Z} = \mat{X} \transpose{\mat{Y}}.
	\end{equation}
	Uniqueness properties are studied in the exact setting, hence for the rest of the paper, we assume that $\mat{Z}$ admits such an exact factorization. 
	
	We are interested in the particular problem variation where the constraint set $\Sigma$ encodes some chosen sparsity patterns for the factorization. 
	For a given binary matrix $\mat{S} \in \mathbb{B}^{m \times r}$ associated to a sparsity pattern, denote 
	\begin{equation}
		\label{eq:model-set}
		\Sigma_{\mat{S}} := \{ \mat{M} \in \mathbb{C}^{m \times r}\; | \; \supp(\mat{M}) \subseteq \supp(\mat{S}) \},
	\end{equation}
	which is the set of matrices with a support 
	included in $\mat{S}$. A pair of sparsity patterns is written $\pair{S} := (\leftsupp{S}, \rightsupp{S})$, where $\leftsupp{S}$ and $\rightsupp{S}$ are the left and right sparsity patterns respectively, also referred to as a left and a right (allowed) support.
	Given any pair of allowed supports represented by binary matrices $\pair{S} := (\leftsupp{S}, \rightsupp{S}) \in \mathbb{B}^{m \times r} \times \mathbb{B}^{n \times r}$, the set
	\begin{equation}
		\label{eq:model-set-pair}
		\Sigma_{\pair{S}} := \Sigma_{\leftsupp{S}} \times \Sigma_{\rightsupp{S}} \subseteq \mathbb{C}^{m \times r} \times \mathbb{C}^{n \times r}
	\end{equation}
	is a linear subspace.
	Since the support of a matrix is unchanged under arbitrary rescaling of its columns, $\Sigma_{\pair{S}}$ is invariant by column scaling for any pair of supports $\pair{S}$. 
	Uniqueness of a solution to \eqref{eq:EMF} with such sparsity constraints will always be considered up to unavoidable scaling ambiguities. Using the terminology from the tensor decomposition literature \cite{kruskal1977three}, such a uniqueness property will be referred to as \emph{essential uniqueness}.
	
	\begin{definition}[Essential uniqueness of a two-layer factorization in $\Sigma$]
		\label{def:essential-uniqueness-two}
		Let $\Sigma$ be a set of pairs of factors, and $\mat{Z}$ be a matrix admitting a factorization $\mat{Z} := \mat{X} \transpose{\mat{Y}}$ such that $(\mat{X}, \mat{Y}) \in \Sigma$. This factorization is \emph{essentially unique} in $\Sigma$, if any solution $(\bar{\mat{X}}, \bar{\mat{Y}})$ to \eqref{eq:EMF} with $\mat{Z}$ and $\Sigma$ is equivalent to $(\mat{X}, \mat{Y})$, written $(\bar{\mat{X}}, \bar{\mat{Y}}) \sim (\mat{X}, \mat{Y})$, in the sense that there is an invertible diagonal matrix $\mat{D}$ such that $(\bar{\mat{X}}, \bar{\mat{Y}}) = \left(\mat{X} \mat{D}, \mat{Y} \mat{D}^{-1} \right)$.
	\end{definition}

	\begin{remark}
		\label{rmk:uniqueness-perm?}
		Another
		definition of essential uniqueness \cite{kruskal1977three} 
		involves permutation ambiguity in addition to scaling ambiguity.
		Nevertheless, we only consider scaling ambiguity in our definition of essential uniqueness, because all the fixed-support constraints $\Sigma_\pair{S}$ considered in this paper remove this permutation ambiguity, as detailed in \Cref{rmk:remove-perm-ambiguity} below.
	\end{remark}

	For any set $\Sigma$ of pairs of factors, the set of all pairs $(\mat{X}, \mat{Y}) \in \Sigma$ such that the factorization $\mat{Z} := \mat{X} \transpose{\mat{Y}}$ is essentially unique in $\Sigma$ is denoted $\unique(\Sigma)$. In other words, we define:
	\begin{equation}
		\label{eq:unique}
		\unique(\Sigma) := \left \{ (\mat{X}, \mat{Y}) \in \Sigma \; | \; \forall (\bar{\mat{X}}, \bar{\mat{Y}}) \in \Sigma, \; \bar{\mat{X}} \transpose{\bar{\mat{Y}}} = \mat{X} \transpose{\mat{Y}} \implies (\bar{\mat{X}}, \bar{\mat{Y}}) \sim (\mat{X}, \mat{Y})  \right \}.
	\end{equation}
	
	To characterize $\unique(\Sigma_{\pair{S}})$ for $\Sigma_{\pair{S}}$ defined as in \eqref{eq:model-set-pair} with $\pair{S}$ any pair of supports, we first establish a non-degeneration property, i.e., a necessary condition for identifiability, involving the so-called \emph{column support}. Define the set of pairs of factors with \emph{identical}, resp.~\emph{maximal},
	column supports in $\Sigma_{\pair{S}}$ as  
	\begin{eqnarray}
		\qquad \idcolsupp{\pair{S}} 
		&:= 
		& \{ (\mat{X}, \mat{Y}) \in \Sigma_{\pair{S}} \; | \; \colsupp (\mat{X}) = \colsupp (\mat{Y}) \}, 	
		\label{eq:DefIC} \\
		\qquad \maxcolsupp{\pair{S}} 
		&:= 
		& \{ (\mat{X}, \mat{Y}) \in \Sigma_{\pair{S}} \; | \;   \colsupp(\mat{X}) = \colsupp(\leftsupp{S}) \text{ and } \colsupp(\mat{Y}) = \colsupp(\rightsupp{S})\}. 	\label{eq:DefMC}
	\end{eqnarray}

	\begin{lemma}
		\label{lemma:identical-colsupp-maximal-colsupp}
		For any pair of supports $\pair{S}$, we have: $\unique(\Sigma_{\pair{S}}) \subseteq \idcolsupp{\pair{S}} \cap \maxcolsupp{\pair{S}}$.
	\end{lemma}

	\begin{remark}
		\label{rmk:equal-supp-colsupp}
		Consequently, if $\colsupp(\leftsupp{S}) \neq \colsupp(\rightsupp{S})$, then the set $\unique(\Sigma_{\pair{S}})$ is empty.
	\end{remark}

	In other words, if the factorization $\mat{Z} := \mat{X} \transpose{\mat{Y}}$ is essentially unique in $\Sigma_{\pair{S}}$, then the left and right supports have necessarily the same column support, and $\mat{X}$, $\mat{Y}$ do not have a zero column inside this column support. The proof is deferred to \Cref{app:identical-colsupp-maximal-colsupp}.

	As the product $\mat{X} \transpose{\mat{Y}}$ is the sum of rank-one matrices $\sum_{i=1}^r \matcol{\mat{X}}{i} \transpose{\matcol{\mat{Y}}{i}}$, the lifting procedure \cite{choudhary2014identifiability, malgouyres2016identifiability} suggests to represent the pair $(\mat{X}, \mat{Y})$ by its $r$-tuple of so-called \emph{rank-one contributions} 
	\begin{equation}\label{eq:DefRankOneMapping}
		\varphi(\mat{X}, \mat{Y}) := (\matcol{\mat{X}}{i} \transpose{\matcol{\mat{Y}}{i}})_{i=1}^r \in (\mathbb{C}^{m \times n})^{r}.
	\end{equation} 
	Indeed, one can identify, up to scaling ambiguities, the columns $\matcol{\mat{X}}{i}$, $\matcol{\mat{Y}}{i}$ from their outer product $\rankone{C}{i} := \matcol{\mat{X}}{i} \transpose{\matcol{\mat{Y}}{i}}$ ($1 \leq i \leq r$), \emph{as long as} the rank-one contribution $\rankone{C}{i}$ is not zero.
	\begin{lemma}[Reformulation of {\cite[Chapter 7, Lemma 1]{le2016matrices}}]
		\label{lemma:id-from-outerproduct}
		Consider $\rankone{C}{}$ the outer product of two vectors $\vctor{a}$, $\vctor{b}$. If $\rankone{C}{} = \vctor{0}$, then $\vctor{a} = \vctor{0}$ or $\vctor{b} = \vctor{0}$. If $\rankone{C}{} \neq \vctor{0}$, then $\vctor{a}$, $\vctor{b}$ are nonzero, and for any $(\vctor{a'}, \vctor{b'})$ such that $\vctor{a'} \transpose{\vctor{b'}} = \rankone{C}{}$, there exists a scalar $\lambda \neq 0$ such that $\vctor{a'} = \lambda \vctor{a}$ and $\vctor{b'} = \frac{1}{\lambda} \vctor{b}$.
	\end{lemma}
	
	With this lifting approach, each support constraint $\pair{S} = (\leftsupp{S}, \rightsupp{S})$ is represented by the $r$-tuple of rank-one support constraints $\tuplerkone{S} = \varphi(\leftsupp{S}, \rightsupp{S}) = (\rankone{S}{i})_{i=1}^{r}$. 
	Thus, if $(\mat{X}, \mat{Y}) \in \Sigma_{\pair{S}} \subseteq \mathbb{C}^{m \times r} \times \mathbb{C}^{n \times r}$, then the $r$-tuple of rank-one matrices $\varphi(\mat{X}, \mat{Y}) \in (\mathbb{C}^{m \times n})^{r}$ belongs to the set:
	\begin{equation}
		\label{eq:gamma-s}
		\Gamma_{\tuplerkone{S}} := \left \{ (\rankone{C}{i})_{i=1}^r \; | \; \forall i \in \integerSet{r}, \;\rank(\rankone{C}{i}) \leq 1, \, \supp(\rankone{C}{i}) \subseteq \rankone{S}{i} \right \} \subseteq (\mathbb{C}^{m \times n})^{r}.
	\end{equation}
	
	As explained in the general framework of \cite{choudhary2014identifiability} established to analyze identifiability in bilinear inverse problems, the main advantage of this approach is to remove the inherent scaling ambiguity, while preserving a one-to-one correspondence between a pair of factors $(\mat{X}, \mat{Y})$ and its rank-one contributions representation $\varphi(\mat{X}, \mat{Y})$. Our work specializes this general framework to matrix factorization problem with two factors and support constraints. Details about the lifting procedure for fixed-support two-layer matrix factorization are in \Cref{app:lifting}.
	
	From this lifting procedure, we show that it is possible to derive a simple necessary and sufficient condition for identifiability.
	Despite its simplicity, this condition is the key to derive identifiability results in butterfly sparse matrix factorization using a hierarchical approach,  
	see \cref{section:butterfly-factorization} and \Cref{lemma:butterfly-supp-disjoint}. 
	The proof is deferred to \Cref{app:simple-sc-identifiability}.

	\begin{proposition}
		\label{prop:disjoint-rank-one-supports-unique-EMF}
		Assuming $\colsupp(\leftsupp{S}) = \colsupp(\rightsupp{S})$ (which is natural by \Cref{rmk:equal-supp-colsupp}), $\unique(\Sigma_{\pair{S}}) = \idcolsupp{\pair{S}} \cap \maxcolsupp{\pair{S}}$ if, and only if, the tuple  $\varphi(\pair{S})= (\rankone{S}{i})_{i=1}^{r}$ has disjoint rank-one supports, i.e., $\supp(\rankone{S}{i}) \cap \supp(\rankone{S}{j}) = \emptyset$ for every $i \neq j$. 
	\end{proposition}

 More conditions on fixed-support identifiability are given in \cite{le2016matrices,LeonPart1}.

	\begin{remark}
		\label{rmk:remove-perm-ambiguity}
		Following the discussion of \Cref{rmk:uniqueness-perm?}, the assumption 
		that $\varphi(\pair{S})$ has disjoint rank-one supports
		is sufficient to remove any potential permutation ambiguity in the factorization of $\mat{Z} := \mat{X} \transpose{\mat{Y}}$ in $\Sigma$ for any $(\mat{X}, \mat{Y}) \in \idcolsupp{\pair{S}} \cap \maxcolsupp{\pair{S}}$, in the sense that: for any $(\mat{X}, \mat{Y}) \in \idcolsupp{\pair{S}} \cap \maxcolsupp{\pair{S}}$, any invertible diagonal matrix $\mat{D}$ and any permutation matrix $\mat{P}$ such that $(\mat{X} \mat{D} \mat{P}, \mat{Y} \inverse{\mat{D}} \mat{P}) \in \Sigma_{\pair{S}}$, the permutation matrix $\mat{P}$ is necessarily the identity matrix. This is true because otherwise, denoting $(\rankone{S}{i})_{i=1}^r := \varphi(\pair{S})$, $(\rankone{C}{i})_{i=1}^r := \varphi(\mat{X}, \mat{Y})$ and noticing that $\varphi(\mat{X}\mat{D} \mat{P}, \mat{Y} \inverse{\mat{D}} \mat{P})$ is equal to $(\rankone{C}{i})_{i=1}^r$ up to a permutation on the index $i$, there would exist two indices $j \neq k$ such that $\supp(\rankone{C}{j}) \subseteq \supp(\rankone{S}{j})$ and $\supp(\rankone{C}{j}) \subseteq \supp(\rankone{S}{k})$ with $\supp(\rankone{C}{j}) \neq \emptyset$ (because $(\mat{X}, \mat{Y}) \in \idcolsupp{\pair{S}} \cap \maxcolsupp{\pair{S}}$), which would contradict the fact that $\supp(\rankone{S}{j}) \cap \supp(\rankone{S}{k}) = \emptyset$.
	\end{remark}
	
	\section{Hierarchical identifiability of butterfly factors}
	\label{section:multilayer}
	
	The main contribution of the paper is to establish some identifiability results in the multi-layer sparse matrix factorization when the sparse factors are constrained to have the so-called \emph{butterfly supports}.
	
	\subsection{Definition and properties of the butterfly structure}
	\label{section:butterfly-factorization}
	We now formally introduce the butterfly structure and its important properties used to establish identifiability results. 
	
	\begin{definition}[Butterfly supports]
		\label{def:butterfly-suppport}
		The \emph{butterfly supports} of size $N = 2^J$ are the $J$-tuple of supports $\bflytuple := ( \bflyindex{1}, \ldots,  \bflyindex{J} ) \in (\mathbb{B}^{N \times N})^J$ defined by:
		\begin{equation}
			\label{eq:butterfly-supports}
			\bflyindex{\ell} := \identity{2^{\ell - 1}} \otimes \begin{pmatrix}  1 & 1\\ 1 &1 \end{pmatrix} \otimes \identity{N/2^\ell}, \; 1 \leq \ell \leq J.
		\end{equation}
	\end{definition}
	
	\Cref{fig:butterfly-support} is an illustration of the butterfly supports. They are 2-regular \cite{le2021structured}, i.e., they have at most 2 nonzero entries per row and per column, and they are also block-diagonal, 
	including the leftmost factor if we consider a single block that covers the matrix entirely.
	
	\begin{definition}[Butterfly structure]
		\label{def:butterfly-structure}
		We say that a matrix $\mat{Z}$ of size $N = 2^J$ admits a butterfly structure if it can be factorized into $J$ factors $( \matseq{\mat{X}}{1}, \ldots, \matseq{\mat{X}}{J})$ that have a support included in the butterfly supports $\bflytuple := (  \bflyindex{1}, \ldots,  \bflyindex{J})$, in the sense that:
		\begin{equation*}
			\mat{Z} = \matseq{\mat{X}}{1} \ldots \matseq{\mat{X}}{J}, \quad \text{with} \quad ( \matseq{\mat{X}}{1}, \ldots, \matseq{\mat{X}}{J}) \in \Sigma_{\bflyindex{1}} \times \ldots \times \Sigma_{\bflyindex{J}}.
		\end{equation*}
		For the rest of the paper we write by slight abuse of notation:
			\begin{equation}
				\Sigma_{\bflytuple} := \Sigma_{\bflyindex{1}} \times \ldots \times \Sigma_{\bflyindex{J}}.
		\end{equation}
	\end{definition}

	With such a structure, matrix-vector multiplication has complexity $\mathcal{O}(N \log N)$ \cite{dao2019learning}: 
	there are $J = \log_2(N)$ butterfly factors 
	and the complexity of matrix-vector multiplication by each butterfly factor $\matseq{\mat{X}}{\ell}$ ($\ell \in \integerSet{J}$) is 
	$\mathcal{O}(N)$, since $\matseq{\mat{X}}{\ell}$ has at most $2N$ nonzero entries.
	As explained in the introduction, the butterfly structure is involved in many matrices associated to fast linear transforms \cite{dao2019learning}, such as the discrete Fourier transform (DFT) matrix.
	
	\begin{example}[Butterfly factorization of the DFT matrix \cite{dao2019learning}]
		\label{ex:butterfly-dft}
		Consider the DFT matrix of size $N \times N$ with $N = 2^J$ denoted as $\dft{N}$, defined by: 
		\begin{equation}
			\label{eq:DFT-matrix}
			\dft{N} := ({\omega_N}^{(k-1)(l-1)})_{k, l \in \integerSet{N}}, \quad \text{where } \omega_N := e^{- i \frac{2\pi}{N}}.
		\end{equation}
		The butterfly factorization of $\dft{N}$ \cite{dao2019learning} relies on the recursive relation
		\begin{equation}
			\label{eq:fourier-recursive}
			\dft{N} = \matsize{B}{N} \begin{pmatrix}
				\dft{N/2} & 0 \\ 0 & \dft{N/2}
			\end{pmatrix} \matsize{P}{N},
		\end{equation}
		where $\matsize{P}{N} \in \mathbb{B}^{N \times N}$ is the permutation matrix that sorts the odd then the even indices\footnote{For instance, for $N=4$, it permutes $(1, 2, 3, 4)$ to $(1, 3, 2, 4)$.},
		and 
		\begin{equation}
			\label{eq:dft-butterfly-block}
			\matsize{B}{N} := \begin{pmatrix}
				\identity{N/2} & \matsize{A}{N/2} \\ \identity{N/2} & - \matsize{A}{N/2}
			\end{pmatrix},
		\end{equation}
		with $\matsize{A}{N/2}$ the diagonal matrix with diagonal entries $1, \omega_N, \omega_N^2, \ldots, {\omega_N}^{\frac{N}{2} - 1}$. 
		Applying recursively \eqref{eq:fourier-recursive} to the block $\dft{N/2}$, we obtain the butterfly factorization of $\dft{N}$:
		\begin{equation}
			\label{eq:butterfly-factorization-dft}
			\begin{split}
			\dft{N} = \matseq{\mat{F}}{1} \ldots \matseq{\mat{F}}{J} \matsize{R}{N}, \quad \text{where } \matseq{\mat{F}}{\ell} := \identity{2^{\ell-1}} \otimes \matsize{B}{N / 2^{\ell-1}}, \; \ell \in \integerSet{J},\\
			\text{and}\ \matsize{R}{N} :=  \matseq{\mat{Q}}{J} \matseq{\mat{Q}}{J-1} \ldots \matseq{\mat{Q}}{1},\quad \text{with}\  \matseq{\mat{Q}}{\ell} := \identity{2^{\ell-1}} \otimes \matsize{P}{N / 2^{\ell-1}}.
			\end{split}
		\end{equation}
		Matrix $\matsize{R}{N} :=  \matseq{\mat{Q}}{J} \matseq{\mat{Q}}{J-1} \ldots \matseq{\mat{Q}}{1}$ is the so-called \emph{bit-reversal} permutation matrix.
	\end{example}

\begin{figure}[!h]
		\centering
		\begin{subfigure}[t]{0.23\textwidth}
			\centering
			\includegraphics[width=0.9\textwidth]{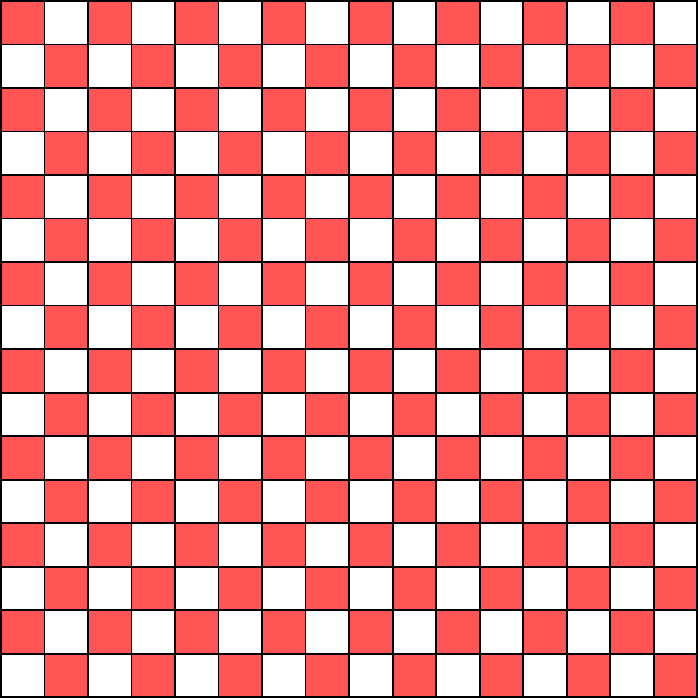}
			\caption{$\bflypartial{1}{3}$}
		\end{subfigure}%
		~ 
		\begin{subfigure}[t]{0.23\textwidth}
			\centering
			\includegraphics[width=0.9\textwidth]{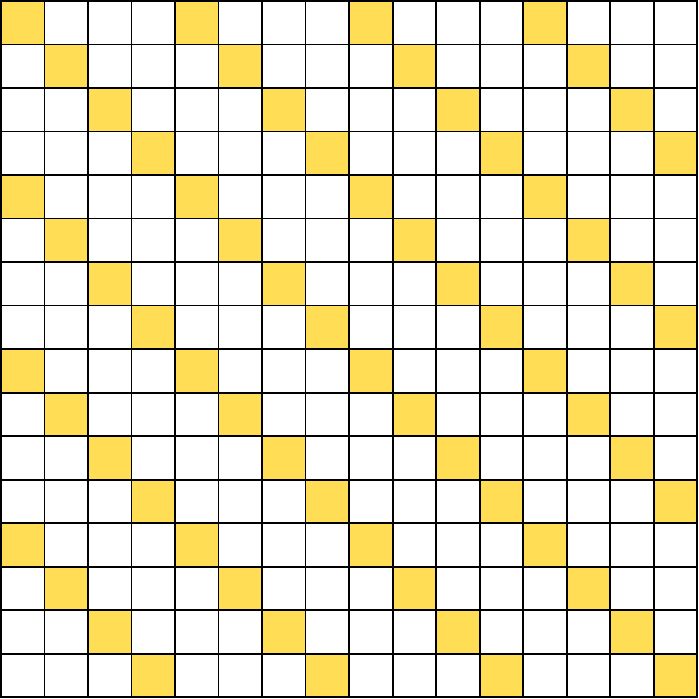}
			\caption{$\bflypartial{1}{2}$}
		\end{subfigure}%
		~ 
		\begin{subfigure}[t]{0.23\textwidth}
			\centering
			\includegraphics[width=0.9\textwidth]{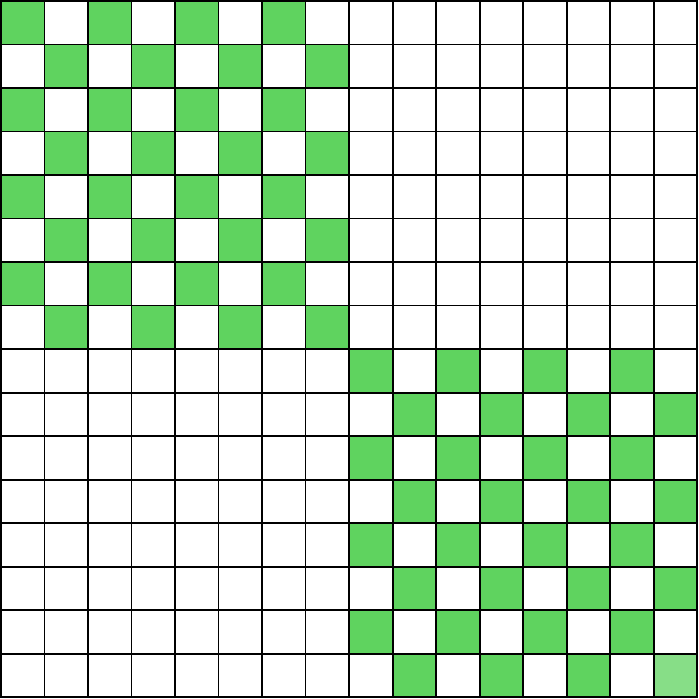}
			\caption{$\bflypartial{2}{3}$}
		\end{subfigure}%
		~ 
		\begin{subfigure}[t]{0.23\textwidth}
			\centering
			\includegraphics[width=0.9\textwidth]{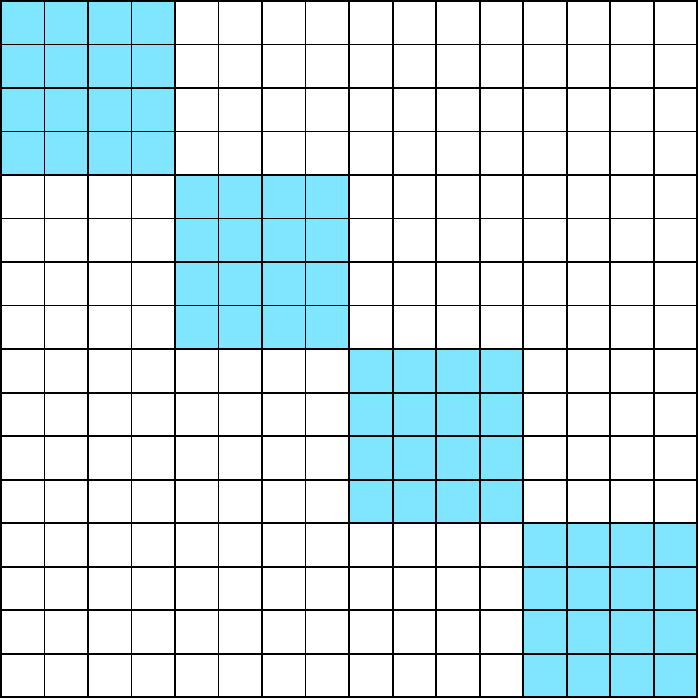}
			\caption{$\bflypartial{3}{4}$}
		\end{subfigure}%
		\caption{Examples of supports $\bflypartial{p}{q}$ defined by \eqref{eq:W_p_q} ($1 \leq p \leq q \leq 4$) of size $N \times N$ with $N = 16$. Nonzero entries are in color, and zero entries are in white.}
		\label{fig:structure-W}
	\end{figure}

	For any $(\matseq{\mat{X}}{1}, \ldots, \matseq{\mat{X}}{J}) \in \Sigma_{\bflytuple}$, the partial product of any consecutive factors $\matseq{\mat{X}}{p} \ldots \matseq{\mat{X}}{q}$ ($1 \leq p \leq q \leq J$) is shown to have a very precise structure encoded by the support $\bflypartial{p}{q}$ as detailed in the following lemma (proved in \Cref{app:block-structure-product-butterfly}). \Cref{fig:structure-W} illustrates this structure on some examples of size $N=16$. In particular this structure is involved in the hierarchical matrix factorization method, as explained below in \cref{section:hierarchical-method}.
	
	\begin{lemma}
		\label{lemma:block-structure-product-butterfly}
		Let $\bflytuple := (  \bflyindex{1}, \ldots,  \bflyindex{J})$ be the butterfly supports of size $N = 2^J$. Then, for any tuple $(\matseq{\mat{X}}{1}, \ldots, \matseq{\mat{X}}{J}) \in \Sigma_{\bflytuple}$, for any $1 \leq p \leq q \leq J$: $\supp(\matseq{\mat{X}}{p} \ldots \matseq{\mat{X}}{q}) \subseteq \bflypartial{p}{q}$, where
		\begin{align}
			\bflypartial{p}{q} 
			&:= \identity{2^{p-1}} \otimes \blockofidentity{p}{q} \in \mathbb{B}^{N \times N}, \; 
			\text{and} \label{eq:W_p_q}\\ 
			\blockofidentity{p}{q} &:= \matsize{U}{2^{q - p + 1}} \otimes \identity{N / 2^q}  \in \mathbb{B}^{\frac{N}{2^{p-1}} \times \frac{N}{2^{p-1}}}, \label{eq:V_p_q}
		\end{align}
		denoting, for any $n$, $\matsize{U}{n} \in \mathbb{B}^{n \times n}$ as the binary matrix full of ones. 
	\end{lemma}
	
	\begin{remark}
		\label{rmk:support-partial-prod}
		We have $\bflypartial{\ell}{\ell} = \bflyindex{\ell}$ for $1 \leq \ell \leq J$, and viewing matrix supports as binary matrices, one can verify that $\bflypartial{p}{q} = \bflyindex{p} \ldots \bflyindex{q}$ for $1 \leq p \leq q \leq J$.
		Also, by \eqref{eq:V_p_q}, $\blockofidentity{p}{q}$ is symmetric, i.e., $\blockofidentity{p}{q} = \transpose{(\blockofidentity{p}{q})}$. By \eqref{eq:W_p_q}, $\bflypartial{p}{q}$ is block-diagonal with blocks $\blockofidentity{p}{q}$, so $\bflypartial{p}{q}$ is also symmetric. Finally, $\bflypartial{p}{q}$ and $\blockofidentity{p}{q}$ are both $2^{q-p+1}$-sparse by column.
	\end{remark}

	\subsection{Hierarchical matrix factorization method}
	\label{section:hierarchical-method}
	The so-called \emph{butterfly sparse matrix factorization} problem is the following special instance of \eqref{eq:smf-general}:
	\begin{equation}
		\label{eq:butterfly-smf}
		\min_{\matseq{\mat{X}}{1}, \ldots, \matseq{\mat{X}}{J}} \| \mat{Z} - \matseq{\mat{X}}{1} \ldots \matseq{\mat{X}}{J} \|_F, \quad \text{such that }  ( \matseq{\mat{X}}{1}, \ldots, \matseq{\mat{X}}{J}) \in \Sigma_{\bflytuple}.
	\end{equation}
	
	\begin{remark}
		As shown in \cite[Remark A.1]{papierTung}, there exists a support constraint $\pair{S} = (\leftsupp{S}, \rightsupp{S})$ and a matrix $\mat{Z}$ such that: (a) $\mat{Z}$ \emph{cannot be written exactly} as $\mat{Z}=\mat{X} \transpose{\mat{Y}}$ for any $(\mat{X},\mat{Y}) \in \Sigma_{\pair{S}}$; (b) $\mat{Z}$ can be approximated arbitrarily well by such a product, i.e., $0 = \inf_{(\mat{X},\mat{Y}) \in \Sigma_{\pair{S}}} \|\mat{Z}-\mat{X} \transpose{\mat{Y}} \|_{F}$.
		This corresponds to a lack of closure of the set $\{\mat{X} \transpose{\mat{Y}}: (\mat{X},\mat{Y}) \in \Sigma_{\pair{S}}\}$. Fortunately this pathological behavior does not happen here since $\mat{Z}$ is assumed to admit an exact factorization. It is left for future work whether in the 
		case of butterfly supports, problem~\eqref{eq:butterfly-smf} always admits a minimizer even when $\mat{Z}$ does not admit an exact 
		butterfly 
		factorization.	
	\end{remark}
	
	As proposed in \cite{le2016flexible}, instead of directly optimizing over the $J$ factors, the hierarchical matrix factorization method is a heuristic approach that performs successive two-layer matrix factorizations, 
	until $J$ sparse factors are obtained. Let us illustrate this method in the scenario where we want to recover a tuple of sparse factors $( \matseq{\mat{X}}{1}, \ldots, \matseq{\mat{X}}{J}) \in \Sigma_{\bflytuple}$ from $\mat{Z} := \matseq{\mat{X}}{1} \ldots \matseq{\mat{X}}{J}$.
	 At the first level, the factors $\matseq{\mat{X}}{1}$ and $\partialproduct{\mat{X}}{2}{J} := \matseq{\mat{X}}{2} \ldots \matseq{\mat{X}}{J}$ are recovered from their known product $\matseq{\mat{X}}{1} \partialproduct{\mat{X}}{2}{J}=\mat{Z}$. At the second level, the factors $\matseq{\mat{X}}{2}$ and $\partialproduct{\mat{X}}{3}{J} := \matseq{\mat{X}}{3} \ldots \matseq{\mat{X}}{J}$ are in turn recovered from their known product $\matseq{\mat{X}}{2} \partialproduct{\mat{X}}{3}{J} = \partialproduct{\mat{X}}{2}{J}$. The process is repeated recursively, until all the sparse factors are recovered. 
	At each level, adequate support constraints are enforced on the factors: they correspond to sparsity patterns obtained from the partial products of several sparse factors. In our case with the butterfly constraint, these adequate support constraints correspond to $\bflypartial{p}{q}$ defined at \eqref{eq:W_p_q}.
	
	It was shown \cite{le:hal-03438881} that \Cref{algo:butterfly-fact}, which is based on the hierarchical approach, performs empirically better than gradient-based optimization methods \cite{dao2019learning} to address problem \eqref{eq:butterfly-smf}.
	Before discussing the details of \Cref{algo:butterfly-fact}, note that in the specific case of the butterfly support constraint,
	the hierarchical factorization method can be generalized to other factorization orders: for instance, in the case of four factors $\matseq{\mat{X}}{1}, \ldots, \matseq{\mat{X}}{4}$ of size $16 \times 16$, one can instead factorize $\mat{Z} := \matseq{\mat{X}}{1} \matseq{\mat{X}}{2} \matseq{\mat{X}}{3} \matseq{\mat{X}}{4}$ into $\matseq{\mat{X}}{1}\matseq{\mat{X}}{2}$ 
	and $\matseq{\mat{X}}{3} \matseq{\mat{X}}{4}$ at the first level, then $\matseq{\mat{X}}{1} \matseq{\mat{X}}{2}$ into $\matseq{\mat{X}}{1}, \matseq{\mat{X}}{2}$, and finally $\matseq{\mat{X}}{3} \matseq{\mat{X}}{4}$ into $\matseq{\mat{X}}{3}, \matseq{\mat{X}}{4}$. Let us formally introduce a tree structure that describes the factorization order in the hierarchical method.
	
	\begin{definition}[Factor-bracketing binary tree]
		\label{def:partitioning-bin-tree}
		A \emph{factor-bracketing binary tree} of a set of consecutive integers
		$\{p,\ldots,q\}$, with $1 \leq p \leq q$,
		is a binary tree, where nodes are non-empty subsets of $\{p,\ldots,q\}$,
		that satisfies the axioms:
		(a) each node is a subset of consecutive indices in $\{ p, \ldots, q \}$;
		(b) the root is the set $\{p,\ldots,q\}$;
		(c) a node is a singleton if, and only if, it is a leaf;
		(d) for each non-leaf node, the left and right children form a partition of their parent, in such a way that the indices of the left child are \emph{smaller} than those in the right child.
	\end{definition}
	
\begin{figure}[!h]
		\centering
		\begin{subfigure}[t]{0.24\textwidth}
			\centering
			\includegraphics[width=\textwidth]{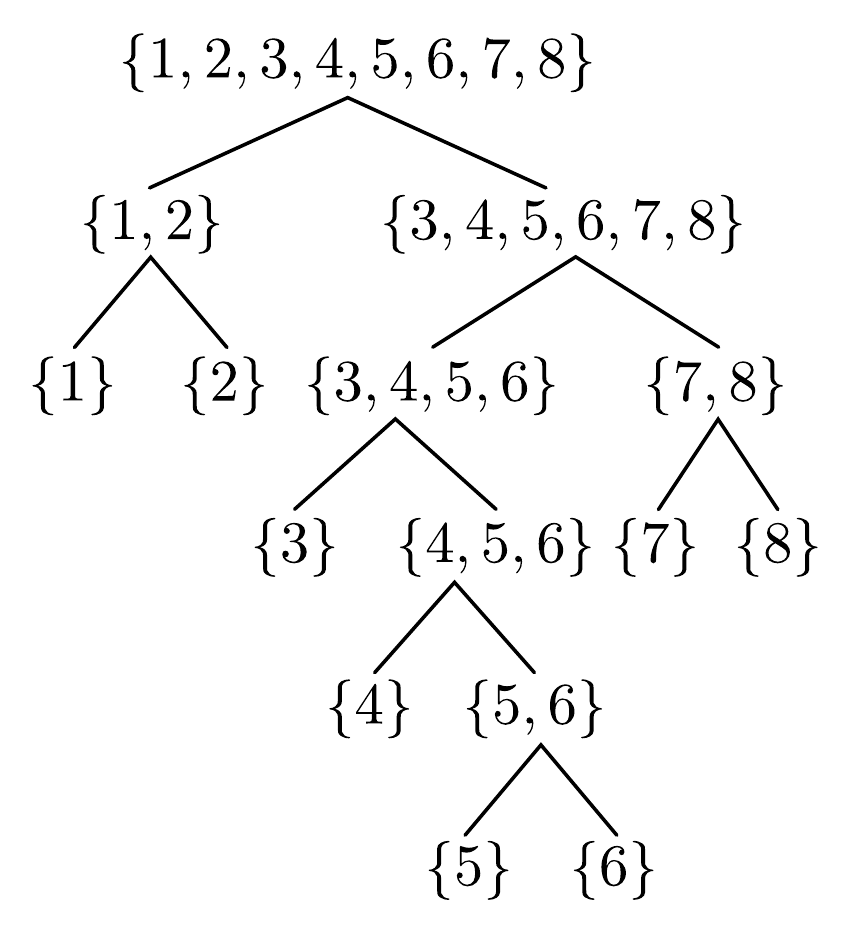}
			\caption{Arbitrary}
		\end{subfigure}%
		~ 
		\begin{subfigure}[t]{0.24\textwidth}
			\centering
			\includegraphics[width=\textwidth]{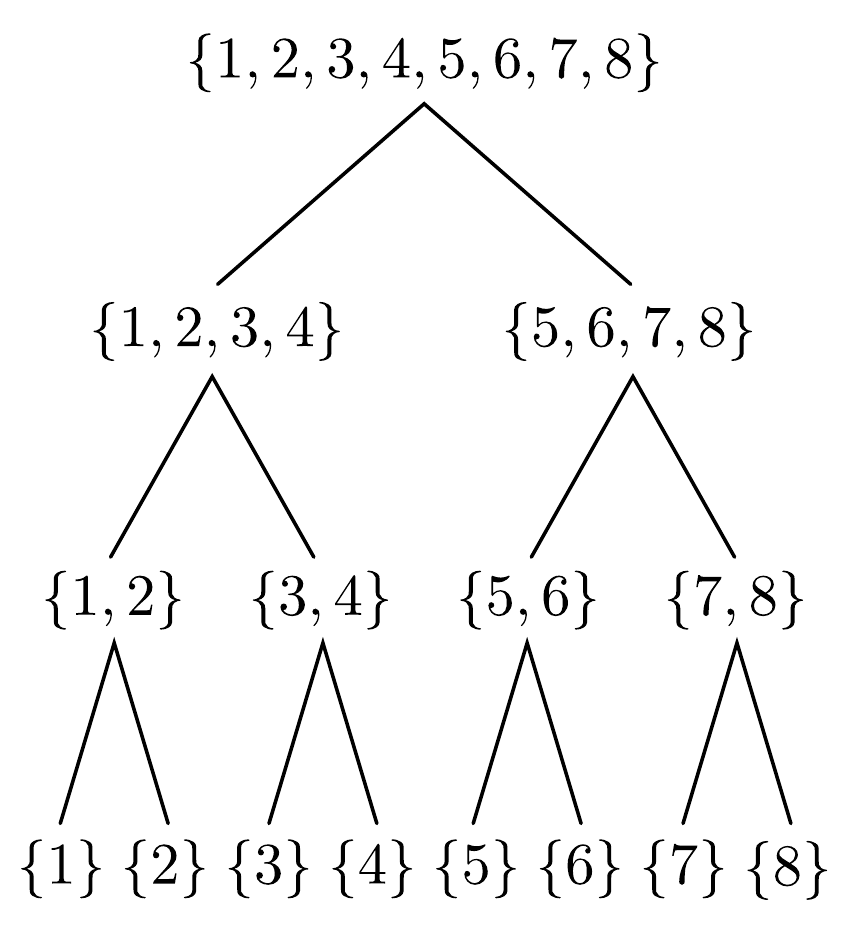}
			\caption{Balanced}
			\label{fig:balanced-tree}
		\end{subfigure}%
		~ 
		\begin{subfigure}[t]{0.24\textwidth}
			\centering
			\includegraphics[width=\textwidth]{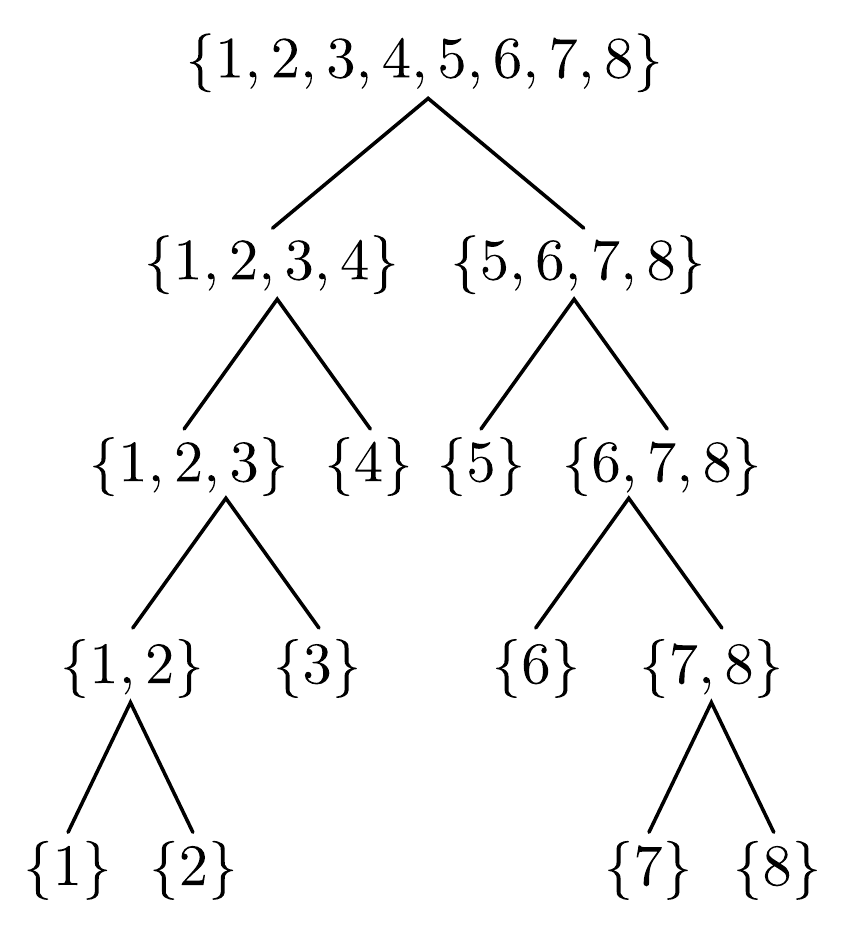}
			\caption{Symmetric}
			\label{fig:symmetric-tree}
		\end{subfigure}%
		~ 
		\begin{subfigure}[t]{0.24\textwidth}
			\centering
			\includegraphics[width=\textwidth]{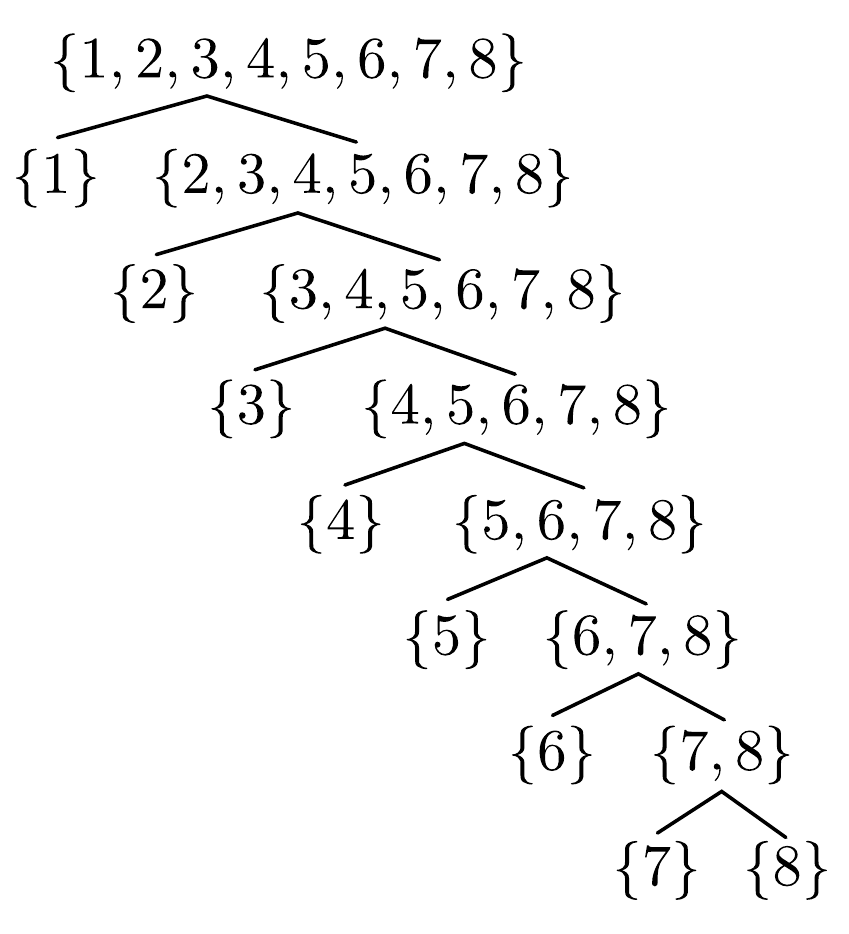}
			\caption{Unbalanced}
			\label{fig:unbalanced-tree}
		\end{subfigure}%
		\caption{Four possible factor-bracketing binary trees of $[J]$, $J=8$, that can serve as input to \Cref{algo:butterfly-fact}. NB: the algorithm is applicable to any number of factors $J$ (not only $J=8$).}
		\label{fig:factor-bracketing-tree}
	\end{figure}

	Examples of factor-bracketing binary trees are illustrated in \Cref{fig:factor-bracketing-tree}. 
	
	Given as inputs a factor-bracketing binary tree $\mathcal{T}$ of $\integerSet{J}$ and any target matrix $\mat{Z}$, \Cref{algo:butterfly-fact} visits the nodes of $\mathcal{T}$ in a breadth-first search order, starting by the root node. 
	At each non-leaf node $n := \{p, \ldots, q \} \subseteq \integerSet{J}$ characterized 
	its ``splitting index" $\ell$, which is the maximum value of its left child, \Cref{algo:hierarchical-step} approximates an intermediate matrix $\partialproduct{\mat{M}}{p}{q}$ by a product $\partialproduct{\mat{M}}{p}{\ell} \partialproduct{\mat{M}}{\ell+1}{q}$, where the left and right factor satisfy the support constraints encoded by  $\bflypartial{p}{\ell}$ and $\bflypartial{\ell+1}{q}$ (line \ref{line:apply-fsmf} of \Cref{algo:hierarchical-step}). The procedure used to compute these two factors is described in \Cref{algo:fsmf}, and is proved to be optimal in \cite{papierTung} in the sense that $\|\partialproduct{\mat{M}}{p}{q}-\mat{X}\transpose{\mat{Y}}\|_{F}^{2}$ is minimized among all $\mat{X},\mat{Y}$ satisfying the support constraints. 
	In essence, denoting $(\rankone{S}{i})_{i=1}^N := \varphi(\bflypartial{p}{\ell}, \transpose{\bflypartial{\ell+1}{q}})$, the procedure consists of successive best rank-one approximations (in the Frobenius norm) of submatrices $\partialproduct{\mat{M}}{p}{q} \odot \rankone{S}{i}$ for each $i \in \integerSet{N}$ (line \ref{line:best-rank1} of \Cref{algo:fsmf}), which can be computed for instance via a truncated SVD. 
	The hierarchical procedure is then repeated recursively on 
	$\partialproduct{\mat{M}}{p}{\ell}$ and $\partialproduct{\mat{M}}{\ell+1}{q}$, with their respective trees.

	\begin{remark}
	\label{remark:algo-icassp}
	One can exploit \Cref{algo:butterfly-fact} \emph{beyond the exact setting} to approximate \emph{any} matrix $\mat{Z}$ of size $2^J$ by a matrix having the butterfly structure, 
	since the procedure in \Cref{algo:fsmf} for two-layer fixed-support matrix factorization is optimal \cite{papierTung}.
	But because this procedure is used in a recursive greedy fashion in \Cref{algo:butterfly-fact}, global optimality of the resulting multi-layer factorization is not necessarily guaranteed. Understanding the stability of the algorithm beyond exact recovery is an interesting challenge left to future work.
\end{remark}

\begin{algorithm}[h]
	\centering
	\caption{Hierarchical butterfly factorization method, size $N = 2^J$.} \label{algo:butterfly-fact}
	\begin{algorithmic}[1]
		\Procedure{Butterfly}{$\mat{Z} \in \mathbb{C}^{N \times N}$, $\mathcal{T}$ factor-bracketing binary tree of $\integerSet{J}$}
		\State \Return $\Call{Hierarchical}{\mat{Z}, \mathcal{T}}$ from \Cref{algo:hierarchical-step}
		\EndProcedure
	\end{algorithmic}
\end{algorithm}

\begin{algorithm}[h]
	\centering
	\caption{Hierarchical step for indices $1 \leq p \leq q \leq J$ in the butterfly factorization method of size $N = 2^J$.} \label{algo:hierarchical-step}
	\begin{algorithmic}[1]
		\Procedure{Hierarchical}{$\partialproduct{\mat{M}}{p}{q} \in \mathbb{C}^{N \times N}$, factor-bracketing tree $\tree{p}{q}$ of $\{p, \ldots, q\}$}
		\If{$p=q$}
			\Return $\partialproduct{\mat{M}}{p}{q}$
		\EndIf
		\State $\ell \gets$ maximum value in the left child of the root of $\tree{p}{q}$
		\State $\left( \tree{p}{\ell}, \tree{\ell+1}{q} \right) \gets$  left and right subtrees of the root of $\tree{q}{p}$
		\State $(\bflypartial{p}{\ell}, \bflypartial{\ell+1}{q}) \gets$ supports defined by \eqref{eq:W_p_q}
		\State $(\partialproduct{\mat{M}}{p}{\ell}, \transpose{\partialproduct{\mat{M}}{\ell+1}{q}}) \gets \Call{FSMF}{\partialproduct{\mat{M}}{p}{q}, \bflypartial{p}{\ell}, \transpose{\bflypartial{\ell+1}{q}}}$ from \Cref{algo:fsmf} \label{line:apply-fsmf} 
		\State $(\matseq{\bar{\mat{X}}}{p}, \ldots, \matseq{\bar{\mat{X}}}{\ell}) \gets \Call{Hierarchical}{\partialproduct{\mat{M}}{p}{\ell}, \tree{p}{\ell}}$
		\State $(\matseq{\bar{\mat{X}}}{\ell+1}, \ldots, \matseq{\bar{\mat{X}}}{q}) \gets \Call{Hierarchical}{\partialproduct{\mat{M}}{\ell+1}{q}, \tree{\ell+1}{q}}$
		\State \Return $(\matseq{\bar{\mat{X}}}{p}, \ldots, \matseq{\bar{\mat{X}}}{\ell}, \matseq{\bar{\mat{X}}}{\ell+1}, \ldots, \matseq{\bar{\mat{X}}}{q})$
		\EndProcedure
	\end{algorithmic}
\end{algorithm}

\begin{algorithm}[!h]
	\centering
	\caption{Fixed-support matrix factorization under assumptions of \Cref{prop:disjoint-rank-one-supports-unique-EMF} \cite[Algorithm 3.1]{papierTung}.} \label{algo:fsmf}
	\begin{algorithmic}[1]
		\Procedure{FSMF}{$\mat{Z} \in \mathbb{C}^{m \times n}, \leftsupp{S} \in \mathbb{B}^{m \times r}, \rightsupp{S} \in \mathbb{B}^{n \times r}$}
		\State $(\rankone{S}{i})_{i=1}^{N} \gets \varphi \left( \leftsupp{S}, \rightsupp{S} \right)$ as defined in \eqref{eq:DefRankOneMapping}
		\For{$i \in \integerSet{r}$}
			\State $(\vctor{x}_i, \vctor{y}_i) \gets \argmin \left \{ \| \mat{Z} \odot \rankone{S}{i} - \vctor{x}_i \transpose{\vctor{y}_i} \|_F, \; \text{such that} \; \supp(\vctor{x}_i \transpose{\vctor{y}_i}) \subseteq \rankone{S}{i} \right \}$ \label{line:best-rank1}
		\EndFor
		\State $\mat{X} \gets (\vctor{x}_1, \ldots, \vctor{x}_r) \in \mathbb{C}^{m \times r}$
		\State $\mat{Y} \gets (\vctor{y}_1, \ldots, \vctor{y}_r) \in \mathbb{C}^{n \times r}$
		\State \Return $(\mat{X}, \mat{Y})$
		\EndProcedure
	\end{algorithmic}
\end{algorithm}

\subsection{Uniqueness of the butterfly factorization}
	As the main contribution of the paper, we now show that the exact factorization $\mat{Z} = \matseq{\mat{X}}{1} \ldots \matseq{\mat{X}}{J}$ into $J$ factors constrained to the butterfly supports is essentially unique, and that these factors can be recovered by \Cref{algo:butterfly-fact}.
	Let us first generalize \Cref{def:essential-uniqueness-two} to the multi-layer case with $J \geq 3$.
	
	\begin{definition}[Essential uniqueness of a multi-layer factorization in $\Sigma$]
		\label{def:essential-uniquess-multi}
		Consider integers $ N_0,\dots,N_{J}$, a set $\Sigma \subseteq \mathbb{C}^{N_0 \times N_1} \times \ldots \times \mathbb{C}^{N_{J-1} \times N_J}$ of $J$-tuples of factors, and 
		a matrix $\mat{Z}$ admitting a factorization 
		$\mat{Z} := \matseq{\mat{X}}{1} \ldots \matseq{\mat{X}}{J}$ such that $(\matseq{\mat{X}}{\ell})_{\ell=1}^J \in \Sigma$. We say that this factorization is \emph{essentially unique} in $\Sigma$, if any $(\matseq{\bar{\mat{X}}}{1}, \ldots, \matseq{\bar{\mat{X}}}{J}) \in \Sigma$ such that $\matseq{\bar{\mat{X}}}{1} \ldots \matseq{\bar{\mat{X}}}{J} = \mat{Z}$ is equivalent to $(\matseq{\mat{X}}{1}, \ldots, \matseq{\mat{X}}{J})$, written $(\matseq{\mat{X}}{\ell})_{\ell=1}^J \sim (\matseq{\bar{\mat{X}}}{\ell})_{\ell=1}^J$, in the sense that  there exist invertible diagonal matrices $\matseq{\mat{D}}{1}, \ldots, \matseq{\mat{D}}{J-1}$ 
		such that $\matseq{\bar{\mat{X}}}{\ell} = \inverse{\matseq{\mat{D}}{\ell-1}} \matseq{\mat{X}}{\ell} \matseq{\mat{D}}{\ell}$ for all $\ell \in \integerSet{J}$, with the convention that $\matseq{\mat{D}}{0}$ and $\matseq{\mat{D}}{J}$ are identity matrices.
	\end{definition}

	\begin{theorem}
		\label{thm:identifiability-multilayer-butterfly}
		Consider $\bflytuple$ the butterfly supports of size $N = 2^J$ and $(\matseq{\mat{X}}{1}, \ldots, \matseq{\mat{X}}{J}) \in \Sigma_{\bflytuple}$. Assume that $\matseq{\mat{X}}{\ell}$ does not have a zero column for $1 \leq \ell \leq J-1$, and not a zero row for $2 \leq \ell \leq J$.
		Then, the factorization $\mat{Z} := \matseq{\mat{X}}{1} \ldots \matseq{\mat{X}}{J}$ is essentially unique in $\Sigma_{\bflytuple}$. 		
		These factors can be recovered from $\mat{Z}$, up to scaling ambiguities only, using the procedure $\Call{Butterfly}{\mat{Z}, \mathcal{T}}$ detailed in \Cref{algo:butterfly-fact}, where  $\mathcal{T}$ is \emph{any} factor-bracketing tree of $\integerSet{J}$.
	\end{theorem}
	
	In other words, \Cref{algo:butterfly-fact} is endowed with \emph{exact recovery guarantees}.  
	In particular, \Cref{thm:identifiability-multilayer-butterfly} can be applied to show identifiability of the butterfly factorization of the DFT matrix as suggested in \cite[Chapter 7]{le2016matrices}, but also the one of the Hadamard matrix. In both cases, the butterfly factors of the DFT or the Hadamard matrix can be recovered up to scaling ambiguities via \Cref{algo:butterfly-fact} with \emph{any} factor-bracketing binary tree of $\integerSet{J}$ as input.
	Before proving \Cref{thm:identifiability-multilayer-butterfly}, we show that \Cref{algo:butterfly-fact} has controlled complexity bounds.
	
	\subsection{Complexity bounds}
	\label{subsec:complexity}
	Existing algorithms for butterfly factorization \cite{dao2019learning, le2016flexible} are based on gradient descent, and as such they require to tune several criteria such as learning rate or stopping criteria. In contrast, \Cref{algo:butterfly-fact} has a bounded complexity as it essentially consists in a controlled number of truncated SVDs to compute rank-one approximations of submatrices. 
	While the full SVD of a matrix of size $m \times n$ would require $\mathcal{O}(mn \min(m,n))$ flops, truncated SVD with numerical rank $k$ requires only $\mathcal{O}(kmn)$ flops (see e.g. \cite{halko2011finding} and references therein).
	Hence, in our complexity analysis of \Cref{algo:butterfly-fact}, the theoretical complexity of computing the best rank-one approximation of a matrix of size $m \times n$ will be $\mathcal{O}(mn)$.
	Below we estimate and compare the complexity for two types of factor-bracketing binary trees.
	
	\paragraph{Unbalanced tree} 
	First we consider running \Cref{algo:butterfly-fact} with a matrix $\mat{Z}$ of size $N \times N$, $N = 2^J$ ($J \geq 2$), and the \emph{unbalanced} factor-bracketing binary tree $\mathcal{T}$ of $\integerSet{J}$ (defined as the factor-bracketing binary tree where the left child of each non-leaf node is a singleton, see \Cref{fig:unbalanced-tree}) as inputs. 
	There are in total $J-1$ non-leaf nodes in this tree. 
	At the non-leaf node of depth $j \in \{0, \ldots, J-2\}$, the algorithm computes the best rank-one approximation of $N$ submatrices of size $2 \times N/2^{j+1}$, which yields a cost of the order of $N \times (2 \times N/2^{j+1}) = N^2 / 2^{j}$. Hence, the total cost of \Cref{algo:butterfly-fact} with the unbalanced factor-bracketing binary tree is of the order of:
$		\sum_{j=0}^{J-2} \frac{N^2}{2^{j}} 
		= 2 (1 - 2^{-J + 1}) N^2 = 2 \left(1 - \frac{2}{N} \right) N^2 =
		\mathcal{O}(N^2)$.
	Similarly, the complexity is $\mathcal{O}(N^2)$ when best rank-one approximations are computed with full SVDs  (see \Cref{app:complexity-complete-SVD}).
	
	\paragraph{Balanced tree}
	Consider now \Cref{algo:butterfly-fact} with an $N \times N$ matrix $\mat{Z}$, $N = 2^J$ where $J$ is also a power of $2$, and the \emph{balanced} factor-bracketing binary tree $\mathcal{T}$ of $\integerSet{J}$ (i.e., all children of each non-leaf node have the same cardinality, see \Cref{fig:balanced-tree}).
	At each non-leaf node of depth $k \in \{0, \ldots, \log_2(J) -1 \}$, the best rank-one approximation of $N$ square submatrices of size $\sqrt{N^{1/2^k}}$ is computed, at a cost of the order of $N \times (\sqrt{N^{1/2^k}} \times \sqrt{N^{1/2^k}}) = N \times N^{1/2^k}$. At each  depth $k \in \{0, \ldots, \log_2(J) -1 \}$, there are $2^{k}$ nodes. As $2^{k} \leq J/2 = \log_{2}(N)/2$ the total cost of \Cref{algo:butterfly-fact} is of the order of: 
$		\sum_{k=0}^{\log_2(J) - 1} 2^{k} N \times N^{1/2^k} = N^2 + \sum_{k=1}^{\log_2(J) - 1} 2^{k} N^{1 + 1/2^k}
		\leq N^{2} + \sum_{k=1}^{\log_2(\log_{2}(N)) - 1} \frac{\log_{2}(N)}{2} N^{3/2} 
		= \mathcal{O}(N^2)$.
	This contrasts with the complexity $\mathcal{O}(N^{5/2})$ when 	using full SVDs (see \Cref{app:complexity-complete-SVD}).
	
	\paragraph{Discussion}
	The complexity of \Cref{algo:butterfly-fact} is of the same order of magnitude as that of matrix-vector multiplications of size $N \times N$, which is $\mathcal{O}(N^{2})$.
	Assume that we want to compute the product $\mat{A} \mat{B}$, where $\mat{A}, \mat{B}$ are of size $N \times N$, and $\mat{A}$ admits the butterfly structure. 
	The naive computation requires $\mathcal{O}(N^3)$. But the data-sparse representation of $\mat{A}$ as a product of butterfly factors can be recovered by \Cref{algo:butterfly-fact} in $\mathcal{O}(N^2)$, in order to enable fast $\mathcal{O}(N \log N)$ matrix-vector multiplication \cite{dao2019learning}. In other words, this method allows for a computation of $\mat{A} \mat{B}$ in $\mathcal{O}(N^2 (\log N + 1))$ flops only, instead of $\mathcal{O}(N^3)$.

	Finally, it is possible to implement \Cref{algo:butterfly-fact} in a distributed fashion. Indeed, the computation of the best rank-one approximation of each submatrix at line \cref{line:best-rank1} can be performed in parallel: in the setting with $T$ threads, each thread computes the best rank-one approximation of $\lfloor N / T \rfloor$ submatrices. Moreover, when running \Cref{algo:butterfly-fact} with a balanced factor-bracketing binary tree, the implementation can be further parallelized, since the factorization at each node of the same depth can be performed by independent threads.

	\subsection{Proof of uniqueness (\Cref{thm:identifiability-multilayer-butterfly})}\label{section:proof-uniqueness}

	This subsection
	is dedicated to the proof of \Cref{thm:identifiability-multilayer-butterfly}.
	The reader more interested in the numerical aspects of the proposed method can skip this part and jump to
	\cref{section:experiments}.
	In order to prove \Cref{thm:identifiability-multilayer-butterfly}, consider $(\matseq{\mat{X}}{1}, \ldots, \matseq{\mat{X}}{J}) \in \Sigma_{\bflytuple}$ satisfying the butterfly constraint. These factors are assumed to verify the assumption of \Cref{thm:identifiability-multilayer-butterfly}. For any $1 \leq p \leq q \leq J$, denote $\partialproduct{\mat{X}}{p}{q} := \matseq{\mat{X}}{p} \ldots \matseq{\mat{X}}{q}$, and $\mat{Z} := \partialproduct{\mat{X}}{1}{J} = \matseq{\mat{X}}{1} \ldots \matseq{\mat{X}}{J}$. Fix any factor-bracketing binary tree $\mathcal{T}$ of $\integerSet{J}$. Given $\mat{Z}$ and $\mathcal{T}$ as the inputs of \Cref{algo:butterfly-fact}, we write $\partialproduct{\mat{M}}{p}{q}$ the intermediate matrix 
	obtained at each node $n := \{p, \ldots, q\}$ of $\mathcal{T}$ from the hierarchical factorization procedure.
	The proof is now separated into two steps. Firstly, we prove that \Cref{algo:butterfly-fact} recovers the butterfly factors $(\matseq{\mat{X}}{1}, \ldots, \matseq{\mat{X}}{J})$ from $\mat{Z}$, up to scaling ambiguities. Secondly, we prove that the butterfly factorization $\mat{Z} = \matseq{\mat{X}}{1} \ldots \matseq{\mat{X}}{J}$ 
	is indeed essentially unique in the sense of \Cref{def:essential-uniquess-multi}.
	
	\subsubsection{The algorithm recovers the butterfly factors}
	The proof of the first part consists in conducting an induction over the non-leaf nodes $n_1, \ldots, n_{J-1}$ of the tree, ordered in a breadth-first-search order (there are indeed in total $J-1$ non-leaf nodes in $\mathcal{T}$). 
	The general idea is to show that \Cref{algo:butterfly-fact} reconstructs recursively from $\mat{Z}$ the partial products $\partialproduct{\mat{X}}{p}{q}$ up to scaling ambiguities at each node $n = \{p, \ldots, q\}$ of the tree $\mathcal{T}$. 
	To that end, we need to prove a crucial lemma (\Cref{lemma:crux}) that essentially reduces the analysis of the multi-layer factorization to the case with only two factors. Then, the proof of \Cref{lemma:crux} itself relies on two key ingredients about optimality (\Cref{thm:optimality}) and essential uniqueness (\Cref{prop:key-step-for-identifiability-hierarchy}) of the considered two-layer factorization problem.
	
	For each $v \in \integerSet{J-1}$, denote 
	$\minnode{v}$, $\maxnode{v}$ the minimum and maximum
	index of node 
	$n_{v} = \{\minnode{v}, \ldots, \maxnode{v}\}$, and
	$\splitnode{v}$ as the ``splitting'' index of node $n_v$, which is the maximum value of its \emph{left} child.
	In the proof we will use the following 
	consequence of the breadth-first-search order.
	\begin{fact}
		\label{lemma:bfs}
		For any $u \in \{2, \ldots, J-1 \}$, the node $n_{u} =\{\minnode{u}, \ldots, \maxnode{u}\}$ is a child of 
		some node $n_{v} =\{\minnode{v},\ldots \splitnode{v}, \splitnode{v}+1, \dots, \maxnode{v}\}$ with $v \in \{1, \ldots, u-1\}$. 
		If $n_{u}$ is the left child of $n_{v}$, then $(\minnode{u}, \maxnode{u}) = (\minnode{v}, \splitnode{v})$.
		If $n_{u}$ is the right	child of $n_{v}$, then $(\minnode{u}, \maxnode{u}) = (\splitnode{v} + 1, \maxnode{v})$. 
	\end{fact}
	
	Define for any $V \in \integerSet{J-1}$, the assertion $P_V$: ``there exist invertible diagonal matrices $\matseq{\mat{D}}{\splitnode{1}}, \ldots, \matseq{\mat{D}}{\splitnode{V}}$ such that, for each $v \in \integerSet{V}$, we have: $\partialproduct{\mat{M}}{\minnode{v}}{\splitnode{v}} = \inverse{\matseq{\mat{D}}{\minnode{v}-1}} \partialproduct{\mat{X}}{\minnode{v}}{\splitnode{v}} \matseq{\mat{D}}{\splitnode{v}}$ and 
	$\partialproduct{\mat{M}}{\splitnode{v}+1}{\maxnode{v}} = \inverse{\matseq{\mat{D}}{\splitnode{v}}}   \partialproduct{\mat{X}}{\splitnode{v}+1}{\maxnode{v}} \matseq{\mat{D}}{\maxnode{v}}$", 
	with the convention $\matseq{\mat{D}}{0} = \matseq{\mat{D}}{J} = \identity{N}$\footnote{Remark that for any $v \in \integerSet{V}$ the node $n_v = \{\minnode{v},\ldots,\maxnode{v}\}$ is either the root node (when $v=1$) or a child of a node $n_{w}$ with $w < v$. In the latter case, by \Cref{lemma:bfs}, either $(\minnode{v}, \maxnode{v}) = (\minnode{w}, \splitnode{w})$, or $(\minnode{v}, \maxnode{v}) = (\splitnode{w} + 1, \maxnode{w})$, with $w \in \integerSet{v-1} \subseteq \integerSet{V}$. In other words, $\minnode{v}-1, \splitnode{v}, \maxnode{v} \in \{ 0, \splitnode{1}, \ldots, \splitnode{V}, J\}$ for all $v \in \integerSet{V}$, meaning that the diagonal matrices $\matseq{\mat{D}}{\minnode{v}-1}$, $\matseq{\mat{D}}{\splitnode{v}}$ and $\matseq{\mat{D}}{\maxnode{v}}$ used
		in the definition of $P_V$ above are well defined.}.
	The principle of the proof is to show $P_V$ by induction for all $V \in \integerSet{J-1}$. In particular, proving $P_{J-1}$ yields our claim, as we now explain.
	Indeed, any leaf node $n = \{ \ell \}$ is either a left child or a right child of a non-leaf node $n_v = \{\minnode{v},\ldots, 
	\maxnode{v}\}$ with $v \in \integerSet{J-1}$. 
	In the case of a left child, $\{\ell\} = \{\minnode{v},\ldots, \splitnode{v}\}$ hence $\ell = \minnode{v} = \splitnode{v}$, and in the other case $\{\ell\} = \{\splitnode{v}+1,\ldots, \maxnode{v}\}$ hence $\ell = \splitnode{v} + 1 = \maxnode{v}$.
	In both cases assertion $P_{J-1}$ implies that $\partialproduct{\mat{M}}{\ell}{\ell} = \inverse{\matseq{\mat{D}}{\ell-1}} \partialproduct{\mat{X}}{\ell}{\ell} \matseq{\mat{D}}{\ell} = \inverse{\matseq{\mat{D}}{\ell-1}} \matseq{\mat{X}}{\ell} 
	\matseq{\mat{D}}{\ell}$, hence $(\partialproduct{\mat{M}}{\ell}{\ell})_{\ell=1}^J \sim (\matseq{\mat{X}}{\ell})_{\ell=1}^J$, meaning that the algorithm recovers the butterfly factors up to scaling ambiguities only.
		The crux of the proof is the following lemma.
		\begin{lemma}
			\label{lemma:crux}
			Under the assumptions of \Cref{thm:identifiability-multilayer-butterfly}, consider $V \in \integerSet{J-1}$ and assume that there are invertible diagonal matrices $\matseq{\mat{D}}{\minnode{V}-1}$ and $\matseq{\mat{D}}{\maxnode{V}}$ such that
$				\partialproduct{\mat{M}}{\minnode{V}}{\maxnode{V}} 
				=  
				\inverse{\matseq{\mat{D}}{\minnode{V} - 1}} 
				\partialproduct{\mat{X}}{\minnode{V}}{\maxnode{V}}
				\matseq{\mat{D}}{\maxnode{V}}.$
			Then the pair $( \partialproduct{\mat{M}}{\minnode{V}}{\splitnode{V}}, \transpose{\partialproduct{\mat{M}}{\splitnode{V} +1}{\maxnode{V}}})$ computed at line \ref{line:apply-fsmf} of \Cref{algo:hierarchical-step} such that the product $\partialproduct{\mat{M}}{\minnode{V}}{\splitnode{V}} \partialproduct{\mat{M}}{\splitnode{V}+1}{\maxnode{V}}$ approximates $\partialproduct{\mat{M}}{\minnode{V}}{\maxnode{V}}$ is equal (up to scaling ambiguities) to the pair $(\bar{\mat{X}}, \bar{\mat{Y}})$  where $\bar{\mat{X}} := \inverse{\matseq{\mat{D}}{\minnode{V}-1}} \partialproduct{\mat{X}}{\minnode{V}}{\splitnode{V}}$, $\bar{\mat{Y}} := \transpose{(\partialproduct{\mat{X}}{\splitnode{V}+1}{\maxnode{V}}  \matseq{\mat{D}}{\maxnode{V}})}$.
		\end{lemma}
	
	Indeed, since $\partialproduct{\mat{M}}{\minnode{1}}{\maxnode{1}} = \mat{Z} =  \partialproduct{\mat{X}}{\minnode{1}}{\maxnode{1}}$ and since $
		\matseq{\mat{D}}{\minnode{1}-1} = \matseq{\mat{D}}{\maxnode{1}} =  \identity{N}$ (recall that $\minnode{1} = 1$ and $\maxnode{1} = J$),  this lemma applied to $V=1$ shows that $P_{1}$ is true. This starts the induction, and we now show that the lemma
		can similarly be used to proceed to the induction. Assume that $P_{V-1}$ is true where $V\in \{2, \ldots, J-1\}$ and consider $\matseq{\mat{D}}{\splitnode{1}}, \ldots, \matseq{\mat{D}}{\splitnode{V-1}}$ the corresponding invertible diagonal matrices.
		By \Cref{lemma:bfs}, the parent of node $n_V$ is necessarily
		some node $n_v$ with $v \in \integerSet{V-1}$ and, depending on whether $n_V$ is a left or right child of $n_v$, we have either $n_V = \{ \minnode{v}, \ldots, \splitnode{v} \}$ or  $n_V = \{ \splitnode{v} + 1, \ldots, \maxnode{v} \}$. 
		Without loss of generality assume the former (the proof is similar if we suppose the latter)
		so that $(\minnode{V}, \maxnode{V}) = (\minnode{v}, \splitnode{v})$. Since $P_{V-1}$ is true we have
		 that $\partialproduct{\mat{M}}{\minnode{V}}{\maxnode{V}} = \partialproduct{\mat{M}}{\minnode{v}}{\splitnode{v}} = \inverse{\matseq{\mat{D}}{\minnode{v}-1}} \partialproduct{\mat{X}}{\minnode{v}}{\splitnode{v}} \matseq{\mat{D}}{\splitnode{v}} = \inverse{\matseq{\mat{D}}{\minnode{V}-1}} \partialproduct{\mat{X}}{\minnode{V}}{\maxnode{V}} \matseq{\mat{D}}{\maxnode{V}}$.
	By \Cref{lemma:crux}, there exists an invertible diagonal matrix $\matseq{\mat{D}}{\splitnode{V}}$ such that $\partialproduct{\mat{M}}{\minnode{V}}{\splitnode{V}} = \inverse{\matseq{\mat{D}}{\minnode{V}-1}} \partialproduct{\mat{X}}{\minnode{V}}{\splitnode{V}} \matseq{\mat{D}}{\splitnode{V}}$ and 
		$\partialproduct{\mat{M}}{\splitnode{V}+1}{\maxnode{V}} = \inverse{\matseq{\mat{D}}{\splitnode{V}}}   \partialproduct{\mat{X}}{\splitnode{V}+1}{\maxnode{V}} \matseq{\mat{D}}{\maxnode{V}}$, which proves $P_V$.
	
	We now focus on the proof of \Cref{lemma:crux}. It relies on two key ingredients formulated in \Cref{thm:optimality} and \Cref{prop:key-step-for-identifiability-hierarchy} below. They are both derived from the following property of the butterfly supports, proved in \Cref{app:butterfly-supp-disjoint}.
	
	\begin{lemma}
		\label{lemma:butterfly-supp-disjoint}
		Given $1 \leq p \leq \ell < q \leq J$, $\leftsupp{S} := \bflypartial{p}{\ell}$ and $\rightsupp{S} := \transpose{\bflypartial{\ell+1}{q}}$ where we recall \eqref{eq:W_p_q}, the tuple of rank-one supports $\varphi \left( \leftsupp{S} , \rightsupp{S} \right)$ has disjoint rank-one supports.
	\end{lemma}
	
	The first consequence of this property is the following optimality theorem.
	
	\begin{theorem}[{\cite[Application of Theorem 3.3]{papierTung}}]
		\label{thm:optimality}
		Denote $\pair{S} := ( \bflypartial{p}{\ell}, \transpose{\bflypartial{\ell+1}{q}} )$. For any matrix $\mat{M}$, the procedure $\Call{FSMF}{\mat{M}, \bflypartial{p}{\ell}, \transpose{\bflypartial{\ell+1}{q}}}$ described in \Cref{algo:fsmf} computes in polynomial time a pair of factors solving the problem:
		$\min_{(\mat{X}, \mat{Y})  \in \Sigma_{\pair{S}}} \| \mat{M} - \mat{X} \transpose{\mat{Y}}\|_{F}$.
	\end{theorem}

	\Cref{lemma:butterfly-supp-disjoint} also allows us to characterize the set $\unique(\Sigma_{\pair{S}})$ when $\pair{S} := ( \bflypartial{p}{\ell}, \transpose{\bflypartial{\ell+1}{q}} )$.
	
	\begin{lemma}
		\label{lemma:hierarchical-level-id-butterfly-supports}
		Denote $\pair{S} = (\leftsupp{S},\rightsupp{S}) := ( \bflypartial{p}{\ell}, \transpose{\bflypartial{\ell+1}{q}} )$. We have
		\begin{equation*}
			\unique \left(\Sigma_{\pair{S}} \right) = \left \{ (\mat{X}, \mat{Y}) \in \Sigma_{\pair{S}} \; | \; \colsupp(\mat{X}) = \colsupp(\mat{Y}) = \integerSet{N} \right \}.
		\end{equation*}
	\end{lemma}
	
	\begin{proof}
		By \Cref{prop:disjoint-rank-one-supports-unique-EMF}, $\unique \left(\Sigma_{\pair{S}} \right) = \idcolsupp{\pair{S}} \cap \maxcolsupp{\pair{S}}$. Since $\colsupp(\leftsupp{S}) = \colsupp(\rightsupp{S}) = \integerSet{N}$, by definition of 
			$\idcolsupp{\pair{S}}$ and $\maxcolsupp{\pair{S}}$ (cf~\eqref{eq:DefIC}-\eqref{eq:DefMC})
		we have $(\mat{X}, \mat{Y}) \in \idcolsupp{\pair{S}} \cap \maxcolsupp{\pair{S}}$ if, and only if, $\mat{X}$ and $\mat{Y}$ do not have a zero column, i.e., $\colsupp(\mat{X}) = \colsupp(\mat{Y}) = \integerSet{N}$.
	\end{proof}
	
	Recall that the factors $(\matseq{\mat{X}}{1}, \ldots, \matseq{\mat{X}}{J})$ verify the assumption of \Cref{thm:identifiability-multilayer-butterfly}, i.e., $\matseq{\mat{X}}{\ell}$ does not have a zero column for $1 \leq \ell \leq J-1$, and not a zero row for $2 \leq \ell \leq J$. As claimed in the following lemma proved in \Cref{app:no-zero-col-row}, this assumption is in fact a necessary and sufficient condition to ensure that each pair of partial products $(\partialproduct{\mat{X}}{p}{\ell}, \transpose{\partialproduct{\mat{X}}{\ell+1}{q}})$ with $1 \leq p \leq \ell < q \leq J$ is non-degenerate (the left and right factor do not have a zero column).
	
	\begin{lemma}
		\label{lemma:no-zero-col-row}
		Let $\bflytuple$ be the butterfly supports of size $N = 2^J$, and $(\matseq{\mat{X}}{1}, \ldots, \matseq{\mat{X}}{J}) \in \Sigma_{\bflytuple}$. The following are equivalent:
		\begin{enumerate}[label=(\roman*)]
			\item for each $1 \leq p \leq \ell < q \leq J$, $\partialproduct{\mat{X}}{p}{\ell} := \matseq{\mat{X}}{p} \ldots \matseq{\mat{X}}{\ell}$ does not have a zero column, and $\partialproduct{\mat{X}}{\ell+1}{q} := \matseq{\mat{X}}{\ell+1} \ldots \matseq{\mat{X}}{q}$ does not have a zero row;
			\item  $\matseq{\mat{X}}{\ell}$ does not have a zero column for $1 \leq \ell \leq J-1$, and not a zero row for $2 \leq \ell \leq J$.
		\end{enumerate}
	\end{lemma}

	Consequently, we obtain the second key ingredient for the proof of \Cref{lemma:crux}.
	
	\begin{proposition}
		\label{prop:key-step-for-identifiability-hierarchy}
		Assume that $(\matseq{\mat{X}}{1}, \ldots, \matseq{\mat{X}}{J}) \in \Sigma_{\bflytuple}$ verifies the hypothesis of \Cref{thm:identifiability-multilayer-butterfly}.		
		Let $1 \leq p \leq \ell < q \leq J$. Denote $\pair{S} := ( \bflypartial{p}{\ell}, \transpose{\bflypartial{\ell+1}{q}} )$.
		Then, for any invertible diagonal matrices $\mat{D}, \bar{\mat{D}}$, denoting $ \bar{\mat{X}} := \inverse{\mat{D}} \partialproduct{\mat{X}}{p}{\ell}$ and $\bar{\mat{Y}} = \transpose{(\partialproduct{\mat{X}}{\ell+1}{q} \bar{\mat{D}})}$, the factorization $\inverse{\mat{D}} \partialproduct{\mat{X}}{p}{q} \bar{\mat{D}}=\bar{\mat{X}} \transpose{\bar{\mat{Y}}}$ into two factors $(\bar{\mat{X}},\bar{\mat{Y}})$ is essentially unique in $\Sigma_{\pair{S}}$. 
	\end{proposition}
	
	\begin{proof}
		By \Cref{lemma:block-structure-product-butterfly},  $(\partialproduct{\mat{X}}{p}{\ell}, \transpose{\partialproduct{\mat{X}}{\ell+1}{q}}) \in \Sigma_{\pair{S}}$.
		By \Cref{lemma:no-zero-col-row} and the assumption on the factors $\matseq{\mat{X}}{\ell}$, $1 \leq \ell \leq J$, 
		the matrices $\partialproduct{\mat{X}}{p}{\ell}$ and $\transpose{\partialproduct{\mat{X}}{\ell+1}{q}}$ do not have a zero column. The same is true for $\inverse{\mat{D}} \partialproduct{\mat{X}}{p}{\ell}$ and $\bar{\mat{D}} \transpose{\partialproduct{\mat{X}}{\ell+1}{q}}$, as the multiplication of a matrix by $\inverse{\mat{D}}$ or $\bar{\mat{D}}$ does not change its support.
		By \Cref{lemma:hierarchical-level-id-butterfly-supports}, $(\inverse{\mat{D}} \partialproduct{\mat{X}}{p}{\ell}, \bar{\mat{D}} \transpose{\partialproduct{\mat{X}}{\ell+1}{q}}) \in \unique(\Sigma_{\pair{S}})$.
	\end{proof}

	We now have all the ingredients to prove the crucial \Cref{lemma:crux}.
	
	\begin{proof}[Proof of \Cref{lemma:crux}]
		Since $\partialproduct{\mat{X}}{\minnode{V}}{\maxnode{V}} = \partialproduct{\mat{X}}{\minnode{V}}{\splitnode{V}} \partialproduct{\mat{X}}{\splitnode{V}+1}{\maxnode{V}}$, we can factorize the matrix $\partialproduct{\mat{M}}{\minnode{V}}{\maxnode{V}} 
			=  
			\inverse{\matseq{\mat{D}}{\minnode{V}-1}} 
			\partialproduct{\mat{X}}{\minnode{V}}{\maxnode{V}}
			\matseq{\mat{D}}{\maxnode{V}}$ as
$			\partialproduct{\mat{M}}{\minnode{V}}{\maxnode{V}} =
\bar{\mat{X}} \transpose{\bar{\mat{Y}}}$,
		with $\bar{\mat{X}} := \inverse{\matseq{\mat{D}}{\minnode{V}-1}} \partialproduct{\mat{X}}{\minnode{V}}{\splitnode{V}}$, $\bar{\mat{Y}} := \transpose{ (\partialproduct{\mat{X}}{\splitnode{V}+1}{\maxnode{V}}  \matseq{\mat{D}}{\maxnode{V}})}$. 
		By \Cref{lemma:block-structure-product-butterfly}, 
		$\left( \bar{\mat{X}}, \bar{\mat{Y}} \right) \in \Sigma_{\pair{S}}$ where $\pair{S} := (\bflypartial{\minnode{V}}{\splitnode{V}}, \transpose{\bflypartial{\splitnode{V}  + 1}{\maxnode{V}}})$. 
		By \Cref{thm:optimality}, the factors $( \partialproduct{\mat{M}}{\minnode{V}}{\splitnode{V}}, \transpose{\partialproduct{\mat{M}}{\splitnode{V}+1}{\maxnode{V}}}) \in \Sigma_{\pair{S}}$ computed at line \ref{line:apply-fsmf} of \Cref{algo:hierarchical-step} 
		minimimize the optimization problem: $\min_{(\mat{X}, \mat{Y}) \in \Sigma_{\pair{S}}} \| \partialproduct{\mat{M}}{\minnode{V}}{\maxnode{V}} - \mat{X} \transpose{\mat{Y}}\|_{F}$. Since $\partialproduct{\mat{M}}{\minnode{V}}{\maxnode{V}} = \bar{\mat{X}} \transpose{\bar{\mat{Y}}}$ with $\left( \bar{\mat{X}}, \bar{\mat{Y}} \right) \in \Sigma_{\pair{S}}$, this minimum is zero hence the computed pair satisfies
		$\partialproduct{\mat{M}}{\minnode{V}}{\maxnode{V}} = \partialproduct{\mat{M}}{\minnode{V}}{\splitnode{V}} \partialproduct{\mat{M}}{\splitnode{V}+1}{\maxnode{V}}$. By \Cref{prop:key-step-for-identifiability-hierarchy}, the factorization $\partialproduct{\mat{M}}{\minnode{V}}{\maxnode{V}} = \bar{\mat{X}} \transpose{\bar{\mat{Y}}}$ into two factors is essentially unique in $\Sigma_{\pair{S}}$, so $( \partialproduct{\mat{M}}{\minnode{V}}{\splitnode{V}}, \transpose{\partialproduct{\mat{M}}{\splitnode{V}+1}{\maxnode{V}}}) \sim (\bar{\mat{X}}, \bar{\mat{Y}})$.
	\end{proof}
	
	This ends the inductive proof of the first part of \Cref{thm:identifiability-multilayer-butterfly} showing that \Cref{algo:butterfly-fact} recovers the butterfly factors $(\matseq{\mat{X}}{\ell})_{\ell=1}^J$ from $\mat{Z} = \matseq{\mat{X}}{1} \ldots \matseq{\mat{X}}{J}$, up to scaling ambiguities.

	\subsubsection{The factorization is essentially unique} We now show that the factorization $\mat{Z} = \matseq{\mat{X}}{1} \ldots \matseq{\mat{X}}{J}$ is essentially unique in $\Sigma_{\bflytuple}$ in the sense of \Cref{def:essential-uniquess-multi}. To that end, consider the factors $(\partialproduct{\mat{M}}{\ell}{\ell})_{\ell=1}^J$ computed using \Cref{algo:butterfly-fact} with $\mat{Z}$ as input, as well as arbitrary factors  $(\matseq{\bar{\mat{X}}}{\ell})_{\ell=1}^J \in \Sigma_{\bflytuple}$ such that $\matseq{\bar{\mat{X}}}{1} \ldots \matseq{\bar{\mat{X}}}{J} = \mat{Z}$. 
	We will show that $(\matseq{\bar{\mat{X}}}{\ell})_{\ell=1}^J$ verifies the assumptions of \Cref{thm:identifiability-multilayer-butterfly}: this will imply that $(\partialproduct{\mat{M}}{\ell}{\ell})_{\ell=1}^J$ is rescaling-equivalent \emph{both} to $(\matseq{\mat{X}}{\ell})_{\ell=1}^J$ and to $(\matseq{\bar{\mat{X}}}{\ell})_{\ell=1}^J$, hence, by transitivity, $(\matseq{\mat{X}}{\ell})_{\ell=1}^J \sim (\matseq{\bar{\mat{X}}}{\ell})_{\ell=1}^J$ as claimed.
	
	Denote $\partialproduct{\bar{\mat{X}}}{p}{q} := \matseq{\bar{\mat{X}}}{p} \ldots \matseq{\bar{\mat{X}}}{q}$ for any $1 \leq p \leq q \leq J$.
	For any $\ell \in \integerSet{J-1}$, we have $\partialproduct{\mat{X}}{1}{\ell} \partialproduct{\mat{X}}{\ell+1}{J} = \mat{Z} = \partialproduct{\bar{\mat{X}}}{1}{\ell} \partialproduct{\bar{\mat{X}}}{\ell+1}{J}$. 
	By \Cref{lemma:block-structure-product-butterfly},
	$(\partialproduct{\bar{\mat{X}}}{1}{\ell}, \transpose{\partialproduct{\bar{\mat{X}}}{\ell+1}{J}}) \in \Sigma_{\pair{S}}$ 
	where $\pair{S} := (\bflypartial{1}{\ell}, \transpose{\bflypartial{\ell+1}{J}})$.
	Besides, by assumption and by \Cref{lemma:no-zero-col-row}, $\partialproduct{\mat{X}}{1}{\ell}$ and $\transpose{\partialproduct{\mat{X}}{\ell+1}{J}}$ do not have a zero column, so by \Cref{lemma:hierarchical-level-id-butterfly-supports}, $(\partialproduct{\mat{X}}{1}{\ell}, \transpose{\partialproduct{\mat{X}}{\ell+1}{J}}) \in \unique(\Sigma_{\pair{S}})$. By definition of 	the set $\unique(\Sigma_{\pair{S}})$ of essentially unique factors, $(\partialproduct{\bar{\mat{X}}}{1}{\ell}, \transpose{\partialproduct{\bar{\mat{X}}}{\ell+1}{J}}) \sim (\partialproduct{\mat{X}}{1}{\ell}, \transpose{\partialproduct{\mat{X}}{\ell+1}{J}})$.
	Since $\partialproduct{\mat{X}}{1}{\ell}$ and $\transpose{\partialproduct{\mat{X}}{\ell+1}{J}}$ do not have a zero column, this implies that $\partialproduct{\bar{\mat{X}}}{1}{\ell}$ and $\transpose{\partialproduct{\bar{\mat{X}}}{\ell+1}{J}}$ also do not have a zero column.
	Consequently, $\matseq{\bar{\mat{X}}}{\ell}$ does not have a zero column (otherwise $\partialproduct{\bar{\mat{X}}}{1}{\ell}$ would have a zero column) and similarly $\matseq{\bar{\mat{X}}}{\ell+1}$ does not have a zero row. As this holds for any $\ell \in \integerSet{J-1}$, $(\matseq{\bar{\mat{X}}}{\ell})_{\ell=1}^J$ verifies the assumption of \Cref{thm:identifiability-multilayer-butterfly}. This ends the proof of \Cref{thm:identifiability-multilayer-butterfly}.

	\subsection{Relation with the complementary low-rank property}
	\label{subsec:complementary-low-rank}
		
	We now relate the butterfly structure (\Cref{def:butterfly-structure}) to the complementary low-rank property \cite{candes2009fast,li2015butterfly},  formally introduced here for a square matrix of size $N \times N$ for $N=2^J$ using the notations from \cite{li2015butterfly}.

	\begin{definition}[Complementary low-rank property \cite{li2015butterfly}]
		\label{def:complementary-low-rank}
			Consider two binary trees $T_X$ and $T_\Omega$ of maximum depth $J$, called \emph{index-partitioning} binary trees, such that each node is a non-empty subset of indices of $[N]$, $N := 2^J$, with the root being $\integerSet{N}$ and the children of each non-leaf node forming a partition of their parent. A matrix $\mat{Z}$ of size $N \times N$ satisfies the \emph{complementary low-rank property} for the trees $T_X$ and $T_\Omega$ if, for any level $\ell \in \{ 0, \ldots, J \}$, any node $A$ in $T_X$ at level $J - \ell$, and any node $B$ in $T_\Omega$ at level $\ell$, the submatrix $\matindex{\mat{Z}}{A}{B}$ obtained by restricting $\mat{Z}$ to the rows indexed by $A$ and to the columns indexed by $B$ is of low rank.
	\end{definition}

	We claim that a matrix having the butterfly structure satisfies the complementary low-rank property for specific choices of index-partitioning binary trees $T_X$ and $T_\Omega$.
	
	\begin{proposition}
		\label{prop:complementary-low-rank}
		Construct the index-partitioning binary trees $T_X$ and $T_\Omega$ in the following way:
		(a) each node $A := \{ a_1, a_2, \ldots, a_{N / 2^\ell} \}$ of $T_X$ at level $\ell \in \{0, \ldots, J-1 \}$, where $a_1 < \ldots < a_{N / 2^\ell}$, has $\{ a_1, a_3, \ldots, a_{N / 2^\ell - 1}\}$ as its left child and $\{ a_2, a_4, \ldots, a_{N / 2^\ell} \}$ as its right child;
		(b) each node $B := \{ b_1, b_2, \ldots, b_{N / 2^\ell} \}$ of $T_\Omega$ at level $\ell \in \{0, \ldots, J-1 \}$, where $b_1 < \ldots < b_{N / 2^\ell}$, has $\{ b_1, b_2, \ldots, b_{N / 2^{\ell+1}} \}$ as its left child and $\{ b_{N / 2^{\ell+1} + 1}, b_{N / 2^{\ell+1} + 2}, \ldots, b_{N / 2^\ell} \}$ as its right child.
		Consider $\bflytuple$ the butterfly supports of size $N = 2^J$ and $(\matseq{\mat{X}}{1}, \ldots, \matseq{\mat{X}}{J}) \in \Sigma_{\bflytuple}$.
		Then, $\mat{Z} := \matseq{\mat{X}}{1} \ldots \matseq{\mat{X}}{J}$ satisfies the complementary low-rank property for the trees $T_X$ and $T_\Omega$.
	\end{proposition}

	\begin{remark}
		\label{rmk:tree_T_X}
		Each node $A$ in $T_X$ 
		at level $J - \ell$ for $\ell \in \{0, \ldots, J \}$ is of the form $\{ c + k N / 2^\ell, \, k=0, \ldots, 2^\ell - 1\}$ for an index $c \in \integerSet{N / 2^\ell}$. Each node $B$ in $T_\Omega$ at level $\ell \in \{0, \ldots, J \}$ is of the form $\{ (c - 1) N / 2^\ell + k, \, k=1, \ldots, N / 2^\ell \}$ for an index $c \in \integerSet{2^\ell}$.
	\end{remark}
	
	\begin{proof}
		Denoting $\partialproduct{\mat{X}}{p}{q} := \matseq{\mat{X}}{p} \ldots \matseq{\mat{X}}{q}$ for any $1 \leq p \leq q \leq J$, 
		we have $\mat{Z} = \partialproduct{\mat{X}}{1}{\ell} \partialproduct{\mat{X}}{\ell+1}{J}$ for any $\ell \in \integerSet{J-1}$. 
		Fix $\ell \in \integerSet{J-1}$, and denote $\pair{S} := (\bflypartial{1}{\ell}, \transpose{\bflypartial{\ell+1}{J}})$, where we recall the notation \eqref{eq:W_p_q}. 
		By \Cref{lemma:block-structure-product-butterfly}, $(\partialproduct{\mat{X}}{1}{\ell}, \transpose{\partialproduct{\mat{X}}{\ell+1}{J}}) \in \Sigma_{\pair{S}}$. This means that $(\rankone{C}{i})_{i=1}^N := \varphi(\partialproduct{\mat{X}}{1}{\ell}, \transpose{\partialproduct{\mat{X}}{\ell+1}{J}}) \in \Gamma_{\tuplerkone{S}}$ where $\tuplerkone{S} = (\rankone{S}{i})_{i = 1}^N := \varphi(\bflypartial{1}{\ell}, \transpose{\bflypartial{\ell+1}{J}})$. 
		Consider any node $A$ in $T_X$ at level $J - \ell$ and any node $B$ in $T_\Omega$ at level $\ell$. By \Cref{rmk:tree_T_X} and \eqref{eq:W_p_q}, one can verify that there exists $i \in \integerSet{N}$ 
		such that the support of the $i$-th column of $\bflypartial{1}{\ell}$ and the support of the $i$-th row of $\bflypartial{\ell+1}{J}$ are respectively the node $A$ and $B$, which means that $\mat{Z} \odot \rankone{S}{i}$ is the submatrix $\matindex{\mat{Z}}{A}{B}$. 
		But by \Cref{lemma:butterfly-supp-disjoint}, the rank-one supports $(\rankone{S}{i})_{i = 1}^N$ are pairwise disjoint, so $\mat{Z} \odot \rankone{S}{i} = ( \sum_{i'=1}^{N} \rankone{C}{i'} ) \odot \rankone{S}{i} = \rankone{C}{i}$, which by definition is of rank one. This is true for any level $\ell$, any node $A$ in $T_X$ at level $J - \ell$, and any node $B$ in $T_\Omega$ at level $\ell$. 
		By \Cref{def:complementary-low-rank}, $\mat{Z}$ satisfies the complementary low-rank property for the trees $T_X$ and $T_\Omega$.
	\end{proof}

	With the index-partitioning binary trees $T_X$ and $T_\Omega$ of \Cref{prop:complementary-low-rank}, the butterfly algorithm from \cite{li2015butterfly} (with non-randomized SVDs) is \Cref{algo:butterfly-fact} with the \emph{symmetric} factor-bracketing binary tree $\mathcal{T}$ illustrated in \Cref{fig:symmetric-tree} and defined for $N=2^{J}$ with an even integer $J$ as: (a) the children of the root node of $\mathcal{T}$ are $\{1, \ldots, J/2 \}$ and $\{ J/2 + 1, \ldots, J \}$; (b) in the left (resp.~right) subtree of the root of $\mathcal{T}$, every right (resp.~left) child of a non-leaf node is a singleton. 
	Applying the procedure of line \ref{line:apply-fsmf} in \Cref{algo:hierarchical-step} at the root node of $\mathcal{T}$ corresponds to the ``middle-level factorization'' in the butterfly algorithm \cite{li2015butterfly}, and applying successively this procedure at the other nodes of $\mathcal{T}$ corresponds to the ``recursive factorization" in \cite{li2015butterfly}.
	
	\section{Numerical experiments}
	\label{section:experiments}
	The empirical behavior of \Cref{algo:butterfly-fact} was first studied in \cite{le:hal-03438881}, showing this superiority of the algorithm for solving \eqref{eq:butterfly-smf} compared to gradient-based 
	optimization \cite{dao2019learning}, in terms of running time and approximation error, both in the exact and noisy setting.  As an original contribution of this paper, the analysis in \cref{subsec:complexity} explains this improved running time, as we show that the total complexity of \Cref{algo:butterfly-fact} is $\mathcal{O}(N^2)$ 
	using truncated SVDs. 
	However, as empirically shown below, it is important to use an appropriate implementation of SVD in order to achieve this complexity bound.

	\begin{figure}[t]
		\centering
		\begin{subfigure}[t]{0.48\textwidth}
			\centering
			\includegraphics[width=\textwidth]{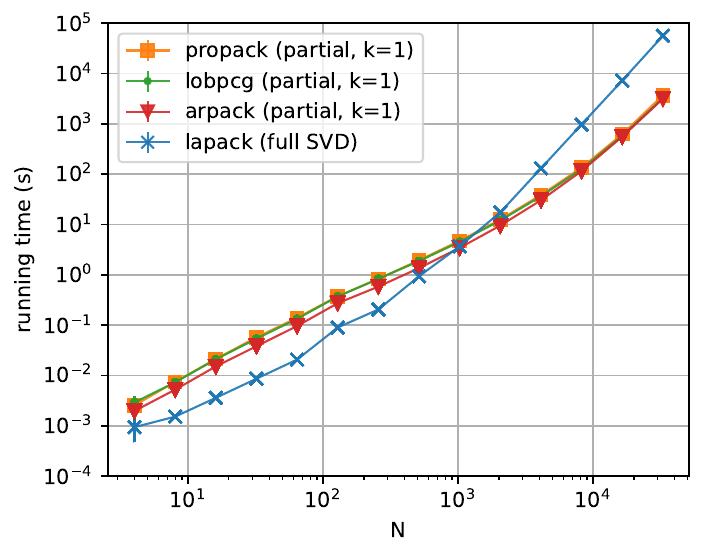}
			\caption{Unbalanced tree, matrix of size $N \times N$}
			\label{fig:unbalanced}
		\end{subfigure}%
		~ 
		\begin{subfigure}[t]{0.48\textwidth}
			\centering
			\includegraphics[width=\textwidth]{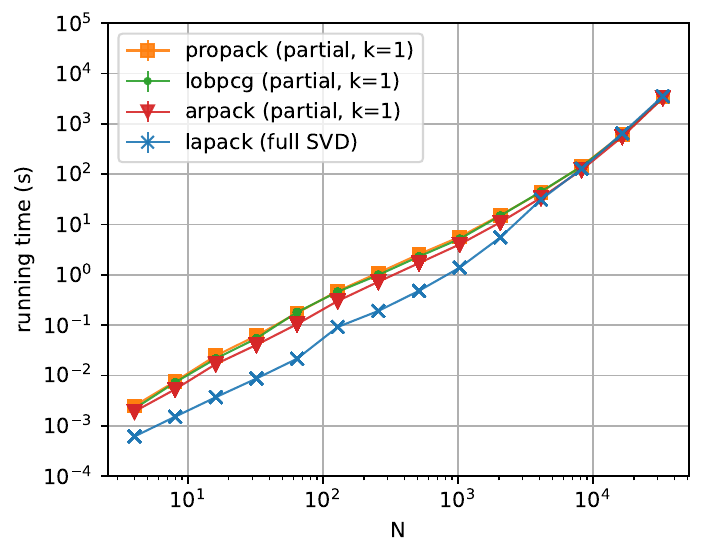}
			\caption{Balanced tree, matrix of size $N \times N$}
			\label{fig:balanced}
		\end{subfigure}%
		~
		\newline
		\begin{subfigure}[t]{0.48\textwidth}
			\centering
			\includegraphics[width=\textwidth]{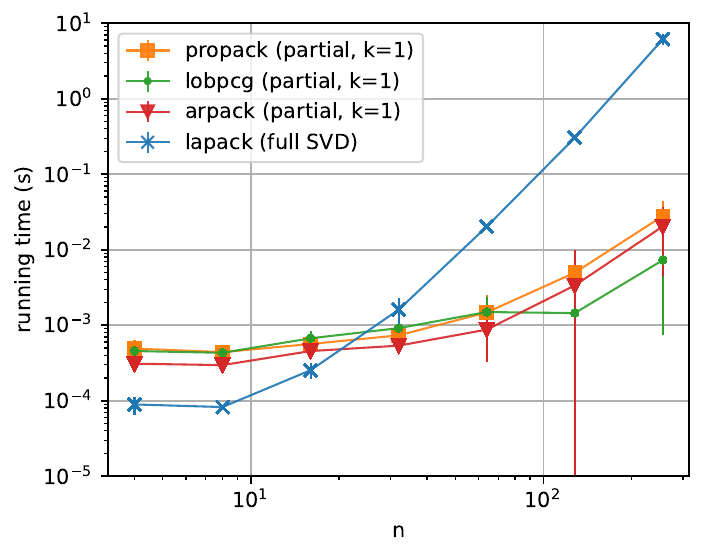}
			\caption{SVD for matrices of size $2 \times n^2 / 2$}
			\label{fig:svd_rectangle}
		\end{subfigure}%
		~ 
		\begin{subfigure}[t]{0.48\textwidth}
			\centering
			\includegraphics[width=\textwidth]{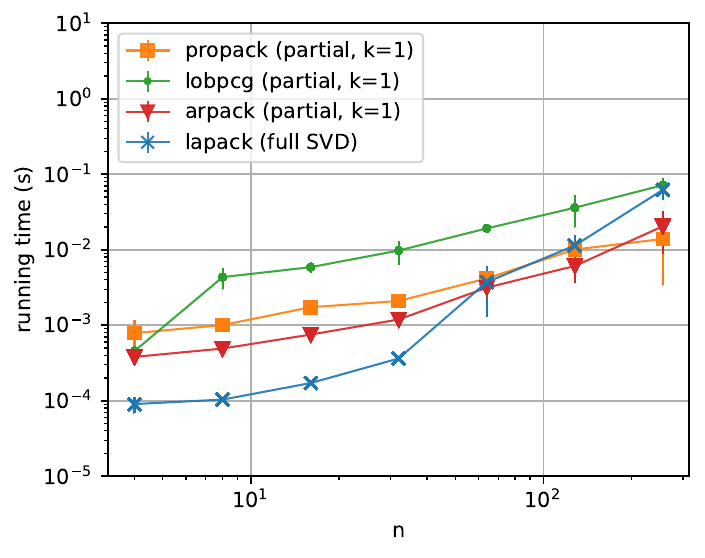}
			\caption{SVD for matrices of size $n \times n$}
			\label{fig:svd_square}
		\end{subfigure}%
		\caption{Comparing different solvers - running time of our implementation of \Cref{algo:butterfly-fact}. For a given matrix size, each factorization (resp.~SVD) is repeated 3 (resp.~10) times in order to plot the mean running time with a vertical error bar showing the standard deviation.}
		\label{fig:compare_solvers}
	\end{figure}

	\paragraph{Protocol}
	We implement \Cref{algo:butterfly-fact} in Python using the \texttt{Scipy 1.8.0} package to compute the SVD for the best rank-one 
	approximation. We measure the running time of our implementation of \Cref{algo:butterfly-fact} to approximate $\mat{Z} = \mat{H} + \sigma \mat{W}$ as a product of $J$ butterfly factors, where $\mat{H}$ is the Hadamard matrix of size $N=2^J$ with $J \in \{2, \ldots, 15 \}$, $\mat{W}$ is a random matrix with i.i.d.\ standard Gaussian entries (zero mean and variance equal to 1), and $\sigma = 0.01$. 
	We use different SVD solvers in our experiments, among: LAPACK, ARPACK, PROPACK, LOBPCG. The first one performs a full SVD before truncating it, while the last ones perform partial SVD at a given order $k$ ($k=1$ in our case). 
	Experiments are run on an Intel(R) Xeon(R) Gold 5218 CPU @ 2.30GHz. In the interest of reproducible research, our implementation is available in open source \cite{zheng:hal-03620052v1}. An optimized C++ implementation with Python and Matlab wrappers is also provided in the FAµST 3.25 toolbox at \url{https://faust.inria.fr/}.
	
		\paragraph{Comparing different solvers} 
We compare the running time of the hierarchical factorization implemented with different SVD solvers. In \Cref{fig:unbalanced}, the unbalanced hierarchical factorization is faster with the LAPACK solver for $N \leq 1024$, while the other solvers are faster for $N \geq 1024$. In \Cref{fig:balanced}, the balanced hierarchical factorization is faster with the LAPACK solver for $N \leq 4096$, and all the solvers have a similar running time for $N \geq 4096$. The difference in running time can be empirically explained by our benchmark of these SVD solvers on a rectangular matrix (size $2 \times n^2 / 2$) in \Cref{fig:svd_rectangle}, and on a square matrix (size $n \times n$) in \Cref{fig:svd_square}. These matrix sizes correspond to the size of submatrices on which an SVD is performed in the hierarchical algorithm. We indeed observe that the full SVD with LAPACK is faster than partial SVDs with ARPACK, PROPACK, LOBPCG for lower matrix size.
	
	\begin{figure}[!t]
		\centering
		\begin{subfigure}[!t]{0.48\textwidth}
			\centering
			\includegraphics[width=\textwidth]{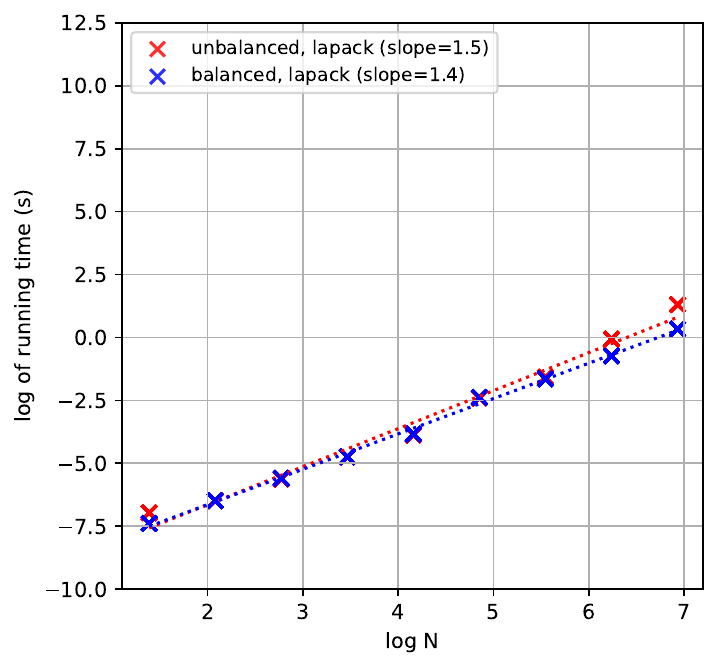}
			\caption{Small matrix size $N \times N$}
			\label{fig:low_dim}
		\end{subfigure}%
		~ 
		\begin{subfigure}[!t]{0.48\textwidth}
			\centering
			\includegraphics[width=\textwidth]{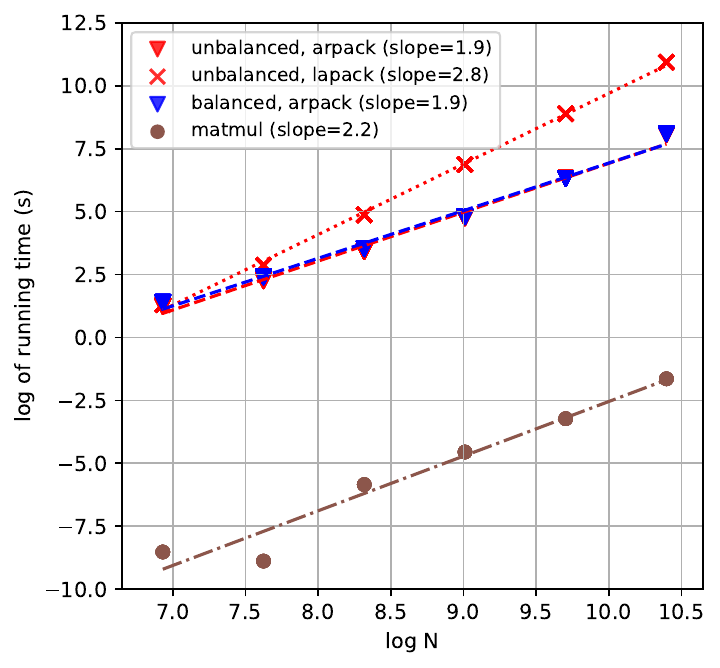}
			\caption{Large matrix size $N \times N$}
			\label{fig:large_dim}
		\end{subfigure}%
		\caption{Comparing balanced and unbalanced trees - running time of our implementation of \Cref{algo:butterfly-fact}, in logarithmic scale. For better visualization a least-square regression is performed for each set of measurements, and we report the estimated slope $a$ of the regression line $Y = a X + b$. For slope comparison, we measure the running time for matrix-vector multiplication, computed with \texttt{numpy.matmul} method from \texttt{Numpy 1.22.3}. 
		}
		\label{fig:experience}
	\end{figure}

	\paragraph{Comparing balanced and unbalanced tree}
	We now compare the running time of the hierarchical factorization algorithm with the unbalanced and balanced tree. In \Cref{fig:low_dim}, the comparison is performed in a setting with small matrix size with $N \leq 1024$, using the LAPACK solver to compute full SVDs. Hierarchical factorization is slightly faster with a balanced tree compared to an unbalanced tree.
	In \Cref{fig:large_dim}, the comparison is performed in a setting with large matrix size with $N \geq 1024$. When using the ARPACK solver, the running time for hierarchical factorization is the same for both trees. This is coherent with our analysis of complexity bounds for \Cref{algo:butterfly-fact}, which is $\mathcal{O}(N^2)$ for both trees. 
	For completeness \Cref{fig:large_dim} also compares the running time of hierarchical factorization with the one of matrix-vector multiplication for a matrix of size $N \times N$, whose complexity is also $\mathcal{O}(N^2)$. Finally we see in \Cref{fig:large_dim} that using an unbalanced tree with the LAPACK solver for large matrix size, as done in experiments of \cite{le:hal-03438881}, is not optimal. This explains why it was observed in \cite{le:hal-03438881} that a balanced tree is preferable to an unbalanced one, while the finer theoretical and computational analysis performed here leads to a different conclusion.
	
	\section{Conclusion}
	\label{section:conclusion}
	We established hierarchical identifiability in the butterfly sparse matrix factorization.
	We proved that the butterfly factors $(\matseq{\mat{X}}{\ell})_{\ell=1}^J$ of $\mat{Z} := \matseq{\mat{X}}{1} \ldots \matseq{\mat{X}}{J}$ can be recovered up to scaling ambiguities from $\mat{Z}$ with a hierarchical factorization method, described by \cref{algo:butterfly-fact}, which is endowed with exact recovery guarantees, and has controlled time complexity of $\mathcal{O}(N^2)$. We now provide some interesting future research directions.
		
	\paragraph{Stability results}
	As discussed in \Cref{remark:algo-icassp}, a challenge is to explore stability properties of the matrix factorization problem with a butterfly structure, to ensure robustness of \Cref{algo:butterfly-fact} with respect to noise. Such guarantees would justify why 
	\Cref{algo:butterfly-fact} still performs well in some noisy settings, as shown empirically in \cite{le:hal-03438881}.
	As the balanced and unbalanced variants of the proposed hierarchical method have similar computation times, it would be interesting to determine if their stability to noise is also similar.
	
	\paragraph{Implementation of the hierarchical algorithm}
	Our benchmark of different solvers of SVD in \Cref{fig:svd_rectangle,fig:svd_square} suggests that our implementation of the hierarchical algorithm can be further improved by choosing the optimal SVD solver depending on the block's dimension.
	One could also envision randomized SVDs in the spirit of the butterfly algorithm \cite{li2015butterfly}.
	\paragraph{Recovering latent sparse butterfly factors}
	When a data matrix is generated from the product of some latent butterfly factors, our identifiability results for butterfly sparse matrix factorization (\Cref{thm:identifiability-multilayer-butterfly}) could be useful to recover with minimal ambiguity these latent factors. One example of such a setting is neural network distillation, where a dense pre-trained teacher network, assumed to involve latent butterfly factors, is distilled into a sparse student network with butterfly structure. In this scenario, identifying the student butterfly factors can be based on sampling of the teacher network's derivatives, in the spirit of \cite{fornasier2021robust, fornasier2022robust}.
	
	\paragraph{Identifiability results for $J=2$ factors}
	Our analysis of identifiability using the hierarchical approach relies on identifiability results in the case with $J=2$ factors. This motivates further explorations of identifiability conditions in this setting.
	Our work focuses on fixed-support constraints, and we considered in \Cref{prop:disjoint-rank-one-supports-unique-EMF} the case of disjoint rank-one supports. 
	More relaxed conditions can be envisioned, such as supports satisfying the so-called \emph{complete equivalence class} condition introduced in \cite{papierTung}. Beyond the fixed-support setting, one can also explore essential uniqueness by considering a \emph{family} of sparsity patterns like in \cite{LeonPart1}.
	One can then use results from matrix completion literature \cite{eldar2012uniqueness, kiraly2015algebraic, cosse2020stable} to establish more elaborate conditions for identifiability in the case with two factors.

	\section*{Acknowledgments}
	The authors thank Hakim Hadj-Djilani for his valuable help in our numerical experiments and the implementation of \Cref{algo:butterfly-fact} in the FAµST 3.25 toolbox. 
	The authors gratefully acknowledge the support of the Centre Blaise Pascal's IT test platform at ENS de Lyon  for Machine Learning facilities. The platform operates the SIDUS solution \cite{sidus} developed by Emmanuel Quemener.
	The authors also thank the anonymous reviewers for the insightful comments and suggestions during the revision of the manuscript.
	
	\bibliographystyle{siamplain}
	
	\bibliography{references}

\begin{thebibliography}{10}

\bibitem{ahmed2013blind}
{\sc A.~Ahmed, B.~Recht, and J.~Romberg}, {\em Blind deconvolution using convex
  programming}, IEEE Transactions on Information Theory, 60 (2013),
  pp.~1711--1732, \url{https://doi.org/10.1109/TIT.2013.2294644}.

\bibitem{bahmani2015lifting}
{\sc S.~Bahmani and J.~Romberg}, {\em Lifting for blind deconvolution in random
  mask imaging: Identifiability and convex relaxation}, SIAM Journal on Imaging
  Sciences, 8 (2015), pp.~2203--2238, \url{https://doi.org/10.1137/141002165}.

\bibitem{candes2009fast}
{\sc E.~Candes, L.~Demanet, and L.~Ying}, {\em A fast butterfly algorithm for
  the computation of {Fourier} integral operators}, Multiscale Modeling \&
  Simulation, 7 (2009), pp.~1727--1750,
  \url{https://doi.org/10.1137/080734339}.

\bibitem{candes2015phase}
{\sc E.~J. Candes, Y.~C. Eldar, T.~Strohmer, and V.~Voroninski}, {\em Phase
  retrieval via matrix completion}, SIAM review, 57 (2015), pp.~225--251,
  \url{https://doi.org/10.1137/151005099}.

\bibitem{candes2013phaselift}
{\sc E.~J. Candes, T.~Strohmer, and V.~Voroninski}, {\em Phaselift: Exact and
  stable signal recovery from magnitude measurements via convex programming},
  Communications on Pure and Applied Mathematics, 66 (2013), pp.~1241--1274,
  \url{https://doi.org/10.1002/cpa.21432}.

\bibitem{chen2022pixelated}
{\sc B.~Chen, T.~Dao, K.~Liang, J.~Yang, Z.~Song, A.~Rudra, and C.~Re}, {\em
  Pixelated butterfly: Simple and efficient sparse training for neural network
  models}, in International Conference on Learning Representations, 2022,
  \url{https://openreview.net/forum?id=Nfl-iXa-y7R}.

\bibitem{choromanski2019unifying}
{\sc K.~Choromanski, M.~Rowland, W.~Chen, and A.~Weller}, {\em Unifying
  orthogonal {Monte} {Carlo} methods}, in International Conference on Machine
  Learning, PMLR, 2019, pp.~1203--1212,
  \url{https://proceedings.mlr.press/v97/choromanski19a/choromanski19a.pdf}.

\bibitem{choudhary2014identifiability}
{\sc S.~Choudhary and U.~Mitra}, {\em Identifiability scaling laws in bilinear
  inverse problems}, arXiv preprint arXiv:1402.2637,  (2014),
  \url{https://arxiv.org/abs/1402.2637}.

\bibitem{choudhary2014sparse}
{\sc S.~Choudhary and U.~Mitra}, {\em Sparse blind deconvolution: What cannot
  be done}, in 2014 IEEE International Symposium on Information Theory, IEEE,
  2014, pp.~3002--3006, \url{https://doi.org/10.1109/ISIT.2014.6875385}.

\bibitem{cosse2020stable}
{\sc A.~Cosse and L.~Demanet}, {\em {Stable rank-one matrix completion is
  solved by the level 2 Lasserre relaxation}}, Foundations of Computational
  Mathematics,  (2020), pp.~1--50,
  \url{https://doi.org/10.1007/s10208-020-09471-y}.

\bibitem{dao2022monarch}
{\sc T.~Dao, B.~Chen, N.~S. Sohoni, A.~D. Desai, M.~Poli, J.~Grogan, A.~Liu,
  A.~Rao, A.~Rudra, and C.~R{\'{e}}}, {\em Monarch: Expressive structured
  matrices for efficient and accurate training}, in International Conference on
  Machine Learning, {PMLR}, 2022, pp.~4690--4721,
  \url{https://proceedings.mlr.press/v162/dao22a.html}.

\bibitem{dao2019learning}
{\sc T.~Dao, A.~Gu, M.~Eichhorn, A.~Rudra, and C.~R{\'e}}, {\em Learning fast
  algorithms for linear transforms using butterfly factorizations}, in
  International Conference on Machine Learning, PMLR, 2019, pp.~1517--1527,
  \url{http://proceedings.mlr.press/v97/dao19a/dao19a.pdf}.

\bibitem{dao2020kaleidoscope}
{\sc T.~Dao, N.~Sohoni, A.~Gu, M.~Eichhorn, A.~Blonder, M.~Leszczynski,
  A.~Rudra, and C.~Ré}, {\em Kaleidoscope: An efficient, learnable
  representation for all structured linear maps}, in International Conference
  on Learning Representations, 2020,
  \url{https://openreview.net/forum?id=BkgrBgSYDS}.

\bibitem{devlin2019bert}
{\sc J.~Devlin, M.~Chang, K.~Lee, and K.~Toutanova}, {\em {BERT:} pre-training
  of deep bidirectional transformers for language understanding}, in North
  American Chapter of the Association for Computational Linguistics: Human
  Language Technologies, Association for Computational Linguistics, 2019,
  pp.~4171--4186, \url{https://doi.org/10.18653/v1/n19-1423},
  \url{https://doi.org/10.18653/v1/n19-1423}.

\bibitem{dosovitskiy2021an}
{\sc A.~Dosovitskiy, L.~Beyer, A.~Kolesnikov, D.~Weissenborn, X.~Zhai,
  T.~Unterthiner, M.~Dehghani, M.~Minderer, G.~Heigold, S.~Gelly, J.~Uszkoreit,
  and N.~Houlsby}, {\em An image is worth 16x16 words: Transformers for image
  recognition at scale}, in International Conference on Learning
  Representations, 2021, \url{https://openreview.net/forum?id=YicbFdNTTy}.

\bibitem{elad2010sparse}
{\sc M.~Elad}, {\em Sparse and redundant representations: from theory to
  applications in signal and image processing}, Springer Science \& Business
  Media, 2010, \url{https://doi.org/10.1007/978-1-4419-7011-4}.

\bibitem{eldar2012uniqueness}
{\sc Y.~C. Eldar, D.~Needell, and Y.~Plan}, {\em Uniqueness conditions for
  low-rank matrix recovery}, Applied and Computational Harmonic Analysis, 33
  (2012), pp.~309--314, \url{https://doi.org/10.1016/j.acha.2012.04.002}.

\bibitem{fornasier2022robust}
{\sc M.~Fornasier, T.~Klock, and M.~Rauchensteiner}, {\em Robust and
  resource-efficient identification of two hidden layer neural networks},
  Constructive Approximation, 55 (2022), pp.~475--536,
  \url{https://doi.org/10.1007/s00365-021-09550-5}.

\bibitem{fornasier2021robust}
{\sc M.~Fornasier, J.~Vyb{\'\i}ral, and I.~Daubechies}, {\em Robust and
  resource efficient identification of shallow neural networks by fewest
  samples}, Information and Inference: A Journal of the IMA, 10 (2021),
  pp.~625--695, \url{https://doi.org/10.1093/imaiai/iaaa036}.

\bibitem{foucart_mathematical_2013}
{\sc S.~Foucart and H.~Rauhut}, {\em A mathematical introduction to compressive
  sensing}, Applied and Numerical Harmonic Analysis, Birkhäuser Basel, 2013,
  \url{https://doi.org/10.1007/978-0-8176-4948-7}.

\bibitem{genz2000methods}
{\sc A.~Genz}, {\em Methods for generating random orthogonal matrices}, in
  Monte-Carlo and Quasi-Monte Carlo Methods 1998, Springer, 2000, pp.~199--213,
  \url{https://doi.org/10.1007/978-3-642-59657-5_13}.

\bibitem{gillis2020nonnegative}
{\sc N.~Gillis}, {\em Nonnegative matrix factorization}, SIAM, 2020,
  \url{https://doi.org/10.1137/1.9781611976410}.

\bibitem{hackbusch2002data}
{\sc W.~Hackbusch and S.~B{\"o}rm}, {\em Data-sparse approximation by adaptive
  {H}2-matrices}, Computing, 69 (2002), pp.~1--35,
  \url{https://doi.org/10.1007/s00607-002-1450-4}.

\bibitem{halko2011finding}
{\sc N.~Halko, P.-G. Martinsson, and J.~A. Tropp}, {\em Finding structure with
  randomness: Probabilistic algorithms for constructing approximate matrix
  decompositions}, SIAM review, 53 (2011), pp.~217--288,
  \url{https://doi.org/10.1137/090771806}.

\bibitem{jing2017tunable}
{\sc L.~Jing, Y.~Shen, T.~Dubcek, J.~Peurifoy, S.~Skirlo, Y.~LeCun, M.~Tegmark,
  and M.~Solja{\v{c}}i{\'c}}, {\em Tunable efficient unitary neural networks
  and their application to {RNN}s}, in International Conference on Machine
  Learning, PMLR, 2017, pp.~1733--1741,
  \url{https://proceedings.mlr.press/v70/jing17a/jing17a.pdf}.

\bibitem{kech2017optimal}
{\sc M.~Kech and F.~Krahmer}, {\em Optimal injectivity conditions for bilinear
  inverse problems with applications to identifiability of deconvolution
  problems}, SIAM Journal on Applied Algebra and Geometry, 1 (2017),
  pp.~20--37, \url{https://doi.org/10.1137/16M1067469}.

\bibitem{kiraly2015algebraic}
{\sc F.~J. Kir{\'a}ly, L.~Theran, and R.~Tomioka}, {\em The algebraic
  combinatorial approach for low-rank matrix completion.}, J. Mach. Learn.
  Res., 16 (2015), pp.~1391--1436,
  \url{https://dl.acm.org/doi/10.5555/2789272.2886794}.

\bibitem{kruskal1977three}
{\sc J.~B. Kruskal}, {\em Three-way arrays: rank and uniqueness of trilinear
  decompositions, with application to arithmetic complexity and statistics},
  Linear algebra and its applications, 18 (1977), pp.~95--138,
  \url{https://doi.org/10.1016/0024-3795(77)90069-6}.

\bibitem{le2013fastfood}
{\sc Q.~Le, T.~Sarlos, and A.~Smola}, {\em Fastfood - computing hilbert space
  expansions in loglinear time}, in International Conference on Machine
  Learning, PMLR, 2013, pp.~244--252,
  \url{https://proceedings.mlr.press/v28/le13.html}.

\bibitem{le2021structured}
{\sc Q.-T. Le and R.~Gribonval}, {\em Structured support exploration for
  multilayer sparse matrix factorization}, in ICASSP 2021-2021 IEEE
  International Conference on Acoustics, Speech and Signal Processing (ICASSP),
  IEEE, 2021, pp.~3245--3249,
  \url{https://doi.org/10.1109/ICASSP39728.2021.9414238}.

\bibitem{papierTung}
{\sc Q.-T. Le, E.~Riccietti, and R.~Gribonval}, {\em {Spurious Valleys,
  Spurious Minima and NP-hardness of Sparse Matrix Factorization With Fixed
  Support}}.
\newblock preprint, 2022, \url{https://hal.inria.fr/hal-03364668}.

\bibitem{le:hal-03438881}
{\sc Q.-T. Le, L.~Zheng, E.~Riccietti, and R.~Gribonval}, {\em {Fast learning
  of fast transforms, with guarantees}}, in {ICASSP, IEEE International
  Conference on Acoustics, Speech and Signal Processing}, Singapore, Singapore,
  May 2022, \url{https://hal.inria.fr/hal-03438881}.

\bibitem{le2016matrices}
{\sc L.~Le~Magoarou}, {\em Matrices efficientes pour le traitement du signal et
  l'apprentissage automatique}, PhD thesis, INSA de Rennes, 2016,
  \url{https://hal.inria.fr/tel-01412558}.
\newblock Written in French.

\bibitem{7178579}
{\sc L.~Le~Magoarou and R.~Gribonval}, {\em Chasing butterflies: In search of
  efficient dictionaries}, in 2015 IEEE International Conference on Acoustics,
  Speech and Signal Processing (ICASSP), 2015, pp.~3287--3291,
  \url{https://doi.org/10.1109/ICASSP.2015.7178579}.

\bibitem{le2016flexible}
{\sc L.~Le~Magoarou and R.~Gribonval}, {\em Flexible multilayer sparse
  approximations of matrices and applications}, IEEE Journal of Selected Topics
  in Signal Processing, 10 (2016), pp.~688--700,
  \url{https://doi.org/10.1109/JSTSP.2016.2543461}.

\bibitem{li2013sparse}
{\sc X.~Li and V.~Voroninski}, {\em Sparse signal recovery from quadratic
  measurements via convex programming}, SIAM Journal on Mathematical Analysis,
  45 (2013), pp.~3019--3033, \url{https://doi.org/10.1137/120893707}.

\bibitem{li2016identifiability_bis}
{\sc Y.~Li, K.~Lee, and Y.~Bresler}, {\em Identifiability in bilinear inverse
  problems with applications to subspace or sparsity-constrained blind gain and
  phase calibration}, IEEE Transactions on Information Theory, 63 (2016),
  pp.~822--842, \url{https://doi.org/10.1109/TIT.2016.2637933}.

\bibitem{li2016identifiability}
{\sc Y.~Li, K.~Lee, and Y.~Bresler}, {\em Identifiability in blind
  deconvolution with subspace or sparsity constraints}, IEEE Transactions on
  information Theory, 62 (2016), pp.~4266--4275,
  \url{https://doi.org/10.1109/TIT.2016.2569578}.

\bibitem{li2017identifiability}
{\sc Y.~Li, K.~Lee, and Y.~Bresler}, {\em Identifiability and stability in
  blind deconvolution under minimal assumptions}, IEEE Transactions on
  Information Theory, 63 (2017), pp.~4619--4633,
  \url{https://doi.org/10.1109/TIT.2017.2689779}.

\bibitem{li2015butterfly}
{\sc Y.~Li, H.~Yang, E.~R. Martin, K.~L. Ho, and L.~Ying}, {\em Butterfly
  factorization}, Multiscale Modeling \& Simulation, 13 (2015), pp.~714--732,
  \url{https://doi.org/10.1137/15M1007173}.

\bibitem{lin2021deformable}
{\sc R.~Lin, J.~Ran, K.~H. Chiu, G.~Chesi, and N.~Wong}, {\em Deformable
  butterfly: A highly structured and sparse linear transform}, Advances in
  Neural Information Processing Systems, 34 (2021), pp.~16145--16157,
  \url{https://proceedings.neurips.cc/paper/2021/file/86b122d4358357d834a87ce618a55de0-Paper.pdf}.

\bibitem{ling2015self}
{\sc S.~Ling and T.~Strohmer}, {\em Self-calibration and biconvex compressive
  sensing}, Inverse Problems, 31 (2015), p.~115002,
  \url{https://doi.org/10.1088/0266-5611/31/11/115002}.

\bibitem{malgouyres2020stable}
{\sc F.~Malgouyres}, {\em On the stable recovery of deep structured linear
  networks under sparsity constraints}, in Mathematical and Scientific Machine
  Learning, PMLR, 2020, pp.~107--127,
  \url{http://proceedings.mlr.press/v107/malgouyres20a/malgouyres20a.pdf}.

\bibitem{malgouyres2016identifiability}
{\sc F.~Malgouyres and J.~Landsberg}, {\em On the identifiability and stable
  recovery of deep/multi-layer structured matrix factorization}, in 2016 IEEE
  Information Theory Workshop (ITW), IEEE, 2016, pp.~315--319,
  \url{https://doi.org/10.1109/ITW.2016.7606847}.

\bibitem{malgouyres2019multilinear}
{\sc F.~Malgouyres and J.~Landsberg}, {\em Multilinear compressive sensing and
  an application to convolutional linear networks}, SIAM Journal on Mathematics
  of Data Science, 1 (2019), pp.~446--475,
  \url{https://doi.org/10.1137/18M119834X}.

\bibitem{martinsson2011fast}
{\sc P.-G. Martinsson}, {\em A fast randomized algorithm for computing a
  hierarchically semiseparable representation of a matrix}, SIAM Journal on
  Matrix Analysis and Applications, 32 (2011), pp.~1251--1274,
  \url{https://doi.org/10.1137/100786617}.

\bibitem{mathieu2014fast}
{\sc M.~Mathieu and Y.~LeCun}, {\em Fast approximation of rotations and
  {Hessians} matrices}, arXiv preprint arXiv:1404.7195,  (2014),
  \url{https://arxiv.org/abs/1404.7195}.

\bibitem{michielssen1996multilevel}
{\sc E.~Michielssen and A.~Boag}, {\em A multilevel matrix decomposition
  algorithm for analyzing scattering from large structures}, IEEE Transactions
  on Antennas and Propagation, 44 (1996), pp.~1086--1093,
  \url{https://doi.org/10.1109/8.511816}.

\bibitem{munkhoeva2018quadrature}
{\sc M.~Munkhoeva, Y.~Kapushev, E.~Burnaev, and I.~Oseledets}, {\em
  Quadrature-based features for kernel approximation}, Advances in neural
  information processing systems, 31 (2018),
  \url{https://proceedings.neurips.cc/paper/2018/file/6e923226e43cd6fac7cfe1e13ad000ac-Paper.pdf}.

\bibitem{o2010algorithm}
{\sc M.~O'Neil, F.~Woolfe, and V.~Rokhlin}, {\em An algorithm for the rapid
  evaluation of special function transforms}, Applied and Computational
  Harmonic Analysis, 28 (2010), pp.~203--226,
  \url{https://doi.org/10.1016/j.acha.2009.08.005}.

\bibitem{pan2001structured}
{\sc V.~Pan}, {\em Structured matrices and polynomials: unified superfast
  algorithms}, Springer Science \& Business Media, 2001,
  \url{https://doi.org/10.1007/978-1-4612-0129-8}.

\bibitem{parikh2014proximal}
{\sc N.~Parikh, S.~Boyd, et~al.}, {\em Proximal algorithms}, Foundations and
  trends{\textregistered} in Optimization, 1 (2014), pp.~127--239,
  \url{https://doi.org/10.1561/2400000003}.

\bibitem{Parker95randombutterfly}
{\sc D.~S. Parker}, {\em Random butterfly transformations with applications in
  computational linear algebra}, tech. report, 1995,
  \url{https://citeseerx.ist.psu.edu/viewdoc/download?doi=10.1.1.17.7932&rep=rep1&type=pdf}.

\bibitem{sidus}
{\sc E.~Quemener and M.~Corvellec}, {\em Sidus—the solution for extreme
  deduplication of an operating system}, Linux J., 2013 (2013),
  \url{https://dl.acm.org/doi/abs/10.5555/2555789.2555792}.

\bibitem{5325694}
{\sc R.~Rubinstein, M.~Zibulevsky, and M.~Elad}, {\em Double sparsity: Learning
  sparse dictionaries for sparse signal approximation}, IEEE Transactions on
  Signal Processing, 58 (2010), pp.~1553--1564,
  \url{https://doi.org/10.1109/TSP.2009.2036477}.

\bibitem{tolstikhin2021mlp}
{\sc I.~O. Tolstikhin, N.~Houlsby, A.~Kolesnikov, L.~Beyer, X.~Zhai,
  T.~Unterthiner, J.~Yung, A.~Steiner, D.~Keysers, J.~Uszkoreit, et~al.}, {\em
  {MLP}-mixer: An all-{MLP} architecture for vision}, Advances in Neural
  Information Processing Systems, 34 (2021), pp.~24261--24272,
  \url{https://proceedings.neurips.cc/paper/2021/file/cba0a4ee5ccd02fda0fe3f9a3e7b89fe-Paper.pdf}.

\bibitem{DictionaryLearning}
{\sc I.~{Tošić} and P.~{Frossard}}, {\em Dictionary learning}, IEEE Signal
  Processing Magazine, 28 (2011), pp.~27--38,
  \url{https://doi.org/10.1109/MSP.2010.939537}.

\bibitem{Kronecker}
{\sc C.~F. Van~Loan}, {\em The ubiquitous {Kronecker} product}, Journal of
  Computational and Applied Mathematics, 123 (2000), pp.~85--100,
  \url{https://doi.org/10.1016/s0377-0427(00)00393-9}.

\bibitem{wolfgang1999sparse}
{\sc H.~Wolfgang}, {\em A sparse matrix arithmetic based on {H}-matrices.
  {Part} {I}: Introduction to h-matrices}, Computing, 62 (1999), pp.~89--108,
  \url{https://doi.org/10.1007/s006070050015}.

\bibitem{woolfe2008fast}
{\sc F.~Woolfe, E.~Liberty, V.~Rokhlin, and M.~Tygert}, {\em A fast randomized
  algorithm for the approximation of matrices}, Applied and Computational
  Harmonic Analysis, 25 (2008), pp.~335--366,
  \url{https://doi.org/10.1016/j.acha.2007.12.002}.

\bibitem{ye2016every}
{\sc K.~Ye and L.-H. Lim}, {\em Every matrix is a product of {Toeplitz}
  matrices}, Foundations of Computational Mathematics, 16 (2016), pp.~577--598,
  \url{https://doi.org/10.1007/s10208-015-9254-z}.

\bibitem{LeonPart1}
{\sc L.~Zheng, E.~Riccietti, and R.~Gribonval}, {\em {Identifiability in
  Two-Layer Sparse Matrix Factorization}}, arXiv preprint arXiv:2110.01235,
  (2021), \url{https://arxiv.org/abs/2110.01235}.

\bibitem{zheng:hal-03620052v1}
{\sc L.~Zheng, E.~Riccietti, and R.~Gribonval}, {\em {Code for reproducible
  research - Efficient Identification of Butterfly Sparse Matrix
  Factorizations}}, Mar. 2022, \url{https://hal.inria.fr/hal-03620052}.
\newblock Code repository, updates at
  \url{https://github.com/leonzheng2/efficient-butterfly}.

\end{thebibliography}
	
	\appendix
	
	\section{Proof of \Cref{lemma:identical-colsupp-maximal-colsupp}}
\label{app:identical-colsupp-maximal-colsupp}
We begin with a simple lemma \cite[Chapter 3]{gillis2020nonnegative}. 
\begin{lemma}
	\label{lemma:nec-conditions-EMF}
	Let $\Sigma$ be any set of pairs of factors, and $(\mat{X}, \mat{Y}) \in \Sigma$. We have
	\begin{displaymath}
		(\mat{X}, \mat{Y}) \in \unique(\Sigma) \iff (\mat{X}, \mat{Y}) \in \bigcap_{\substack{\Sigma' \subseteq \Sigma \\ (\mat{X}, \mat{Y}) \in \Sigma'}} \unique(\Sigma').
	\end{displaymath}
\end{lemma}

\begin{proof}
	Let $(\mat{X}, \mat{Y}) \in \unique(\Sigma)$, and consider $\Sigma' \subseteq \Sigma$ such that $(\mat{X}, \mat{Y})  \in \Sigma'$, as well as $(\bar{\mat{X}}, \bar{\mat{Y}}) \in \Sigma'$ such that $\bar{\mat{X}} \transpose{\bar{\mat{Y}}} = \mat{X} \transpose{\mat{Y}}$. Since $\Sigma' \subseteq \Sigma$, we have $(\bar{\mat{X}}, \bar{\mat{Y}}) \in \Sigma$, and because $(\mat{X}, \mat{Y}) \in \unique(\Sigma)$, $(\bar{\mat{X}}, \bar{\mat{Y}}) \sim (\mat{X}, \mat{Y})$. 
	Moreover, $(\mat{X}, \mat{Y}) \in \Sigma'$, hence $(\mat{X}, \mat{Y}) \in \unique(\Sigma')$. This is true for every $\Sigma' \subseteq \Sigma$, proving one implication. 
	The converse is true considering the case $\Sigma' := \Sigma$.
\end{proof}

\Cref{lemma:identical-colsupp-maximal-colsupp} is then derived from the following trivial but crucial observation.

\begin{lemma}
	\label{lemma:on-met-ce-qu-on-veut-a-droite}
	Let $\Sigma$ be any set of pairs of factors, and $\mat{X}$, $\mat{Y}, \bar{\mat{Y}}$ such that $(\mat{X}, \mat{Y}), (\mat{X}, \bar{\mat{Y}}) \in \Sigma$. If $\mat{X}\transpose{\mat{Y}}=\mat{X}\transpose{\bar{\mat{Y}}}$ and $\colsupp(\bar{\mat{Y}})$, $\colsupp(\mat{Y})$ do not have the same cardinality, then $(\mat{X}, \mat{Y}) \not\sim (\mat{X}, \bar{\mat{Y}})$ and hence $(\mat{X}, \mat{Y}) \notin \unique(\Sigma)$ and $(\mat{X}, \bar{\mat{Y}}) \notin \unique(\Sigma)$. This is true in particular if there exists an index $i \in \integerSet{r}$ for which $\matcol{\mat{X}}{i} = \vctor{0}$, $\matcol{\mat{Y}}{i} = \vctor{0}$, $\matcol{\bar{\mat{Y}}}{i} \neq \vctor{0}$, and $\matcol{\mat{Y}}{j} = \matcol{\bar{\mat{Y}}}{j}$ for all $j \neq i$.
\end{lemma}

\begin{proof}[Proof of \Cref{lemma:identical-colsupp-maximal-colsupp}]
	We first show $\unique(\Sigma_{\pair{S}}) \subseteq \idcolsupp{\pair{S}}$ by contraposition. Let $(\mat{X}, \mat{Y}) \in \Sigma_{\pair{S}}$, and suppose that $\colsupp(\mat{X}) \neq \colsupp(\mat{Y})$. Up to matrix transposition, we can suppose without loss of generality that $\colsupp(\mat{Y})$ is not a subset of $\colsupp(\mat{X})$, so there is $i \in \integerSet{r}$ such that $\matcol{\mat{X}}{i} = \vctor{0}$ and $\matcol{\mat{Y}}{i} \neq \vctor{0}$. 
	Define  $\bar{\mat{Y}}$ a right factor such that $\matcol{\bar{\mat{Y}}}{i} = \vctor{0}$ and $\submat{\bar{\mat{Y}}}{\integerSet{r} \backslash \{ i \}} = \submat{\mat{Y}}{\integerSet{r} \backslash \{ i \}}$
	\footnote{By abuse of notations, $\submat{\mat{M}}{I}$ is the submatrix of $\mat{M} \in \mathbb{C}^{m \times n}$ composed of columns of $\mat{M}$ indexed by $I \subset \integerSet{n}$.}. 
	By construction, $\supp(\bar{\mat{Y}}) \subseteq \supp(\mat{Y})$, so $(\mat{X}, \bar{\mat{Y}}) \in \Sigma_{\pair{S}}$.
	Applying \Cref{lemma:on-met-ce-qu-on-veut-a-droite} to $\Sigma = \Sigma_{\pair{S}}$, we obtain $(\mat{X}, \mat{Y}) \notin \unique(\Sigma_{\pair{S}})$.
		We now show $\unique(\Sigma_{\pair{S}}) \subseteq \maxcolsupp{\pair{S}}$, also by contraposition. Let $(\mat{X}, \mat{Y}) \notin \maxcolsupp{\pair{S}}$, and assume $\colsupp(\mat{X}) \neq \colsupp(\leftsupp{S})$. The reasoning would be symmetric if we supposed $\colsupp(\mat{Y}) \neq \colsupp(\rightsupp{S})$. 
	By the previously shown inclusion $\unique(\Sigma_{\pair{S}}) \subseteq \idcolsupp{\pair{S}}$, we also assume $\colsupp(\mat{X}) = \colsupp(\mat{Y})$ without loss of generality. To conclude we treat three cases:
	\begin{itemize}
		\item If $\colsupp(\rightsupp{S}) \not\subseteq \colsupp(\leftsupp{S})$ then we can fix  $i \in \integerSet{r}$ such that $\matcol{\leftsupp{S}}{i} = \vctor{0}$ and $\matcol{\rightsupp{S}}{i} \neq \vctor{0}$. This means $\matcol{\mat{X}}{i} = \matcol{\mat{Y}}{i} = \vctor{0}$. Setting $\bar{\mat{Y}} \in \Sigma_{\rightsupp{S}}$ such that $\matcol{\bar{\mat{Y}}}{i} = \matcol{\rightsupp{S}}{i}$ and $\submat{\bar{\mat{Y}}}{\integerSet{r} \backslash \{ i \}} = \submat{\mat{Y}}{\integerSet{r} \backslash \{ i \}} $, we build an instance as in 
		\Cref{lemma:on-met-ce-qu-on-veut-a-droite} with $\Sigma = \Sigma_{\pair{S}}$, to show $(\mat{X}, \mat{Y}) \notin \unique(\Sigma_{\pair{S}})$.
		\item If $\colsupp(\leftsupp{S}) \not\subseteq \colsupp(\rightsupp{S})$, then the same arguments yield $(\mat{X}, \mat{Y}) \notin \unique(\Sigma_{\pair{S}})$.
		\item There remains the case $\colsupp(\leftsupp{S}) = \colsupp(\rightsupp{S})$. Since $\colsupp(\mat{X}) \subsetneq \colsupp(\leftsupp{S})$, let us fix $i \in \colsupp(\leftsupp{S})$ such that $\matcol{\mat{X}}{i} = \vctor{0}$. Then, $\matcol{\leftsupp{S}}{i} \neq 0$, $\matcol{\rightsupp{S}}{i} \neq 0$, and $\matcol{\mat{X}}{i} = \matcol{\mat{Y}}{i} = \vctor{0}$. Again, we construct $\bar{\mat{Y}} \in \Sigma_{\rightsupp{S}}$ with $\matcol{\bar{\mat{Y}}}{i} = \matcol{\rightsupp{S}}{i}$, $\submat{\bar{\mat{Y}}}{\integerSet{r} \backslash \{ i \}} = \submat{\mat{Y}}{\integerSet{r} \backslash \{ i \}} $, and we obtain an instance as in \Cref{lemma:on-met-ce-qu-on-veut-a-droite}
		with $\Sigma = \Sigma_{\pair{S}}$, showing that $(\mat{X}, \mat{Y}) \notin \unique(\Sigma_{\pair{S}})$.
	\end{itemize}
\end{proof}

The following key proposition will be useful in the lifting approach presented below.

\begin{proposition}
	\label{prop:non-degenerate-properties}
	For any pair of supports $\pair{S}$, we have: $\unique (\Sigma_{\pair{S}}) = \unique (\idcolsupp{\pair{S}}) \cap \maxcolsupp{\pair{S}}$.
\end{proposition}

\begin{proof}
	The direct inclusion is immediate by \Cref{lemma:nec-conditions-EMF,lemma:identical-colsupp-maximal-colsupp}. 		
	For the inverse inclusion, let $(\mat{X^*}, \mat{Y^*}) \in \unique (\idcolsupp{\pair{S}}) \cap \maxcolsupp{\pair{S}}$, and $(\mat{X}, \mat{Y}) \in \Sigma_{\pair{S}}$ such that $\mat{X} \transpose{\mat{Y}} = \mat{X^*} \transpose{\mat{Y^*}}$. The goal is to show $(\mat{X}, \mat{Y}) \sim (\mat{X^*}, \mat{Y^*})$.
	Denote $J = \colsupp(\mat{X}) \cap \colsupp(\mat{Y})$. 
	Define $(\bar{\mat{X}}, \bar{\mat{Y}}) \in \idcolsupp{\pair{S}}$ such that $(\submat{\bar{\mat{X}}}{J}, \submat{\bar{\mat{Y}}}{J}) = (\submat{\mat{X}}{J}, \submat{\mat{Y}}{J})$ and $(\submat{\bar{\mat{X}}}{\integerSet{r} \backslash J}, \submat{\bar{\mat{Y}}}{\integerSet{r} \backslash J}) = (\vctor{0}, \vctor{0})$. 
	Since $\bar{\mat{X}} \transpose{\bar{\mat{Y}}} = \mat{X} \transpose{\mat{Y}} = \mat{X^*} \transpose{\mat{Y^*}}$, and $(\mat{X^*}, \mat{Y^*}) \in \unique(\idcolsupp{\pair{S}})$, we have $(\bar{\mat{X}}, \bar{\mat{Y}}) \sim (\mat{X^*}, \mat{Y^*})$. 
	But $\colsupp(\mat{X}^*) = \colsupp(\leftsupp{S})$ and $\colsupp(\mat{Y}^*) = \colsupp(\rightsupp{S})$, because $(\mat{X^*}, \mat{Y^*}) \in \maxcolsupp{\pair{S}}$. Hence, $J = \colsupp(\bar{\mat{X}}) = \colsupp(\mat{X}^*) = \colsupp(\leftsupp{S})$ and similarly $J = \colsupp(\rightsupp{S})$. This necessarily yields  $\bar{\mat{X}} = \mat{X}$ and $\bar{\mat{Y}} = \mat{Y}$. In conclusion, $(\mat{X}, \mat{Y}) = (\bar{\mat{X}}, \bar{\mat{Y}}) \sim (\mat{X^*}, \mat{Y^*})$.
\end{proof}

\section{Lifting procedure} 
\label{app:lifting}
	We start by claiming two technical lemmas, whose proof is left to the reader.
	Recall that $\idcolsupp{\pair{S}}$, $\maxcolsupp{\pair{S}}$ and $\Gamma_{\tuplerkone{S}}$ are defined in~\eqref{eq:DefIC},~\eqref{eq:DefMC}, and~\eqref{eq:gamma-s}.

	\begin{lemma}
		\label{lemma:rk-contrib-non-deg}
		Let $\pair{S}$ be any pair of supports satisfying $\colsupp(\leftsupp{S}) = \colsupp(\rightsupp{S})$, and denote $\tuplerkone{S} := \varphi(\pair{S})$. Then: $(\mat{X}, \mat{Y}) \in \idcolsupp{\pair{S}} \cap \maxcolsupp{\pair{S}} \iff \varphi(\mat{X}, \mat{Y}) \in \maxrankonesupp{\tuplerkone{S}}$, where $\maxrankonesupp{\tuplerkone{S}}$ denotes the set of tuples $\tuplerkone{C} \in \Gamma_{\tuplerkone{S}}$ with ``maximal" index support\footnote{By analogy with column support, the index support of $\tuplerkone{C} = (\rankone{C}{i})_{i=1}^r$ is the subset of indices $i \in \integerSet{r}$ such that $\rankone{C}{i} \neq \mat{0}$.}:
		\begin{equation}
			\maxrankonesupp{\tuplerkone{S}} := \{ \tuplerkone{C} \in \Gamma_{\tuplerkone{S}} \; | \; \forall i \in \integerSet{r}, \, \rankone{C}{i} = \mat{0} \implies \rankone{S}{i} = \mat{0} \}.
		\end{equation}

	\end{lemma}

	\begin{lemma}
	\label{lemma:varphi-properties}
	The application $\varphi$ is invariant to column rescaling: $\varphi(\mat{X}, \mat{Y}) = \varphi(\bar{\mat{X}}, \bar{\mat{Y}})$ for any equivalent pair of factors $(\mat{X}, \mat{Y}) \sim (\bar{\mat{X}}, \bar{\mat{Y}})$. Moreover, the application $\varphi$ restricted to $\idcolsupp{\pair{S}}$, denoted $\varphi_{\pair{S}} \colon \idcolsupp{\pair{S}} \to \Gamma_{\tuplerkone{S}}$ is surjective, and injective up to equivalences, in the sense that $(\mat{X}, \mat{Y}) \sim (\bar{\mat{X}}, \bar{\mat{Y}})$ for any $(\mat{X}, \mat{Y}), (\bar{\mat{X}}, \bar{\mat{Y}}) \in \idcolsupp{\pair{S}}$ such that $\varphi_{\pair{S}}(\mat{X}, \mat{Y}) = \varphi_{\pair{S}}(\bar{\mat{X}}, \bar{\mat{Y}})$.
\end{lemma}

We now define the operator that sums the $r$ matrices of a tuple $\tuplerkone{C}$ as:
$	\mathcal{A}: \tuplerkone{C} = (\rankone{C}{i})_{i=1}^r\mapsto \sum_{i=1}^r \rankone{C}{i}$.
This operator corresponds to a lifting operator in the terminology of \cite{choudhary2014identifiability}. For any set $\Gamma \subseteq (\mathbb{C}^{m \times n})^{r}$ of $r$-tuples of rank-one matrices, denote (by slight abuse of notation)
$	\unique(\Gamma) := \{ \tuplerkone{C} \in \Gamma \; | \; \forall \tuplerkone{C'} \in \Gamma, \; \mathcal{A}(\tuplerkone{C}) = \mathcal{A}(\tuplerkone{C'}) \implies \tuplerkone{C} = \tuplerkone{C'} \}$.
The previous lemmas lead to the following theorem characterizing essential uniqueness of the factorization $\mat{Z} := \mat{X} \transpose{\mat{Y}}$ in $\Sigma_{\pair{S}}$ as the identifiability of the rank-one contributions $\varphi(\mat{X}, \mat{Y})$ from $\mat{Z}$ with the constraint set $\Gamma_{\tuplerkone{S}}$.

\begin{theorem}
	\label{thm:equivalence-with-EMD}
	For any pair of supports $\pair{S}$ such that $\colsupp(\leftsupp{S}) = \colsupp(\rightsupp{S})$, 
	denoting $\tuplerkone{S} := \varphi(\pair{S})$, we have: $(\mat{X}, \mat{Y}) \in \unique(\Sigma_{\pair{S}}) \iff
	\varphi(\mat{X}, \mat{Y}) \in \unique(\Gamma_{\tuplerkone{S}}) \cap \maxrankonesupp{\tuplerkone{S}}$.
\end{theorem}

\begin{proof}
	By \Cref{lemma:varphi-properties}, one verifies that $(\mat{X}, \mat{Y}) \in \unique(\idcolsupp{\pair{S}}) \iff \varphi(\mat{X}, \mat{Y}) \in \unique(\Gamma_{\tuplerkone{S}})$ hence
	\begin{equation*}
		\begin{split}
			(\mat{X}, \mat{Y}) \in \unique(\Sigma_{\pair{S}}) &\stackrel{ \Cref{prop:non-degenerate-properties}}{\iff} (\mat{X}, \mat{Y}) \in \unique (\idcolsupp{\pair{S}}) \cap \maxcolsupp{\pair{S}} \\
			& \stackrel{ \Cref{lemma:varphi-properties}}{\iff} \varphi(\mat{X}, \mat{Y}) \in \unique(\Gamma_{\tuplerkone{S}}) \text{ and } (\mat{X}, \mat{Y}) \in \idcolsupp{\pair{S}} \cap \maxcolsupp{\pair{S}} \\
			&\stackrel{ \Cref{lemma:rk-contrib-non-deg}}{\iff} \varphi(\mat{X}, \mat{Y}) \in \unique(\Gamma_{\tuplerkone{S}}) \cap \maxrankonesupp{\tuplerkone{S}}.
		\end{split}
	\end{equation*}
\end{proof}

\begin{corollary}
	\label{cor:uniform-id}
	For any pair of supports $\pair{S}$ such that $\colsupp(\leftsupp{S}) = \colsupp(\rightsupp{S})$, 
	denoting $\tuplerkone{S} := \varphi(\pair{S})$, we have: $\unique(\Sigma_{\pair{S}}) = \idcolsupp{\pair{S}} \cap \maxcolsupp{\pair{S}} \iff \maxrankonesupp{\tuplerkone{S}} \subseteq \unique(\Gamma_{\tuplerkone{S}})$.
\end{corollary}

\begin{proof}
	Suppose $\maxrankonesupp{\tuplerkone{S}} \subseteq \unique(\Gamma_{\tuplerkone{S}})$. Let $(\mat{X}, \mat{Y}) \in \idcolsupp{\pair{S}} \cap \maxcolsupp{\pair{S}}$. 
	By \Cref{lemma:rk-contrib-non-deg}, $\varphi(\mat{X}, \mat{Y}) \in \maxrankonesupp{\tuplerkone{S}}$, which implies by assumption that $\varphi(\mat{X}, \mat{Y}) \in \unique(\Gamma_{\tuplerkone{S}}) \cap \maxrankonesupp{\tuplerkone{S}}$. By \Cref{thm:equivalence-with-EMD}, $(\mat{X}, \mat{Y}) \in \unique(\Sigma_{\pair{S}})$.
	This shows $\idcolsupp{\pair{S}} \cap \maxcolsupp{\pair{S}} \subseteq \unique(\Sigma_{\pair{S}})$ and the converse inclusion holds by \Cref{lemma:identical-colsupp-maximal-colsupp}.

	Now suppose $\unique(\Sigma_{\pair{S}}) = \idcolsupp{\pair{S}} \cap \maxcolsupp{\pair{S}}$. Let $\tuplerkone{C} \in \maxrankonesupp{\tuplerkone{S}}$. By \Cref{lemma:varphi-properties}, there is $(\mat{X}, \mat{Y}) \in \idcolsupp{\pair{S}}$ such that $\varphi(\mat{X}, \mat{Y}) = \tuplerkone{C}$. Let $i \in \colsupp(\leftsupp{S})$. By assumption, $i \in \colsupp(\rightsupp{S})$ hence $\rankone{S}{i} \neq \mat{0}$. Since $\tuplerkone{C} \in \maxrankonesupp{\tuplerkone{S}}$, we have $\matcol{\mat{X}}{i} \transpose{\matcol{\mat{Y}}{i}} = \rankone{C}{i} \neq 0$, which means that $i \in \colsupp(\mat{X})$. This is true for any $i \in \colsupp(\leftsupp{S})$, so $\colsupp(\mat{X}) = \colsupp(\leftsupp{S})$. By definition of  $\idcolsupp{\pair{S}}$, $\colsupp(\mat{X}) = \colsupp(\mat{Y})$, and by assumption, $\colsupp(\leftsupp{S}) = \colsupp(\rightsupp{S})$, hence $(\mat{X}, \mat{Y}) \in \maxcolsupp{\pair{S}}$. Since we supposed  $\unique(\Sigma_{\pair{S}}) = \idcolsupp{\pair{S}} \cap \maxcolsupp{\pair{S}}$, we get $(\mat{X}, \mat{Y}) \in \unique(\Sigma_{\pair{S}})$. By \Cref{thm:equivalence-with-EMD}, $\tuplerkone{C} = \varphi(\mat{X}, \mat{Y}) \in \unique(\Gamma_{\tuplerkone{S}})$.
\end{proof}

\section{Proof of \Cref{prop:disjoint-rank-one-supports-unique-EMF}}
\label{app:simple-sc-identifiability}

\begin{proof}
	By \Cref{cor:uniform-id}, it is enough to show that $\maxrankonesupp{\tuplerkone{S}} \subseteq \unique(\Gamma_{\tuplerkone{S}})$ if, and only if, $\tuplerkone{S} := \varphi(\pair{S})$ has disjoint rank-one supports.
	Sufficiency is immediate: given $\tuplerkone{C} \in \maxrankonesupp{\tuplerkone{S}}$ and $\mat{Z} := \mathcal{A}(\tuplerkone{C})$, the entries of $\rankone{C}{i}$ ($1 \leq i \leq r$) can be directly identified from the submatrix $\mat{Z} \odot \rankone{S}{i}$, because the rank-one supports in the tuple $\tuplerkone{S}$ are pairwise disjoint.
	For necessity, suppose there are $i, j \in \integerSet{r}$, $i \neq j$ such that $\supp(\rankone{S}{i}) \cap \supp(\rankone{S}{j}) \neq \emptyset$. Let $(k, l)$ be an index in this intersection. Denote $\mat{E}^{(k, l)} \in \mathbb{B}^{n \times m}$ 
	the canonical binary matrix full of zeros except a one at index $(k,l)$. 
	Observe that $ \mathcal{A}(\tuplerkone{C}) = \sum_{
		t \in \integerSet{r} \backslash \{ i, j\}} \rankone{S}{t} = \mathcal{A}(\bar{\tuplerkone{C}})$, but $\tuplerkone{C} \neq \bar{\tuplerkone{C}}$ with $\tuplerkone{C}, \bar{\tuplerkone{C}} \in \maxrankonesupp{\tuplerkone{S}} \subseteq \Gamma_{\tuplerkone{S}}$ defined as 

	\begin{displaymath}
		\forall t \in \integerSet{r}, \quad \rankone{C}{t} = \begin{cases}
			\mat{0} & \text{if $\rankone{S}{t} = \mat{0}$} \\
			\mat{E}^{(k, l)} & \text{if $t=i$} \\
			- \mat{E}^{(k, l)} & \text{if $t=j$} \\
			\rankone{S}{t} & \text{otherwise}
		\end{cases}, \quad \rankone{\bar{C}}{t} = \begin{cases}
			\mat{0} & \text{if $\rankone{S}{t} = \mat{0}$} \\
			2 \mat{E}^{(k, l)} & \text{if $t=i$} \\
			- 2 \mat{E}^{(k, l)} & \text{if $t=j$} \\
			\rankone{S}{t} & \text{otherwise}
		\end{cases}.
	\end{displaymath}
	 This shows $\mathcal{C}, \bar{\tuplerkone{C}} \neq \unique(\Gamma_{\tuplerkone{S}})$, hence, $\maxrankonesupp{\tuplerkone{S}} \not\subseteq \unique(\Gamma_{\tuplerkone{S}})$, which ends the proof by contraposition.
\end{proof}

\section{Proof of \Cref{lemma:block-structure-product-butterfly}}
\label{app:block-structure-product-butterfly}

The proof of \Cref{lemma:block-structure-product-butterfly} follows from \Cref{rmk:support-partial-prod} and the following lemma.

\begin{lemma}
	\label{lemma:support-of-product} 
	Given two matrix suports $\suppindex{S}{1} \in \mathbb{B}^{m \times r}$ and $\suppindex{S}{2} \in \mathbb{B}^{r \times n}$, for any pair of factors $(\matseq{\mat{X}}{1}, \matseq{\mat{X}}{2}) \in \Sigma_{\suppindex{S}{1}} \times \Sigma_{\suppindex{S}{2}}$, we have $\supp(\matseq{\mat{X}}{1}\matseq{\mat{X}}{2}) \subseteq \supp(\suppindex{S}{1} \suppindex{S}{2})$.
\end{lemma}

\begin{proof}
	If $(i, j) \notin \supp(\suppindex{S}{1} \suppindex{S}{2})$ then $0 = \matindex{(\suppindex{S}{1} \suppindex{S}{2})}{i}{j} = \sum_{k=1}^r \matindex{\suppindex{S}{1}}{i}{k} \matindex{\suppindex{S}{2}}{k}{j}$. As $\suppindex{S}{1}$, $\suppindex{S}{2}$ are binary, this implies that for each $k \in \integerSet{r}$, $\matindex{\suppindex{S}{1}}{i}{k} = 0$ or $\matindex{\suppindex{S}{2}}{k}{j} = 0$. Since $\supp(\matseq{\mat{X}}{\ell}) \subseteq \suppindex{S}{\ell}$ for $\ell \in \{1,2\}$, 
	we obtain $\matindex{\matseq{\mat{X}}{1}}{i}{k} = 0$ or $\matindex{\matseq{\mat{X}}{2}}{k}{j} = 0$ for each $k \in \integerSet{r}$. This yields $\matindex{\left( \matseq{\mat{X}}{1} \matseq{\mat{X}}{2} \right)}{i}{j} = \sum_{k=1}^r \matindex{\matseq{\mat{X}}{1}}{i}{k} \matindex{\matseq{\mat{X}}{2}}{k}{j} = 0$ hence $(i, j) \notin \supp(\matseq{\mat{X}}{1}\matseq{\mat{X}}{2})$. We conclude by contraposition.
\end{proof}

\begin{proof}[Proof of \Cref{lemma:block-structure-product-butterfly}]
	We start by the case $p = 1$ and show that  $\supp(\matseq{\mat{X}}{1} \ldots \matseq{\mat{X}}{\ell}) \subseteq \blockofidentity{1}{\ell}$ for all $1 \leq \ell \leq J$  by induction. This is true for $\ell = 1$, because by \eqref{eq:butterfly-supports} and \eqref{eq:V_p_q}, $\supp(\matseq{\mat{X}}{1}) \subseteq  \bflyindex{1} = \left[ \begin{smallmatrix}
		1 & 1 \\ 1 & 1
	\end{smallmatrix} \right] \otimes \identity{2^{J-1}} = \blockofidentity{1}{1}$. 
	Suppose now that $1 \leq \ell \leq J-1$ and $\supp(\matseq{\mat{X}}{1} \ldots \matseq{\mat{X}}{\ell}) \subseteq \blockofidentity{1}{\ell}$, which is the matrix full of blocks $\identity{N/2^{\ell}}$. 
	Since $\supp(\matseq{\mat{X}}{\ell+1}) \subseteq  \bflyindex{\ell + 1}$, by \Cref{lemma:support-of-product}, we have $\supp(\matseq{\mat{X}}{1}  \ldots \matseq{\mat{X}}{\ell} \matseq{\mat{X}}{\ell + 1} ) \subseteq  \supp(\blockofidentity{1}{\ell}  \bflyindex{\ell+1})$. 
	As $ \bflyindex{\ell+1}$ is block diagonal with blocks of size $\frac{N}{2^{\ell}} \times \frac{N}{2^{\ell}}$ equal to $\matsize{U}{2} \otimes \identity{N/2^{\ell+1}} = \left[\begin{smallmatrix}
		\identity{N/2^{\ell+1}} & \identity{N/2^{\ell+1}} \\
		\identity{N/2^{\ell+1}} & \identity{N/2^{\ell+1}} \\
	\end{smallmatrix} \right]$, we get:
	\begin{equation*}
		\begin{split}
			\blockofidentity{1}{\ell}  \bflyindex{\ell+1} &= \left( \matsize{U}{2^{\ell}} \otimes \identity{N / 2^{\ell}} \right) \left( \identity{2^{\ell}} \otimes \matsize{U}{2} \otimes \identity{N / 2^{\ell+1}} \right) \\
			&\stackrel{\eqref{eq:kronecker}}{=} \left( \matsize{U}{2^{\ell}} \identity{2^{\ell}} \right)  \otimes \left( \identity{N / 2^{\ell}} \left( \matsize{U}{2} \otimes \identity{N / 2^{\ell+1}} \right) \right) \\
			&\stackrel{\eqref{eq:kronecker}}{=} \matsize{U}{2^{\ell}} \otimes \matsize{U}{2} \otimes \identity{N / 2^{\ell+1}} = \matsize{U}{2^{\ell+1}} \otimes \identity{N / 2^{\ell+1}} = \blockofidentity{1}{\ell + 1}.
		\end{split}
	\end{equation*}
	Consequently, $\supp(\matseq{\mat{X}}{1}  \ldots \matseq{\mat{X}}{\ell} \matseq{\mat{X}}{\ell + 1}) \subseteq \supp(\blockofidentity{1}{\ell}  \bflyindex{\ell+1}) = \supp(\blockofidentity{1}{\ell+1}) = \blockofidentity{1}{\ell+1}$. This ends the  induction.
	Now, in the case $p > 1$, the support $\bflyindex{p}$ of size $N \times N$ is block diagonal, where each block is the leftmost butterfly support of size $N / 2^{p-1} \times N / 2^{p-1}$. Applying the previous case to each of these blocks of size $N / 2^{p-1} \times N / 2^{p-1}$ yields the claimed result.
\end{proof}

\section{Complexity bounds of \Cref{algo:butterfly-fact} with full SVDs}
\label{app:complexity-complete-SVD}

\paragraph{Unbalanced tree}
At the (unique) non-leaf node of depth $j \in \{0, \ldots, J-2\}$, the algorithm computes the best rank-one approximation of $N$ submatrices of size $2 \times N/2^{j+1}$. Using a full SVD costs of the order of $2 \times N/2^{j+1} \times 2 = 2 \times N / 2^{j}$, hence the total cost (with unbalanced tree and full SVD) is
$	\sum_{j=0}^{J-2} N \times 2 \times \frac{N}{2^{j}} 
	= 4 (1 - 2^{-J + 1}) N^2 = 4 \left(1 - \frac{2}{N} \right) N^2 =
	\mathcal{O}(N^2)$.

\paragraph{Balanced tree}
At each of the $2^{k}$ non-leaf nodes of depth $k \in \{0, \ldots, \log_2(J) -1 \}$, the best rank-one approximation of $N$ square submatrices of size $\sqrt{N^{1/2^k}}$ is computed: using a full SVD on each submatrix costs of the order of $(\sqrt{N^{1/2^k}})^{3} = N^{3/2^{k+1}}$.
Since $2^{k} \leq J/2 = \log_{2}(N)/2$ and $N^{1 + 3/2^{k+1}} \leq N^{1 + 3/4} = N^{7/4}$ for $k \geq 1$, the total cost with balanced tree and full SVD is
\begin{equation*}
	\begin{split}
		\sum_{k=0}^{\log_2(J) - 1} 2^{k} \times N \times N^{3/2^{k+1}} &=  N^{5/2} + \sum_{k=1}^{\log_2(J) - 1} 2^{k} N^{1 + 3/2^{k+1}} \\ 
		&\leq N^{5/2} + \sum_{k=1}^{\log_2(\log_{2}(N)) - 1} \frac{\log_{2}(N)}{2} N^{7/4} \\
		&= \mathcal{O}(N^{5/2}).
	\end{split}
\end{equation*}

\section{Proof of \Cref{lemma:butterfly-supp-disjoint}}
\label{app:butterfly-supp-disjoint}

\begin{proof}
	Denote $\tuplerkone{S} := \varphi ( \bflypartial{p}{\ell}, \transpose{\bflypartial{\ell+1}{q}} )$, and $I_k := \{ (k-1) N / 2^\ell + 1, \ldots, k N / 2^{\ell} \}$ for $k \in \integerSet{2^\ell}$. 
The right support $\transpose{\bflypartial{\ell+1}{q}}$, which is equal to $\bflypartial{\ell+1}{q}$ by \Cref{rmk:support-partial-prod}, is block diagonal, with blocks $\blockofidentity{\ell+1}{q}$ of size $N / 2^\ell \times N / 2^\ell$. 
	Hence, for $i \in I_k$ and $j \in I_{k'}$ with $k, k' \in \integerSet{2^\ell}$,  $k \neq k'$, the rank-one supports $\rankone{S}{i}$ and $\rankone{S}{j}$ are disjoint. The columns supports $\{ \supp(\matcol{(\bflypartial{p}{\ell})}{i}), \; i \in I_k \} $ are pairwise disjoint for each $k \in \integerSet{2^\ell}$, by definition of $\bflypartial{p}{\ell}$ and $\blockofidentity{p}{\ell}$.
	This means that for $i, j \in I_k$ ($k \in \integerSet{2^\ell})$, the rank-one supports $\rankone{S}{i}$ and $\rankone{S}{j}$ are also disjoint, when $i \neq j$. 
\end{proof}

\section{Proof of \Cref{lemma:no-zero-col-row}}
\label{app:no-zero-col-row}

\begin{proof}
	{(i) $\Rightarrow$ (ii)}: consider $p,\ell,q$ such that the partial products are made of a single factor. 
	{(ii) $\Rightarrow$ (i):}
	Given $\ell \in \{1, \ldots, J-1\}$, we show by backward induction that $\partialproduct{\mat{X}}{p}{\ell}$ has no zero column for each $p \in \{1, \ldots, \ell\}$. By assumption, $\partialproduct{\mat{X}}{\ell}{\ell} = \matseq{\mat{X}}{\ell}$ has no zero column. Let $p \in \{2, \ldots, \ell\}$, and suppose that $\partialproduct{\mat{X}}{p}{\ell}$ has no zero column. For $i \in \integerSet{N}$,
	the $i$-th column of $\partialproduct{\mat{X}}{p-1}{\ell} = \matseq{\mat{X}}{p-1} \partialproduct{\mat{X}}{p}{\ell} $ is a linear combination of columns $\{ \matcol{(\matseq{\mat{X}}{p-1})}{k} \; | \; k \in \supp(\matcol{(\partialproduct{\mat{X}}{p}{\ell})}{i}) \}$. 
	By \Cref{lemma:hierarchical-level-id-butterfly-supports}, the supports of the rank-one contributions in $\varphi ( \matseq{\mat{X}}{p-1} , \transpose{\partialproduct{\mat{X}}{p}{\ell}} )$ are pairwise disjoint, hence the column supports $\{ \supp(\matcol{(\matseq{\mat{X}}{p-1})}{k}) \; | \; k \in \supp(\matcol{(\partialproduct{\mat{X}}{p}{\ell})}{i}) \}$ are pairwise disjoint. 
	But $\supp(\matcol{(\partialproduct{\mat{X}}{p}{\ell})}{i})$ is not empty as $\partialproduct{\mat{X}}{p}{\ell}$ has no zero column. As $\matseq{\mat{X}}{p-1}$ is also not empty,  the $i$-th column of $\partialproduct{\mat{X}}{p-1}{\ell}$ is non-zero, and this is true for each $i \in \integerSet{N}$, which ends the induction.
	A similar induction shows that $\partialproduct{\mat{X}}{\ell+1}{q}$ has no zero row for each $2 \leq \ell +1 \leq q \leq J$. 	
\end{proof}

\end{document}